# Multi-spectral Image Panchromatic Sharpening – Outcome and Process Quality Assessment Protocol

Andrea Baraldi[a], Francesca Despini[b], and Sergio Teggi[b]

***Abstract*— Multi-spectral (MS) image panchromatic (PAN)-sharpening algorithms proposed to the remote sensing community are ever-increasing in number and variety. Their aim is to sharpen a coarse spatial resolution MS image with a fine spatial resolution PAN image acquired simultaneously by a spaceborne/airborne Earth observation (EO) optical imaging sensor pair. Unfortunately, to date, no standard evaluation procedure for MS image PAN-sharpening outcome and process is community-agreed upon, in contrast with the Quality Assurance Framework for Earth Observation (QA4EO) guidelines proposed by the intergovernmental Group on Earth Observations (GEO). In general, process is easier to measure, outcome is more important. The original contribution of the present study is fourfold. First, existing procedures for quantitative quality assessment (Q$^2$A) of the (sole) PAN-sharpened MS product are critically reviewed. Their conceptual and implementation drawbacks are highlighted to be overcome for quality improvement. Second, a novel (to the best of these authors' knowledge, the first) protocol for Q$^2$A of MS image PAN-sharpening product and process is designed, implemented and validated by independent means. Third, within this protocol, an innovative categorization of spectral and spatial image quality indicators and metrics is presented. Fourth, according to this new taxonomy, an original third-order isotropic multi-scale gray-level co-occurrence matrix (TIMS-GLCM) calculator and a TIMS-GLCM texture feature extractor are proposed to replace popular second-order GLCMs.**

## I. Introduction

THE goal of this multidisciplinary investigation is the design, implementation and validation by independent means of a novel (to the best of these authors', the first) quantitative evaluation procedure for Earth observation (EO) multi-spectral (MS) image panchromatic (PAN)-sharpening outcome and process, in compliance with (conditioned by): (i) human vision, considered as a reference baseline, and (ii) the Quality Assurance Framework for Earth Observation (QA4EO) guidelines, delivered by the intergovernmental Group on Earth Observations (GEO) [1]. This technological research and development (TRD) project is of potential interest to the computer vision discipline and to the relevant segment of the remote sensing (RS) community whose demand for effective, efficient and easy-to-use EO image understanding systems (EO-IUSs) is ever-increasing with the quality and quantity of spaceborne/airborne EO images [2], [3].

According to the ongoing Global Earth Observation System of Systems (GEOSS) implementation plan for years 2005-2015 [4] and to the QA4EO guidelines [1], both delivered by GEO, the visionary goal of providing "the right (geospatial) information, in the right format, at the right time, to the right people, to make the right decisions" requires two necessary and sufficient key principles to be met: "accessibility" and "suitability/reliability" of input RS data, processes and output information products. According to philosophical hermeneutics, information is meant to be either quantitative (non-equivocal) *information-as-thing* [5], e.g., values of a leaf area index are estimated from sensory data [6], or qualitative (equivocal) *information-as-data-interpretation* [7], e.g., land cover (LC) classification and LC change (LCC) detection maps are derived from EO images [8], [9]. In greater detail, the GEO's key principle of "suitability/reliability" relies on mandatory calibration and validation (*Cal/Val*) activities, whose implementation becomes critical to sensory data, process and product quality assurance.

(i) *Cal* activities. An appropriate coordinated program of calibration activities throughout all stages of a spaceborne/airborne mission, from sensor building to end-of-life, is considered mandatory to ensure the harmonization and interoperability of multi-source multi-temporal remote sensing

Manuscript received August 1, 2015; revised December 31, 2015 and July 1, 2016. First published December 4, 2016; current version published February 24, 2017.

In this work, Andrea Baraldi was supported in part by the National Aeronautics and Space Administration under Grant/Contract/Agreement No. NNX07AV19G issued through the Earth Science Division of the Science Mission Directorate.

Francesca Despini and Sergio Teggi were funded by the Agenzia Spaziale Italiana (ASI), in the framework of the project "Analisi Sistema Iperspettrali per le Applicazioni Geofisiche Integrate - ASI-AGI" (n. I/016/11/0).

A. Baraldi was with the Dept. of Geographical Sciences, University of Maryland, College Park, MD 20742, USA. He is now with the Dept. of Agricultural and Food Sciences, University of Naples Federico II, Portici (NA), Italy (e-mail: andrea6311@gmail.com).

F. Despini (email: francesca.despini@unimore.it) and S. Teggi (email: sergio.teggi@unimore.it) are with the Dept. of Engineering "Enzo Ferrari" (DIEF), University of Modena and Reggio Emilia, Italy.

(RS) data [1]. By definition, radiometric calibration is the transformation of dimensionless digital numbers (DNs) into a community-agreed physical unit of radiometric measure, e.g., top-of-atmosphere (TOA) radiance (TOARD), TOA reflectance (TOARF), surface reflectance (SURF), etc.

(ii) *Val* activities. By definition, validation (not to be confused with testing. suitable for internal use) is the process of assessing, by independent means to be community-agreed upon, the "standard" quality of process and outcome [10]. In greater detail, each RS data processing stage and output product must be assigned with metrological/statistically-based (quantitative) quality indicators (QIs), to be community-agreed upon, featuring a degree of uncertainty in measurement at a known degree of statistical significance, to comply with the general principles of statistics and provide a documented traceability of the propagation of errors through the information processing chain, in comparison with established "community-agreed reference standards" [1].

Quite strikingly, the GEO's *Cal/Val* recommendations, although based on common knowledge, are neglected or ignored in the RS common practice [8], [9]. About *Cal* activities, on the one hand, the RS community regards as baseline knowledge that "the prerequisite for physically based, quantitative analysis of airborne and satellite sensor measurements in the optical domain is their calibration to spectral radiance" ([11], p. 29). More explicitly, according to related works [8], [9], [12], [13], [14], radiometric calibration is a necessary not sufficient condition for automatic interpretation of (for physical model-based inference from) EO imagery. On the other hand, in common practice, first, the word "calibration" is absent from a large portion of papers published in the RS literature. Second, the large majority of selectable algorithms implemented in commercial EO image processing software products does not consider radiometric calibration as mandatory [14]. Relaxation of the GEO's *Cal* constraint implies that the RS community heavily relies on statistical (inductive inference) systems, which are inherently ill-posed [15], semi-automatic and site-specific [16], whereas physical (deductive inference) models are largely neglected. Although statistical systems do not require as input sensory data provided with a physical meaning, they may benefit from *Cal* activities in terms of augmented robustness to changes in the input dataset. This is tantamount to saying that, whereas dimensionless sensory data, provided with no physical unit of measure, are eligible for use as input to statistical models exclusively, on the contrary, numerical data provided with a physical unit of measure can be input to both physical and statistical models [14]. In compliance with the QA4EO guidelines, the present work considers *Cal* activities mandatory in its further experimental Section V, in spite of the fact it deals with statistical systems exclusively.

With regard to the GEO's *Val* requirements, to date the RS community appears affected by a lack of standard (recognized) evaluation procedures, whose application domain ranges from (qualitative, categorical) RS image classification [12] to (quantitative) RS data fusion [17], [18], [19], [20]. In agreement with other authors [17], [21], Wald defines image fusion as "a formal framework in which are expressed means and tools for the alliance of images originating from different sources. It aims at obtaining information of a greater quality, although the exact definition of 'greater quality' will depend on the application" [22].

The present paper copes with the ongoing lack of a standard *Val* procedure for multi-spectral (MS) image panchromatic (PAN)-sharpening product and process [18]. MS image PAN-sharpening (merging, synthesizing) algorithms aim at taking advantage of the complementary spatial and spectral properties of MS and PAN imaging sensors [23]. Their goal is to deliver as output a fused PAN-sharpened MS image, $MS^*_h$, by injecting into a coarse spatial resolution MS image, $MS_{l,b}$, with b = 1,…,B, where B is the number of spectral channels and l stands for low scale factor (by definition, scale factor = 1 / spatial resolution [24]), the high-pass spatial details conveyed from a fine spatial resolution PAN image, $P_h$, where h > l stands for high scale factor and where the two sensory images, $MS_l$ and $P_h$, are assumed to be acquired (nearly) simultaneously and to depict the same Earth surface. Typical spatial resolutions of a spaceborne MS and PAN imaging sensor pair range from low (> 500 m, e.g., Meteosat Second Generation, MSG) to medium (from 30 m to 500 m, e.g., Landsat-8), high (< 30 m to 5 m, e.g., SPOT-4, SPOT-5, IRS-1C/D LISS III, EO-1 ALI) and very high (< 5 m, e.g., IKONOS-2, QuickBird-2, GeoEye-1, WorldView-2, WorldView-3, PLEIADES-1A/B, SPOT-6/7, FORMOSAT-2) for the MS imaging sensor, while its PAN counterpart features a spatial resolution finer by a factor of two (e.g., Landsat-8, SPOT-4, SPOT-5), three (e.g., MSG, EO-1 ALI) or four (e.g., IKONOS-2, QuickBird-2, GeoEye-1, WorldView-2, WorldView-3, PLEIADES-1A/B, SPOT-6/7, FORMOSAT-2, IRS-1C/D LISS III).

An $MS^*_h$ image synthesized at fine spatial resolution and featuring 'high spectral quality' (whatever this definition means, in line with Wald [22]) is considered crucial for most RS image applications based on the analysis of spectral signatures, from stratigraphic and lithologic mapping [25], to soil and vegetation analysis [26], [27], to digital surface model (DSM) correction [23]. For example, in [23], a 2D elevation map is generated from a stereo PAN image, then it fits also on the PAN-sharpened MS image. In the 2D elevation map, some image-objects (planar segments) are typically affected by no-data, e.g., due to occlusion phenomena. To correct each no-data pixel value in the height map, a neighboring pixel is searched for in the PAN-sharpened MS image which has the most similar color (in all spectral bands) and a non no-data value in the height map. This colorimetric best fitting neighbor's height value is used to fill the missing value in the DSM.

Unfortunately, the peculiar nature of the MS image PAN-sharpening problem requires that no sensory (non-synthesized) MS image "truth" at high spatial resolution, $MS_h$, exists for comparison with the PAN-sharpened MS outcome, $MS^*_h$. If it were not so, the MS image PAN-sharpening problem would cease to exist. It means that MS image PAN-sharpening is an inherently ill-posed problem in the Hadamard sense, whose solution does not exist or is not unique or, if it exists, it is not robust to small changes in the input dataset [28]. As such, it is difficult to solve and requires *a priori* knowledge, in addition to sensory data, $P_h$ and $MS_l$, to become better posed for numerical treatment [15], [29]. Since MS image PAN-sharpening is inherently ill-posed, so it is the quantitative quality assessment ($Q^2A$) of PAN-sharpened MS imagery. This explains why the latter, too, is a much debated issue [18], [19], [20]: due to a lack of interdisciplinary background, the RS community may keep looking for a single "best" quantitative (objective) solution of an inherently ill-posed (visual, cognitive, qualitative, equivocal) problem, where no single "best" solution exists.

To recapitulate, the objective of the present study is to fill the information gap in $Q^2A$ of MS image PAN-sharpening outcome and process, subject to (conditioned by) the following constraints, required to make the inherently difficult (ill-posed) problem at hand better posed for numerical treatment: problem solution(s), if any, must comply with, first, well-known functional principles of human vision, considered as a reference baseline [30]; second, with perceptual visual quality, assessed by human subjects under controlled experimental conditions; third, with the *Cal/Val* requirements, proposed by GEO in the QA4EO guidelines to be enforced by the RS community [1]. In general, process is easier to measure, outcome is more important. Provided with a relevant survey value, the proposed multidisciplinary investigation is of potential interest to the computer vision community, which includes RS scientists and practitioners involved with EO-IUS activities, see Fig. 1 [31], [32].

The rest of this paper is organized as follows. Section II provides the problem background. In Section III, existing PAN-sharpened MS image quality estimation procedures are critically revised. Materials and methods adopted in the experimental session are described in Section IV, where a novel protocol for $Q^2A$ of the MS image PAN-sharpening outcome and process is proposed. Experimental results are presented in Section V and discussed in Section VI. Conclusions are reported in Section VII.

## II. Problem Background

According to Section I, MS image PAN-sharpening algorithms form a subset of the parent-class of inductive (bottom-up) data learning algorithms for function regression [15], [33], where no target 2D function, $MS_h$, exists. Hence, an output product, $MS^*_h$, must be synthesized (extrapolated) based on *a priori* knowledge (assumptions) in addition to sensory data. In the machine learning discipline, it is common knowledge that inductive data learning problems (either supervised data learning for function regression or classification [15], [29], or unsupervised data learning for vector quantization [34], [35], [36], [37], [38], vector clustering [39], [40], [41], [42], density function estimation or entropy maximization [39]) are inherently ill-posed in the Hadamard sense [28]. It means they are difficult to solve and require prior (top-down, deductive) knowledge in addition to data to become better posed for numerical treatment [15], [29]. By definition, *a priori* knowledge is any knowledge available in addition ("from the earlier", top-down) to the (quantitative) dataset at hand. In common practice, inductive data learning algorithms are semi-automatic (depending on system's free-parameters to be user-defined) and site-specific (depending on training data to learn from, by induction) [16], [43]. Typically, an inherently ill-posed inductive data learning algorithm is provided with prior knowledge in three forms [15]: (i) at the level of understanding of the system's design (architecture), where an inference function (e.g., Bayesian inference) is selected for maximization/minimization purposes, (ii) at the level of understanding of the system's algorithm, where a class of approximating functions (e.g., radial basis function, polynomial function, etc.) is selected together with a model complexity term, e.g., a term capable of regularizing (smoothing) the function regression solution to avoid (exact) function interpolation, (iii) in the initialization phase, when the system's free-parameters are user-defined based on heuristic (qualitative) criteria, which decreases the degree of automation of the statistical model. In addition to these traditional forms of prior knowledge adopted by the parent-class of inductive data learning systems, the special subcategory of MS image PAN-sharpening algorithms requires an *a priori* model of the target 2D function, $MS_h$, to be approximated by the fused image, $MS^*_h$. To recapitulate, the subcategory of MS image PAN-sharpening algorithms is "more" ill-posed than traditional inductive data learning algorithms for function regression. Since it lacks a quantitative "truth" to approximate, the former subcategory rather belongs to the class of inherently ill-posed cognitive problems, like vision (image understanding) in general [44], [45], and early-vision in particular [46], e.g., image segmentation [47], where there is no known cost function to minimize. This consideration justifies the interdisciplinary scenario sketched in Fig. 1. If these inter-disciplinary relationships hold, but are not fully acknowledged by individual scientific communities, consequences may be dreadful. For example, due to an underestimation of the inherent complexity (ill-posedness) of cognitive problems, an ever-increasing number of alternative MS image PAN-sharpening algorithms is expected to be submitted for consideration for publication in the RS and computer vision literature in the close future, exactly like tens of "novel", supposedly "better", inherently ill-posed image segmentation and contour detection algorithms are being published each year. Yet-another "better" solution in a class of (inherently ill-posed) inductive data learning algorithms, where no "single best solution" exists, means that alternative solutions differ one another in the degree of prior knowledge employed to become better conditioned for numerical treatment. Hence, when dealing with inductive learning-from-data algorithms, the focus of scientific attention for discrimination and quality improvement should shift from algorithms to initial conditions, consisting of an *a priori* (deductive) knowledge available in addition to data.

As reported in Section I, the recognition by the RS community of standard procedure(s) for $Q^2A$ of PAN-sharpened MS images is a controversial problem whose solution would be of the utmost importance [18], [19], [20], in accordance with the QA4EO recommendations [1]. In general, it is well established that any data enhancement process (data pre-processing stage), including image fusion, whose input and output variables are quantitative (*information-as-thing*, refer to Section I), is required to assess the quality of the output data (expected to be of "greater quality" [22]) in comparison with the quality of the input dataset(s), according to a (dis)similarity metric [44]. Unfortunately, in the specific case of $Q^2A$ of PAN-sharpened MS imagery, this comparison is particularly difficult because, in the absence of a full-resolution image "truth", $MS_h$, the sensory image pair, $P_h$ and $MS_l$, and the output data product, $MS^*_h$, to be compared feature a different spatial or spectral resolution [48]. About image QIs and quality metrics, the following general considerations hold.

### *A. Quantitative Image Quality Metrics: Signal Fidelity Measures and Perceptual Visual Quality Metrics*

In the words of Iqbal and Aggarwal: "frequently, no claim is made about the pertinence or adequacy of the digital models as embodied by computer algorithms to the proper model of human visual perception... This enigmatic situation arises because research and development in computer vision is often considered quite separate from research into the functioning of human vision. A fact that is generally ignored is that biological vision is currently the only measure of the incompleteness of the current stage of computer vision, and illustrates that the problem is still open to solution" [30].

Objective (quantitative) quality evaluation for images and video can be classified into two board types: signal fidelity measures and perceptual visual quality metrics (PVQMs) [49], [55].

The signal fidelity measures refer to the traditional MAE (mean absolute error), MSE (mean square error), SNR (signal-to-noise ratio), PSNR (peak SNR), etc. Although they are simple, well defined, with clear physical meanings and widely accepted, signal fidelity measures can be a poor predictor of perceived visual quality, especially when the noise is not additive. For example, MAE and MSE are pixel-by-pixel differences, i.e., these statistics are non-contextual and position-dependent. Since they consider a (2D) image as a (0D) string of pixels, i.e., they ignore contextual image information, therefore they are inconsistent with visual perception. In addition, being image position-dependent, they are sensitive to image rotations.

According to a relevant portion of the computer vision literature, *the primary use of image quality metrics is to quantitatively measure an image quality that correlates with perceptual visual quality*. So-called perceptual visual quality metrics, PVQMs, are *objective models for predicting subjective visual quality scores*, like the resultant mean opinion score (MOS) obtained by many observers through repeated viewing sessions [47], [50], [51], [55]. In spite of the recent progress in related fields, objective evaluation of picture quality in line with human perception is still a long and difficult odyssey due to the complex, multi-disciplinary nature of the problem (related to physiology, psychology, vision research and computer science) [55]. For example, cognitive understanding, prior knowledge and interactive visual processing (e.g., eye movements) influence the perceived quality of images; this is the so-called cognitive interaction problem [61]. A human observer will give different quality scores to the same image if s/he is provided with different instructions. Prior information regarding the image content, or attention and fixation, may also affect the evaluation of the image quality. But most image quality metrics do not consider these effects, they are difficult to quantify and not well understood [61]. It is clear that, unlike so-called signal fidelity measures, PVQMs have to quantify the spatial difference (e.g. Position difference in image contours) together with the spectral difference (e.g., image-wide difference in spectral means) between a reference and test image pair [12], [52], [53], [54]. There are two major categories of PVQMs with regard to reference requirements: double-ended and single-ended. Double-ended metrics require both the reference (original) signal and the test (processed) signal, and can be further divided into two subclasses: reduced-reference (RR) metrics that need only part of the reference signal and full-reference (FR) ones that need the complete reference signal. Single-ended metrics use only the processed signal, and are therefore also called no-reference (NR) ones. Most existing PVQMs are FR ones [55], e.g., the popular univariate (one-channel) "universal" (scalar) image quality index (UIQI), or Q index for brevity [60], which was further generalized into the so-called structural similarity (SSIM) index [55], [61]. Noteworthy, although SSIM is considered a PVQM, it does not appear to be provided with a perceptual relevance on a strong theoretical ground [55], in fact SSIM bears both a statistical link and a formal connection with traditional signal fidelity measures, such as the conventional pixel-based MSE [137]. Important conclusions reported in [137] are quoted as follows: "In both an empirical study and a formal analysis, evidence of a relationship between the increasingly popular SSIM and the conventional MSE is uncovered. This research is perhaps the first to uncover a statistical link of this nature and likely the only in which a formal connection is established… Collectively, these findings suggest that the performance of the SSIM is perhaps much closer to that of the MSE than some might claim. Consequently, one is left to question the legitimacy of many of the applications of the SSIM. Ultimately, this investigation once again illustrates the enormous gap that continues to exist between an automated measure of image quality and that of the human mind. Until a more radical approach is considered, this problem will likely continue to confound researchers in the field."

To recapitulate, in a PVQM, quantitative spatial and spectral (2D) image QIs must to be estimated jointly, to be validated by the MOS collected from a group of human subjects [55], [61], e.g., refer to [18] for a detailed description of a visual analysis of PAN-sharpened MS images.

### *B. Non-Injective Property of Summary (Gross) Characteristics*

It is common knowledge that any QI (or summary statistic) is inherently non-injective [56]. The non-injective property of summary statistics or (gross) QIs means that no "universal" QI can exist, because two different instantiations of the same target complex phenomenon can feature the same summary statistic. For example, Zhang and Lu duly observe that semantically "simple" (intuitive to use) planar (2D) shape descriptors are not suitable as standalone descriptors, but a combination of descriptors (feature/error pooling [55], [61], 134]) is necessary in order to accurately describe planar shapes [57]. In economic studies, the popular gross domestic product should never be considered *per se*, but in a minimally dependent and maximally informative (mDMI) [58] combination with other QIs like, for example, the Gini index, estimating the inequality of income or wealth, the pollution/environmental quality, etc. [59]. In practice, the design and development of an mDMI vocabulary of QIs allows to enforce a convergence-of-evidence approach, which is a key decision strategy in cognitive systems [44], [58]. The apparently well-known non-injective property of any QI is in contrast with a search for "universal" QIs traditionally pursued by significant portions of the scientific community. For example, the popular univariate (one-channel) "universal" (scalar) image quality index (UIQI), or Q index for brevity [60], which was further generalized into the so-called structural similarity (SSIM) index [55], [61], were both developed by the computer vision community. Evaluation procedures for PAN-sharpened MS outcome, based on the "universal" Q index, are widely adopted by the RS community [17], [48], [52], [62], [63], [64]. In greater detail, the four-channel "universal" image quality index Q4 and its extension to 2n bands, $Q2^n$ [64], are multivariate generalization of the popular univariate Q index [55], [60]. Like Q, its Q4/$Q2^n$ extensions are logical AND-combinations of three different factors [17], [52], [62], [63], [64]. The first is the modulus of the hypercomplex (multivariate pixel-based) correlation coefficient in range [1-, 1]. The second and third terms measure, respectively, the normalized degree of similarity (in range [0, 1]), estimated across spectral bands, between two univariate (one-channel) means and two univariate standard deviations. In practice, any so-called "universal" Q/Q4/$Q2^n$ index is a (weighted) mixture (e.g., a logical AND-combination) of heterogeneous QIs, each featuring its own unit of measure (if any), domain of change and sensitivity to changes in input data, into one "ultimate" (universal) scalar QI, to be dealt with by univariate analysis. While pursuing dimensionality reduction, the Q/Q4/$Q2^n$ indexes can cause an information loss. To comply with the non-injective property of QIs, a viable strategy alternative to searching for a "universal" QI (which cannot exist) is to develop an mDMI set of individual QIs (independent random variables) to be dealt with by multivariate analysis [58]. In this context, multivariate analysis of heterogeneous QIs is intended as a synonym of a convergence-of-evidence approach [44], such that converging sources of weak (fuzzy), but independent evidence allow to infer strong conjectures [43], in accordance with the general principles of fuzzy logic [65], [66].

### *C. Multi-scale Image Statistics*

Perceptual image quality is inherently multi-scale in the 2D spatial domain, known that human pre-attentive vision adopts at least four spatial scales of analysis to capture non-stationary planar (2D) statistics [67], [68], [69], [70]. It means that no "best" spatial scale exists in vision. Rather, a single well-designed battery of multi-scale spatial filters is necessary and sufficient to solve any possible visual problem [49]. Due to the central limit theorem, any "big data" distribution, like a summary (gross) statistic estimated image-wide from non-stationary local statistics, tends to have a Gaussian shape, where individual contributions of independent (non-stationary) random variables (like basis functions) become indistinguishable from the whole [13]. In other words, global (image-wide) statistics are likely to average over non-stationary local patterns in data. For example, global (image-wide) bivariate Pearson's correlation coefficient (PCC) values are scale invariant only when the original image pair is strongly correlated. Otherwise, PCC values may change with spatial scale (e.g., simulated by image resampling) by more than 20%. In [52], [54], [60], it was observed that if the image-wide PCC statistic is replaced with the average of spatially local PCC values, then the latter computation is much less sensitive to changes in scale. Noteworthy, the size of the moving window required to estimate spatially local PCC/Q/Q4/ $Q2^n$ values is one system's free scale parameter. Unfortunately, first, it must be user-defined based on heuristics. Second, no single spatial scale is sufficient to solve visual problems, different from toy problems [67], [68], [69], [70].

Unlike bivariate PCC statistics, popular univariate summary statistics, like image mean and standard deviation also employed in the Q/Q4/ $Q2^n$ indexes [60], are more robust to changes in scale, which is easy to prove when the resampling algorithm is the nearest-neighbor.

### *D. Yellot's Theory of Low-Level Vision for Texture Discrimination: The Triple Autocorrelation Uniqueness (TAU) Theorem*

The long-disproved Julesz conjecture concerning texture discrimination in biological vision states that pre-attentive discrimination of textures is possible only for textures that have different 2nd-order autocorrelation statistics (univariate statistics of the 2nd-order in the spatial domain). Many counter-examples to this theorem have subsequently been discovered by Julesz and co-workers as well as by other independent researchers [71], [72], [73], [74]. In other words, it is possible to construct pairs of physically distinct texture images whose 2nd-order univariate statistics are exactly identical. This simple background knowledge found in existing literature has an important practical consequence: it implies that popular $2^{nd}$-order spatial statistics, extracted from a gray-level co-occurrence matrix (GLCM) implemented in nearly all existing RS image processing software toolboxes, are inadequate for texture assessment and comparison purposes [75], [76]. Actually, in a more recent paper Yellott appeared to reintroduce the validity of $2^{nd}$-order spatial statistics, by proving that every discrete, finite image is uniquely determined by its two-dimensional dipole histogram [135].

In the context of more recent re-thinking on this subject, Julesz synthesized his studies of pre-attentive texture discrimination as follows: "In essence, we found that texture segmentation is not governed by global (statistical) rules, but rather depends on local, nonlinear features (textons)." As a consequence, "contrary to common belief, texture segmentation cannot be explained by differences in power spectra" (which are image-wide statistics, rather than local statistics). In other words, in biological vision, the neural computations are inherently local in the 2D spatial domain; next, a spatial average is superimposed on the local computational processes. For example, the overall amount of contrast is a visually salient feature which survives this averaging process, although the precise position of each contrast element does not survive the averaging process [74].

In a more recent paper, Yellott stated the following [71].

- Given a discrete image (2D) array, I(c, r), c = 1, …, C, r = 1, …, R, consisting of C columns and R rows, the discrete image-wide 1st-, 2nd-, and 3rd-order spatial statistics are defined respectively as:

$$a_{1,I} = \frac{1}{C \cdot R} \sum_{c=1}^{C} \sum_{r=1}^{R} I(c,r), \quad (1)$$

$$a_{2,I}(n,m) = \frac{1}{C \cdot R} \sum_{c=1}^{C} \sum_{r=1}^{R} I(c,r) I(c+n, r+m) \quad (2)$$

$$= \frac{1}{C \cdot R} \cdot AutocrrltnFnctn,$$

$$a_{3,I}(n_1, m_1, n_2, m_2) =$$

$$\frac{1}{C \cdot R} \sum_{c=1}^{C} \sum_{r=1}^{R} I(c,r) I(c+n_1, r+m_1) I(c+n_2, r+m_2)$$

$$= \frac{1}{C \cdot R} \cdot TripleAutoCrrltnFnctn, \quad (3)$$

where Eq. (2) is the so-called continuous autocorrelation function (up to a multiplicative factor), while Eq. (3) is known as the third-order continuous autocorrelation function (up to a multiplicative factor).

- In a black and white (binary) image of finite size, the image-wide third-order statistics are equivalent to the image-wide triple autocorrelation function, which is a generalization of the ordinary image-wide autocorrelation function.
- In a black and white (binary) image of finite size, the image-wide second-order statistics are equivalent to its image-wide autocorrelation function. For images with more than two gray levels, this equivalence breaks down, i.e., two images can have the same autocorrelation function, but different 2nd-order statistics.
- Discrimination between textured images of finite size becomes increasingly difficult as their image-wide third-order statistics become more similar.
- The Yellott's Triple Autocorrelation Uniqueness (TAU) theorem states that every panchromatic (one-channel multi-gray leveled) image of finite size is uniquely determined (up to spatial translation) by its image-wide third-order statistics. Let's consider the two following statements.
  - o Statement A: Two panchromatic images of finite size are visually identical (up to spatial translation).
  - o Statement B: Two panchromatic images feature identical image-wide third-order statistics.

The TAU principle affirms that statement B is a necessary condition of statement A, i.e., statement A implies statement B. On the other hand, statement B is a sufficient condition of statement A, i.e., statement B implies statement A.

- Identical image-wide third-order statistics imply identical image-wide 2nd-order statistics.

In commenting Yellott's work, Victor observes the following [74].

- The TAU theorem is computed image-wide, i.e., it applies to images of finite size, while the Julesz conjecture applies to textures conceived as a single infinite image or as an infinite ensemble of finite images (which relates to the property of ergodic textures, such that averages performed over the infinite ensemble of textures can be replaced by spatial averages over a single spatially infinite image extracted from the ensemble). Thus, the TAU theorem does not apply to texture ensembles, i.e., it does not trivialize the Julesz conjecture based on local, rather than global statistics. In practice, TAU, which refers to image-wide third-order statistics in images of finite size, does not hold true.
- Biological vision consists of a set of ill-posed problems, such as shape from shading, shape from texture, structure from motion, etc. [46]. Due to the inherent ill-posedness of the (3-D) scene reconstruction from (2D) imagery, the visual system necessarily makes inferences from partial (incomplete) information, and the discovery of how these inferences are made is what the study of biological vision is all about [44].

By combining the TAU theorem with the inherently ill-posed problem of texture segmentation in pre-attentive vision whose neural computations are inherently local [49], [67], [68], [69], [70], [74], a new version of the Julesz conjecture, hereafter referred to as the Enhanced TAU (ETAU) theorem, is formulated as follows.

"*Two images of either finite or infinite size are visually identical (up to spatial translation) if their local, non-linear, non-specific elements (textons) of texture perception* ("tokens" in the Marr's terminology [77], where tokens are detected in the raw primal sketch of early vision) *have identical third-order spatial statistics; if this occurs, it means that two different textures* (homogeneous spatial distributions of tokens, detected in the full primal sketch of early vision [77]) *are the same texture*."

In this latter statement, concepts like texture element/ texton/ token and texture, where texture is defined as the visual effect generated by a spatial distribution of tokens, are necessarily vague (fuzzy), to account for the inherent ill-posedness of pre-attentive vision [46], [77]. Analogously, the same vagueness holds in the inherently ill-posed early-vision process of texture detection (texture segmentation), dealt with by the pre-attentive visual second stage, known as full primal sketch [46], [77].

A simple relationship between the aforementioned ETAU thesis and biological vision reinforces the former speculation. To date, the human visual system can be seen as a huge puzzle with a lot of missing pieces. Even in the first processing layers of the primary visual cortex (PVC, area V1 of the visual cortex, striate cortex) there remain many gaps, in spite of knowledge acquired by neuroscience [67], [68], [69], [78]. In part, these information gaps are being filled by developing and studying computational models. For example, models of simple, complex and end-stopped cells have been implemented in the last 10 years [79], [80], [81]. However, if we require that a computational model of vision should be able to predict perceptual effects, like the Mach bands illusion, where bright and dark bands are seen at ramp edges, then the number of published vision models becomes surprisingly small [82]. In a rather schematic summary, V1 is the input layer of the visual cortex in both left and right hemispheres of the brain. It is organized in so-called cortical hypercolumns, with neighboring left-right regions which receive input—via the optic chiasm and the lateral geniculate nucleus (NGL)—from the left and right eyes, with small "islands," called the "chromatic" blobs [67], [68], [78]. Traditionally, blobs are believed to consist of color-sensitive cells, called double-opponent cells, (apparently) non-oriented, but sensitive to colors [49]. More recent studies found that many color cells in V1 are also orientation tuned [83]. Differently from double-opponent cells in blob areas, most cells in the large interblob areas are (apparently) selective for orientation, but are not chromatic. In the interblob hypercolumns there are simple (S-)cells, complex (C-)cells and end-stopped cells. Complex cells are thought to receive convergent excitatory connections from several simple cells [67]. A major difference between S- and C-cells is that the former are quasilinear while the latter exhibit a clear second-order nonlinearity [84], [85]. There is general agreement that S- and C-cells serve for line and edge extraction, to accomplish object segregation, categorization and recognition [79], [80]. Unfortunately, there are tens of different computational models trying to explain how S- and C-cells interact for line and edge extraction, e.g., refer to [79], [80], [81], [82], [84], [85].

About end-stopping, there seems to be no sharp distinction between end-stopped and not end-stopped cell populations. Furthermore, end-stopped cells show the well-known characteristics of either simple or complex cells. All this suggests that end-stopping is an attribute added to the simple and complex types [79], [80]. End-stopped cells respond to singularities, like line/edge crossings, vertices and end points. The so-called multi-scale keypoint representation [86], accomplished via end-stopped cells, serves as Focus-of-Attention (FoA) [79], [80]. The information represented at the keypoints complements the edge representation. The edge signal is weak or undefined at points of strong 2D intensity variations such as corners or terminations produced by occlusion. The result are gaps in the contours, and false connections between foreground and background, which make an interpretation of an edge map difficult. One can see that the representation of keypoints indicates precisely these critical locations, like terminations, corners and junctions. Typically, many of the keypoints are located on occluding contours [79], [80].

Since they have a proactive role in contour detection, where they are claimed to be sufficient for edge detection by zero-crossing [77], [87], and since they provide inputs to both C-cells (whatever this cell type does) and end-stopped cells suitable for keypoint detection, S-cells are of key relevance in pre-attentive vision. Typically, they are modelled by complex Gabor (wavelet) functions, or quadrature filters with a real cosine and an imaginary sine component, both with a Gaussian envelope, see Fig. 2 and Fig. 3. If the even-symmetric (real) part of a Gabor local filter is implemented like a second-order derivative of an oriented Gaussian shape, like that shown in Fig. 3(a), then it is: (i) suitable for detecting image contours as zero-crossings of the even-symmetric filtered image, in agreement with the Marr's theory of early vision [77], [87], and (ii) eligible for collecting 3rd-order spatial statistics, like those envisaged by the Yellott's ETAU principle.

To conclude, the ETAU speculation finds a physical justification in the multi-scale model of even-symmetric S-cells found in the interblob hypercolumns, as those shown in Fig. 2 and Fig. 3(a), in agreement with the Marr's theory of early vision [75], [77]. Noteworthy, the odd-symmetric (imaginary) part of the same Gabor filter, which is equivalent to a first-order derivative of an oriented Gaussian shape, shown in Fig. 3(b), would be eligible for collecting 2nd-order spatial statistics, like those envisaged by the long-disproved Julesz conjecture about texture discrimination.

### *E. Criteria for Quality Improvement of Existing PAN-Sharpened MS Image Estimation Procedures*

Well-grounded in common knowledge and in the existing literature (refer to Section II.A to Section II.D), four criteria are proposed to be adopted in the further Section III for quality improvement of existing estimation procedures for PAN-sharpened MS outcome.

(I) Quantitative planar (2D) spatial and spectral QIs are estimated together with perceptual (qualitative) image quality values collected from a group of human subjects: yes/no. If no, the estimation procedure is lacking in terms of reference (prior) knowledge to be considered as "truth".

(II) The same (homogeneous) multi-scale image statistic, e.g., a spectral local mean, is combined across spatial scales: yes/no. If yes, on theory, this combination of information (feature/error pooling [55], [61], [134]) is acceptable, because an overall information gain can be accomplished. For example, appropriate multi-scale spatial filter combinations allow detection of color image contours [49]. In practice, any multi-scale combination of homogenous information ought to be further scrutinized at the level of understanding of the algorithm's implementation, to check whether or not this combination of information sources leads to an information gain. If spatial statistics are collected either pixel-based (1$^{st}$-order spatial statistics) or at a single spatial scale (like local PCC/Q/Q4/Q2$^{n}$ indexes [55], [60], [64]), then these statistics are likely to be inadequate to capture inter-image similarities featuring up to 3$^{rd}$-order spatial autocorrelation properties, according to the Yellott's ETAU principle (refer to Section II.D).

(III) Multiple heterogeneous statistics, e.g., image-wide spectral mean and variance, each featuring its own unit of measure (if any), domain of variation and sensitivity to changes in input data, are combined into a "universal" QI, which cannot exist, due to the non-injective property of summary statistics: yes/no. This mixture of heterogeneous random variables into one scalar "universal" QI is, in general, subject to a loss of information, due to dimensionality reduction; hence, in general, it is theoretically inconvenient, in particular when either of the following conditions occurs.

• It is based on a heuristic (subjective, equivocal) weighted combination of individual terms, where weights are user-defined based on empirical criteria. These weights increase the number of system's free-parameters to be user-defined. Their total number is monotonically decreasing with the system's degree of automation (ease of use) [8], [9].

• Heterogeneous terms are combined without harmonization of their units of measure, domains of variation and sensitivities to changes in input data. For example, in [88], a multi-source geospatial index of climate change adopts a linear min-max normalization function to render each input dataset comparable in (normalized) range of change and (dimensionless) unit of measure. Unfortunately, a linear min-max normalization function applied to different data sources (random variables) does not harmonize their sensitivities to changes in the input dataset. The so-called z-score (standard score of a raw score, standardized variable) would be a better solution [89].

(IV) The popular bivariate PCC in range [-1, 1] is adopted as an image QI, irrespective of its local spatial scale of analysis, refer to the aforementioned point (II), whether or not it is combined with other QI indexes, refer to the aforementioned point (III): yes/no. In general, exploitation of PCC as an inter-image QI and metric should be considered theoretically inconvenient. The well-known sensitivity of the PCC to linear transformations of the two random variables means that PCC is maximum (in absolute terms) between two images that are either identical or one the linear transformation of the other although, in this latter case, they can look (perceptually) very different. Its macroscopic inconsistency with (its independence from) visual perception should discourage perceptual image (dis)similarity metrics from using the PCC as an input variable. Similar considerations led to neglect the estimation of correlation as a viable texture feature from popular 2nd-order GLCMs [75].

## III. Critical Review of Existing Procedures for $Q^2A$ of PAN-Sharpened MS Images

State-of-the-art procedures for $Q^2A$ of PAN-sharpened MS outcomes belong to two families, depending on whether or not the sensory $MS_l$ image is adopted as a reference dataset.

### *A. Abstract Three-Statement Wald's Protocol, where an Ideal Reference Image at Fine Resolution is Available for Comparison Purposes*

Let us assume that together with the sensory $P_h$ and $MS_l$ images acquired simultaneously by a PAN and MS sensor pair at spatial scales h and l respectively, with h > l, an ideal reference $MS_h$ is also available as "truth", like it were acquired by the same MS sensor capable of working at high and low spatial scales simultaneously. The Wald's protocol is based on the following three PAN-sharpened MS image quality assessment criteria [48].

1. Any fused image, $MS^*_h$, if spatially degraded from scale h to l, identified as $MS^*_{h\text{->}l}$, should be as nearly identical as possible to the original $MS_l$ image. For example, a channel-specific difference between images $MS^*_{h\text{->}l,b}$ and $MS_{l,b}$, b = 1, …, B, can be computed on a per-pixel basis ([48], p. 694). This property, called the consistency property, is a necessary, but not sufficient condition for image fusion, i.e., its fulfillment does not imply a correct fusion [18]. In practice, there is an influence of the downsampling strategy upon the results of comparison between $MS^*_{h\text{->}l}$ and $MS_l$, but this influence can be kept small, provided the MS image downsampling operator is such that, first, an appropriate low-pass filter (LPF), whose transfer function has to match the average modulation transfer function (MTF) of the MS sensor [90], is applied to the MS image ([48], p. 694). Second, a decimation operator, characterized by a sampling factor equal to the spatial scale ratio (h: l) between the two native scales of images [63], is applied to the low-pass filtered MS image.

2. Each band of the synthetic image $MS^*_{h,b}$, b = 1, …, B, should be as identical as possible to its ideal reference counterpart $MS_{h,b}$, b = 1, …, B. Since this property does not cope with the entire set of channels simultaneously, then a third consistency property is required.

3. As a whole (i.e., when all channels are examined simultaneously), the synthetic $MS^*_h$ image should be as identical as possible to the ideal reference image $MS_h$.

### *B. Quantitative Analysis with the Sensory $MS_l$ Image Adopted as Reference: Revised Two-Statement Wald's Protocol and Its One-Statement Simplified Version*

In [48], the second and third virtual properties of the abstract Wald's protocol are implemented as follows.

• Second property proposed in Section III.A. If the original input images, $P_h$ and $MS_l$, are degraded as $P_{h\text{->}l}$ and $MS_{l\text{->}s}$, where the spatial scale s is such that s < l < h, and if a PAN-sharpened MS image, $MS^*_l$, is synthesized at the native spatial scale l < h starting from degraded images $P_{h\text{->}l}$ and $MS_{l\text{->}s}$, then the fused image, $MS^*_l$, should be as nearly identical as possible to the original $MS_l$ image considered as reference. It is recommended that spatial scale ratio (l: s) is chosen equal to (h: l). This is a realistic strategy to check in practice the synthesis property [18], [20], [48]. To obtain $MS_{l\text{->}s}$, the same constraints about the MS image degradation filter listed in Section III.A hold. In addition, the PAN image degradation filter used to generate $P_{h\text{->}l}$ is typically designed as an ideal filter [90].

• Third property proposed in Section III.A. Assuming that the high-frequency (fine resolution) spatial information is conveyed into the MS image to be synthesized at fine resolution, $MS^*_h$, by the sensory PAN image, $P_h$, a realistic spatial quality assessment of the fused image requires the difference to be ideally null between: (i) bivariate PCCs computed between the downscaled PAN image, $P_{h\text{->}l}$, and each band of the synthesized image, $MS^*_{l,b}$, b= 1, ,…, B, and (ii) bivariate PCCs computed between the same downscaled PAN image, $P_{h\text{->}l}$, and each band of the original image $MS_{l,b}$, with b= 1, …, B (see [48], p. 695).

A typical simplified implementation of the Wald's protocol consists of the aforementioned second property exclusively.

Unfortunately, the practical choice of the pair of LPFs applied to the sensory $MS_l$ and $P_h$ images for downsampling is crucial in these realistic adaptations of the ideal Wald's protocol. An erroneous choice of these LPFs may lead to mismatches between the $Q^2A$ of the image fusion at reduced resolution, $MS^*_l$, and the quality, perceived by visual inspection exclusively (due to the absence of an $MS_h$ image "truth"), of the image fusion outcome at full resolution, $MS^*_h$ [91]. This holds true particularly in the case of MS image PAN-sharpening methods exploiting spatial filters [92].

The proposed realistic adaptation of the abstract three-statement Wald's protocol requires exploitation of both so-called "scalar" QIs (i.e., statistics estimated in a single channel) and so-called "vector" QIs (i.e., statistics estimated in all spectral channels simultaneously), together with (dis)similarity metrics [63]. In [18], one-channel "scalar" statistics are otherwise called "unimodal", whereas multi-channel "vector" statistics are otherwise called "multimodal". In the rest of the present work, these expressions are replaced by terms "univariate" and "multivariate" respectively. Examples of univariate statistics are relative bias (difference in mean), difference in variances, relative difference in standard deviation, and many others [50]. The well-known PCC is a bivariate statistic. The popular UIQI, typically identified as Q index, combines into one scalar value several heterogeneous statistics including PCC [60]; hence, the Q index is also a bivariate statistic. Well-known examples of multivariate summary statistics are Q4, as a generalization of Q [62], and $Q2^n$ as a generalization of Q4 [64], the relative dimensionless global error (ERGAS, Erreur Relative Globale Adimensionnalle de Synthèse) [93], the average spectral angle mapper (SAM) cost index [94] and many others [50]. In [53], image QIs are divided into either spectral or spatial, where examples of the latter category are the Zhou spatial correlation coefficient (ZCC) [95] and the true edge (TE) detector [96].

In the simplified implementation of the Wald's protocol, consisting of the aforementioned second property exclusively, univariate QIs can be omitted, i.e., multivariate QIs can be adopted exclusively.

To make this paper self-contained, the popular univariate Q index and the multivariate SAM and ERGAS cost indexes are presented hereafter. The SAM formulation computes the inter-vector angle between two data vectors $\vec{x}$ and $\vec{y}$ as:

$$SAM(\vec{x},\vec{y}) = \arccos\left(\frac{\langle \vec{x},\vec{y} \rangle}{|\vec{x}|\cdot|\vec{y}|}\right), \quad (4)$$

where $|\cdot|$ and $\langle \; \rangle$ indicate, respectively, the Euclidean norm and the scalar vector product [94]. In the application domain of MS image PAN-sharpening, SAM is adopted to quantify the inter-image pixel-specific MS difference (distorsion), irrespective of the pixel-pair difference in modula (color intensities). An image-wide SAM statistic is the average of the pixel-based SAM values [63], [94]. If there is no inter-image spectral distorsion, then average SAM is zero. It means that, in MS image PAN-sharpening applications, average SAM is a cost (error) index to be minimized.

The ERGAS cost (error) index is a heuristic multivariate estimate of a dimensionless pixel-based inter-image difference adopted by several procedures for $Q^2A$ of PAN-sharpened MS outcome [18], [63], [97]. It is defined as:

$$ERGAS = 100 \cdot \frac{h}{l}\sqrt{\frac{1}{B}\sum_{b=1}^{B}\frac{RMSE(MS_b)^2}{(Mean_b)^2}}, \quad (5)$$

where $Mean_b$ is the mean value of the sensory image $MS_{l,b}$, b = 1, …, B, while the root mean square error term, $RMSE(MS_b)$, is defined as:

$$RMSE(MS_b) = \sqrt{\frac{\sum_{i=1}^{NP}(MS_{l,b}(i) - MS^*_{l,b}(i))^2}{NP}}, \quad (6)$$

where i is a pixel identifier and NP is the total number of pixels. According to Wald, ERGAS exhibits a strong tendency to decrease as the inter-image similarity increases. Typical ERGAS values of "good inter-image quality" range below 3 [93].

In the computer vision literature, Wang and Bovik [60] proposed a so-called "universal" (combined scalar value from heterogeneous statistics) image quality index, UIQI, identified as Q. This is a similarity metric instantiated as an AND-combination of heterogeneous bivariate and univariate statistics extracted from a pair of one-channel images x and y, such that it is maximum when the two one-channel images are the same. In particular:

$$Q = \frac{\sigma_{xy}}{\sigma_x\sigma_y} \cdot \frac{2\overline{x}\,\overline{y}}{(\overline{x})^2 + (\overline{y})^2} \cdot \frac{2\sigma_x\sigma_y}{\sigma_x^2 + \sigma_y^2}, \quad (7)$$

where $\sigma_{xy} = \frac{1}{NP-1}\sum_{i=1}^{NP}(x_i - \overline{x})(y_i - \overline{y})$ is the covariance between one-channel images x and y, with $\overline{x}$ = E[x] and standard deviation $\sigma(x) = \sigma_x \geq 0$. In Eq. (7), the first term is the popular Pearson's cross-correlation coefficient, PCC = $\frac{\sigma_{xy}}{\sigma_x\sigma_y} \in$ [-1, 1], whose absolute value increases if there is a linear relationship between the two univariate random variables x and y. The second term is the normalized similarity (belonging to range [0, 1]) between the image means (called luminance), which is equal to one if and only if the two univariate means are the same. The third term is the normalized similarity (in range [0, 1]) of the two image standard deviations (called contrast), which is equal to one if and only if the two univariate standard deviations are the same. Although it is indeed practical to estimate an inter-image similarity as an overall (image-wide) scalar QI value, an inter-image (dis)similarity measure is typically space variant because image signals are generally nonstationary in the 2D spatial domain. Hence, it is more appropriate to measure the Q index values locally, e.g., based on a non-overlapping moving window of size W1 × W2 in pixel unit, to accomplish an image partition into blocks, and then combine these local values at an image-wide spatial scale, e.g., by averaging the sum of local values [60]. This is equivalent to implementing the Q index at a single-scale of analysis, whereas human vision is known to adopt at least four-scale spatial filters (refer to Section II.C). In more recent years, the so-called SSIM index was proposed as a generalization of the bivariate Q index [61]. A multi-scale implementation of the SSIM index was proposed by the same authors [136].

Although SSIM is considered a PVQM, it does not appear to be provided with a perceptual relevance on a strong theoretical ground, in fact SSIM bears certain similarities with traditional signal fidelity measures, such as the MSE [55]. This is clearly explained in [137] whose conclusions are quoted in Section II.A.

To account for widespread criticisms about the SSIM such as those reported in [137], Simoncelli et al. have recently proposed a PVQM based on a normalized Laplacian pyramid for image analysis and synthesis as a viable alternative to the SSIM. The proposed PVQM formulation is:

$$D(I_R, I_T) = \frac{1}{S}\sum_{s=0}^{S-1} \frac{1}{\sqrt{SF_s}} \left\|\hat{I}_{R,s} - \hat{I}_{T,s}\right\|^2,$$

where $I_R$ and $I_T$ are the reference and the test image respectively, $\hat{I}_{R,s}$ and $\hat{I}_{T,s}$ denote vectors containing the transformed reference and distorted image data at scale s = 0, …, S-1, respectively, and where $SF_s$ is the number of spatial filters in the sub-band at scale s. In this equation a root mean squared error is computed for each scale, and then averaged over these scales giving larger heuristic weights to the lower frequency coefficients (which are fewer in number, due to subsampling).

Since the "univeral" Q index is bivariate, i.e., it applies to two one-channel images exclusively, Alparone *et al.* [62] proposed Q4 as a "universal" Q index extended to four-band image pairs. Like Q, Q4 is computed per image block and, next, averaged across blocks. Like Q, also Q4 is a similarity index, to be maximized in MS image PAN-sharpening applications, made of three factors. The first is the modulus of the multivariate (multi-band) hypercomplex PCC in range [-1, 1]. The second and third terms measure, respectively, the normalized difference across bands of univariate (one-band) mean pairs and standard deviation pairss. Typically, an image-wide Q4 value is computed as average of one-scale local Q4 values estimated across an image partition of 16 × 16 or 32 × 32 blocks. In recent years, Garzelli and Nencini [64] presented a "universal" $Q2^n$ similarity index as a generalization of Q4 for image pairs of more than four bands by a power of 2.

### *C. Quantitative Analysis at High Spatial Scale h, Without Reference Image*

In [52], [54], a new "universal" (combined from heterogeneous statistics) QI, called "quality with no reference", QNR, is proposed for $Q^2A$ of PAN-sharpened MS outcome. This second approach exploits no reference image, but relationships among the two sensory images, $P_h$ and $MS_l$, and the synthesized $MS^*_h$ product exclusively. Hence, it is appealing because its inputs are the two sensory images in addition to the output fused image, at their native scales. Unfortunately, this second approach strongly depends on the choice of QIs and quality metrics. In line with [60], summary statistics should be computed as image-wide averages of one-scale local estimates, to better capture the non-stationarity of image statistics. This is especially true for pixel-based bivariate PCC values that may change with scale by more than 20% (e.g., simulated by image resampling, refer to Section II.C).

In [52], [54], [63], the "universal" QNR similarity metric is implemented as an AND-combination of a spatial similarity index with a spectral similarity index:

$$QNR = (1 - D_\lambda)^\alpha \cdot (1 - D_S)^\beta, \qquad (8)$$

where $D_\lambda$ is the spectral distortion, $D_S$ is the spatial distortion and $\alpha$ and $\beta$ are two coefficients to be user-defined based on heuristics to weight the two terms. The spectral distortion is computed as:

$$D_\lambda = \sqrt[p]{\frac{1}{B(B-1)} \sum_{i=1}^{B} \sum_{j=1, j\neq i}^{B} \left| d_{i,j}(MS_l, MS^*_h) \right|^p}, \qquad (9)$$

where p is a metric parameter to be user-defined based on empirical criteria (it is usually set to 1 [54]) and term

$$d_{i,j}(MS_l, MS^*_h) = Q(MS_{l,i}, MS_{l,j}) - Q(MS^*_{h,i}, MS^*_{h,j}) \qquad (10)$$

is a dissimilarity measure between two "universal" one-band Q index values. The spectral cost function (9) is minimized when the inter-band heterogeneous Q combination of spectral properties at high spatial scale h, within the synthetic $MS^*_h$ image, are the same of their spectral counterparts at low spatial scale l, within the sensory $MS_l$ image. The spatial distortion function is defined as:

$$D_S = \sqrt[q]{\frac{1}{B} \sum_{i=1}^{B} \left| Q(MS^*_{h,b}, P_h) - Q(MS_{l,b}, P_{h->l}) \right|^q}, \qquad (11)$$

where q is a metric parameter to be user-defined based on empirical criteria (it is usually set to 1 [54]). The spatial distorsion Eq. (11) is minimized when the similarity Q index computed at high spatial scale h between each band of the fused $MS^*_h$ image and the $P_h$ image is equal to the similarity Q index computed at low spatial scale l between each band of the sensory $MS_l$ image and the downscaled $P_{h->l}$ image. To recapitulate, QNR is a PAN-sharpened MS image QI in range [0, 1]. To maximize QNR, the spectral distortion term, $D_\lambda$, and the spatial distortion term, $D_S$, must be minimized to zero. It means that QNR is monotonically increasing with the combined spatial and spectral qualities of the fused $MS^*_h$ product.

Alternative formulations of the spatial and spectral cost functions (9) and (11) employ QIs and quality metrics like standard deviation, entropy (He), cross entropy (CE), spatial frequency (SF), fusion mutual information (FMI), fusion quality index (FQI), fusion similarity metric (FSM), etc. [50].

## IV. Materials and Methods

To design, implement and validate by independent means an innovative procedure for $Q^2A$ of MS image PAN-sharpening process and outcome, the following materials and methods were selected.

### *A. Validation Dataset*

According to standard review quality criteria adopted by peer-reviewed journals in computer science, experimental results are expected to be shown for a sufficient number of real and standard/appropriate data sets, typically two or more [98], [99]. For example, to assess the best among alternative MS image PAN-sharpening algorithms in terms of QIs of operativeness (QIOs), encompassing accuracy, efficiency, degree of automation, robustness to changes in the input dataset, etc. [8], [9], at least two input dataset should be considered mandatory.

Rather than selection of a best algorithm among alternative solutions, the goal of the experimental session of the present study is validation of an evaluation procedure, refer to Section I. By definition, validation (not to be confused with testing) is the process of assessing, by independent means, the quality of the information processing system's outputs [10]. In this work, the system under investigation is an evaluation procedure whose outputs are quantitative ranks of PAN-sharpened MS images to be validated against qualitative ranks collected from human subjects independent of the authors of the procedure under validation.

For validation of the proposed evaluation protocol and for the sake of paper brevity, only one validation sensory dataset was selected, to be representative of the complexity of the target phenomenon under investigation, namely, $Q^2A$ of MS image PAN-sharpening outcome and process. The selected sensory dataset consists of a very high resolution (VHR) spaceborne MS and PAN image pair, acquired by the QuickBird-2 imaging sensors on 2004-06-13 at 09:58 a.m. over the Campania region, Italy. The QuickBird-2 imaging sensors simultaneously acquire a PAN image at 0.61 m spatial resolution and a four-band image at 2.44 m resolution, whose spectral channels are: visible blue (B), green (G), red (R) and near-infrared (NIR). Based on ancillary calibration metadata files, both PAN and MS images were radiometrically calibrated into TOARF values, in compliance with the QA4EO requirements (refer to Section I). The radiometrically calibrated MS and PAN images are shown in Fig. 4. This VHR image pair was considered appropriate because, first, it depicts a wide variety of land surface classes, ranging from urban areas to forests and agricultural fields, but also includes real-world RS image noise, where chromatic information is saturated (in white-color image areas, like clouds), null (in black-color image areas, like cloud-shadows) or fuzzy, e.g., image areas affected by haze. Second, it consists of four bands. Hence, this four-band sensory $MS_l$ image and its synthesized $MS^*_l$ versions were not too difficult to be visually assessed by a pool of human subjects, who could rely exclusively on a three-channel RGB monitor for image comparison and quality assessment. Last but not least, this four-band validation image allowed estimation of the popular Q4 index [62].

According to Table 1 (also refer to the further Section IV.B), there were fourteen PAN-sharpened MS image instances to be ranked for validation purposes by the proposed quantitative evaluation procedure in comparison with a qualitative (perceptual) assessment by human subjects, adopted as a reference ("truth") and expected to be mimicked (matched) by the proposed quantitative PVQM approach.

Actually, the proposed validation dataset of fourteen PAN-sharpened MS images, provided with perceptual visual quality ranks as "truth", was sufficient to act as counter-example where popular multivariate scalar QIs (or cost indexes), like average SAM, ERGAS and Q4, adopted by state-of-the-art procedures for PAN-sharpened MS image quality estimation [18], [63], were correlated one another, but fail to be uncorrelated with perceptual visual quality assessed by human subjects under controlled experimental conditions.

For the sake of completeness, we mention that an internal testing phase of the proposed evaluation procedure predated the validation phase, documented in this paper. During tests conducted on several PAN-sharpened MS images, the proposed quantitative estimation procedure was compared with a visual assessment (for acquisition of "truth") by the same authors of the estimation procedure. These same authors considered the experimental degree of match of quantitative test results with their own visual assessment in agreement with theoretical expectations (refer to Section II), to be further confirmed in a validation phase by independent means.

For $Q^2A$ of MS image PAN-sharpening outcome we adopted a reduced resolution approach following the simplified one-statement Wald's protocol [93], refer to Section III.B. In this evaluation framework, according to Vivone *et al.* [63], the original QuickBird-2 PAN and MS images were downsampled, by means of a Gaussian low-pass filter (LPF) and a decimation operator, to a spatial resolution of 2.44 m and 9.76 m respectively, to maintain the same fusion ratio (1:4) as in the original QuickBird image pair. The implemented LPF matched the average modulation transfer function (MTF) of the MS and PAN imaging sensors [90], in agreement with Section III.B.

### *B. MS Image PAN-Sharpening Algorithms Selected for Testing*

For testing purposes we surveyed popular MS image PAN-sharpening algorithms available in several RS image processing commercial software toolboxes, specifically, ERDAS Imagine (licensed by ERDAS, Inc.) [100], Environment for Visualizing Images (ENVI, licensed by ITT Industries, Inc.) and Interactive Data Language (IDL, licensed by ITT Industries, Inc.) [101]. This survey led to a selection of eight algorithms.

- Principal Component (PC) transform [102], implemented by ENVI.
- Gram-Schmidt (GS) transform [101], [103], implemented by ENVI.
- Color Normalized Spectral Sharpening (CN) transform [104], implemented by ENVI.
- Discrete Wavelet (DWT) transform [105], [106], implemented by IDL.
- A Trous Wavelet (ATW) transform [106], [107], [108], implemented by IDL.
- Hyperspherical Color Space (HCS) [109], implemented by ERDAS.
- Ehlers (EH) transform [110], implemented by ERDAS.
- Resolution Merge (RM), [111], implemented by ERDAS.

A description of these eight algorithms is beyond the scope of this paper; interested readers can refer to literature.

For each of these algorithms, there were one or more system's free-parameters to be user-defined. One input parameter was selected for discriminative purposes, i.e., its changes in value led to different runs of the same algorithm with different outcomes. Other input parameters, if any, were kept fixed in the different runs by the same algorithm. Table 1 shows that different resampling methods were selected as input parameter by some of the eight algorithms, which increased to fourteen the total number of alternative MS image PAN-sharpening system implementations to be compared.

When applied to the downsampled version of the validation sensory dataset shown in Fig. 4, these test algorithms generated a fused $MS^*_l$ image at low spatial scale l, as shown in Fig. 5 and Fig. 6.

### *C. State-of-the-Art Multivariate Scalar QIs Selected for Comparison Purposes*

In order to compare image quality estimates by the new evaluation protocol with existing QIs, three multivariate (all-band) scalar QIs were selected as the most widely adopted by the scientific community in recent years.

- SAM dissimilarity index [94].
- ERGAS dissimilarity index [97].
- Q4 similarity index [62].

A review of these indexes is provided in Section III.B. For further details, refer to the literature. About Q4, it was calculated as averages on BL × BL image blocks, with BL = 8. Hence, Q4 depends on BL too, denoted as $Q4_{BL}$. Finally, $Q4_{BL}$ was averaged over the whole image to yield the global score index Q4, in agreement with [62].

### *D. Perceptual Image Quality Assessment*

According to Section II.A, human visual analysis is indispensable to provide the inherently ill-posed MS image PAN-sharpening problem with a reference baseline for quality estimation and comparison, also refer to Section I. Although stated in other terms, this concept is widely acknowledged in works like [18], [50], [91]. Unfortunately, visual quality assessment of multiple images is a difficult and lengthy task to handle because, first, the human visual system is not equally sensitive to various types of distortion or color contrast in an image. Second, the perceived image quality is strongly dependent upon the observer and the thematic application at hand (*information-as-data-interpretation*, refer to Section I). Third, technical factors, such as difficulties in image representation or evaluation, e.g., when MS images have more than three spectral bands, may undermine the validity of the experiment with human subjects. To minimize accidental and systematic errors in visual evaluation, protocols for visual quality assessment have been proposed in the fields of television and image compression [55], image segmentation [47] and EO MS image PAN-sharpening [93].

The goal of the following procedure is to estimate the resultant MOS obtained by many observers through repeated viewing sessions [55], [61]. In agreement with [47], sixteen Modena and Reggio Emilia University staff and students served as subjects, equally split by gender. None was paid for his/her participation. All were native Italian speakers and reported having normal or corrected-to-normal vision. Two categorical variables, identified as Spatial QI and Spectral QI, were assigned with seven levels, from A to G, corresponding to numbers 1 to 7, see Table 2. The original PAN and MS image pair and each test PAN-sharpened MS image were partitioned into BL × BL non-overlapping blocks, with BL = 8. performance of the current quality assessment algorithms.

To account for the cognitive interaction problem, where the perceived quality of images is influenced by prior information [61], before beginning the experimental trials, the subjects received six practice trials, each consisting of the same randomly selected block extracted from all the PAN-sharpened MS images in comparison with the same block extracted from the original PAN and MS image pair, to get familiar with concepts like spatial and spectral qualities (similarities) in four-band images shown in a three-channel (RGB) monitor. The test image is required to be as identical as possible to the reference image, neither better nor worse in perceptual terms [60]. Following this practice, the subjects started on the remaining 58 blocks for spatial and spectral quality (similarity) assessment. Each PAN-sharpened MS image-block was ranked (sorted) by each subject, who was free to change band combinations shown in the RGB monitor. The two spatial QI and spectral QI distributions were estimated per image. Each distribution was standardized (to feature zero mean and unit variance) [89] and the standardized range of change was split into seven bins of the same width, labeled A to G again. Finally, a winner-take-all strategy was adopted. If the difference in score between the first-best and the second-best level was less than 10% of the first-best score, than both levels were considered winners, as shown in Table 2. This MOS procedure is very different from that proposed in [61], where subjects were asked to rank an ensemble of images compared with the same reference image, but were asked to provide their perception of quality of each pairwise image comparison on a continuous linear scale that was divided into five equal regions marked with adjectives "Bad", "Poor", "Fair", "Good" and "Excellent". Raw scores for each subject were normalized by the mean and variance of scores for that subject (i.e., raw values were converted to Z-scores [54]) and then the entire data set was rescaled to fill the range from 1 to 100. MOSs were then computed for each image, after removing outliers.

### *E. Expert System in Operating Mode for Prior Knowledge-Based MS Data Space Discretization (Partitioning)*

Prior knowledge-based (top-down, deductive, physical model-based) preliminary classification (pre-classification) has an important role in the operational, comprehensive and timely generation of information products from EO "big data" [112], in compliance with the QA4Eo guidelines [1]. Documented applications of prior knowledge-based image mapping systems date back to the early 1980s [44], [45] and span from RS image enhancement (data pre-processing), like automatic stratified (conditioned, pre-classified) image co-registration, topographic correction [113], [114] and cloud masking [115], [116], [117], to second-stage (high-level) stratified (conditioned) LC classification and LCC detection [8], [9], [12], [13], [14], [44], [45], [118]. *Equivalent to color naming in a natural language* [119], [120], [121], *prior knowledge-based continuous color space discretization (compression, quantization, partitioning)* *is the automatic deductive counterpart of semi-automatic inductive vector quantization algorithms, like the popular k-means algorithm (also known as Linde-Buzo-Gray algorithm, LBG)* [15], [33], [34], [35], [36], [37], [38], *not to be confused with unsupervised data clustering algorithms, where termination is not based on optimizing any model of the process or its data* [39], [40], [41], [42], [122], [123], [124]. In unsupervised (unlabeled) data quantization problems, the target cost function to minimize is known and equal to a data quantization error (typically, a mean square error). In the machine learning literature, it is common knowledge that any inductive data learning problem is inherently ill-posed in the Hadamard sense [28] and requires *a priori* knowledge in addition to data to become better-posed for numerical solution [15], [33]. More specifically, in a generic data quantization error minimization problem, the quantization error is expected to be monotonically decreasing with the number of quantization levels; hence, no number of quantization levels is "optimum" *per se*. As a consequence, the number of quantization levels is typically user-defined based on subjective criteria, like in the well-known k-means unsupervised data learning algorithm, where the number of quantization levels k is an input parameter to be user-defined. Vice versa, it is possible to provide the target data quantization error as a user-defined input parameter, such that it is the system's free-parameter k to be dynamically learned from data by the inductive data quantization algorithm [34], [35].

The expert system for prior knowledge-based MS data quantization selected in this study was the Satellite Image Automatic Mapper (SIAM), proposed to the RS community in recent years [8], [9], [12], [13], [14], [125]. Since it is based on prior knowledge, SIAM is: (i) independent of (non-adaptive to) data and (ii) fully automatic, i.e., it requires neither input parameters nor training data to run. In agreement with the GEOSS implementation plan [4], the SIAM software product is implemented as an integrated system of four subsystems, including one "master" 7-band Landsat-like subsystem, L-SIAM (whose spectral resolution comprises bands visible blue: B, visible green: G, visible red: R, near infra-red: NIR, medium infra-red 1: MIR1, medium infra-red 2: MIR2, and thermal infra-red: TIR) plus three "slave" (downscale) subsystems, namely, a 4-band Satellite Pour l'Observation de la Terre (SPOT)-like (G, R, NIR and MIR1), S-SIAM, a 4-band Advanced Very High Resolution Radiometer (AVHRR)-like (R, NIR, MIR1 and TIR), AV-SIAM, and a 4-band QuickBird-like (B, G, R and NIR), Q-SIAM, whose spectral resolutions overlap with Landsat's, but are inferior to Landsat's. Noteworthy, an expression like "Landsat-like MS image" adopted in this paper means: "an MS image whose spectral resolution mimics the spectral domain of the 7 bands of the Landsat family of imaging sensors", i.e., a spectral resolution where bands visible blue (B), visible green (G), visible red (R), near infra-red (NIR), medium infra-red 1 (MIR1), medium infra-red 2 (MIR2) and thermal infra-red (TIR) overlap (which does not mean coincide) with Landsat's. According to these four families of spectral resolution specifications, the SIAM software product can pre-classify any radiometrically calibrated MS image acquired by past, existing or future-planned optical imaging sensors, either spaceborne or airborne, refer to Table 3. To realistically cope with the fact that there is no "fixed" number of quantization levels which is "optimal" in general, since this number is user- and application-specific (refer to this Section above), each SIAM subsystem delivers as output four pre-classification maps at different levels of color quantization, which are not alternative, but co-exist (like different hierarchical levels of detail co-exist in ontologies of the world [133], like LC class taxonomies [126], [127]): fine, intermediate, coarse and "shared", see Table 3. The latter provides a pre-defined vocabulary of color names "shared" by the four SIAM subsystems. Hence, this "shared" color vocabulary can be employed for inter-sensor post-classification change/no-change detection.

Since it is a physical model, the sole requirement of the prior knowledge-based SIAM color quantizer is to be input with MS data provided with a physical unit of radiometric measure, namely, digital numbers (DNs) radiometrically calibrated into TOARF or SURF values, in agreement with the QA4EO recommendations [1], refer to Section I. Noteworthy, TOARF $\supseteq$ SURF, i.e., SURF is a special case of TOARF in clear sky and flat terrain conditions [128]. In practice, (noisy) TOARF $\approx$ (noiseless) SURF + atmospheric noise + non-flat terrain effects. If SIAM is successful in mapping a MS data space of (noisy) TOARF values into fixed non-overlapping hypervolumes (discrete color names as mutually exclusive and totally exhaustive buffer zones or domains of activation), then (noiseless) SURF values fall around the center of these hypervolumes, see Fig. 7. Examples of the SIAM output products, to be employed in the further experimental Section V, are shown in Fig. 8, Fig. 9 and Fig. 10.

### *F. New Protocol for $Q^2A$ of MS Image PAN-sharpening Outcome and Process*

Summarized in Section II.E, preliminary considerations about existing PAN-sharpened MS image quality estimation procedures, surveyed in Section III, are taken into account to design and implement a novel (to the best of these author's knowledge, the first) procedure for $Q^2A$ of MS image PAN-sharpening outcome and process. The proposed estimation procedure belongs to the class of simplified one-statement Wald's protocols with reference image $MS_l$ at low spatial scale $l < h$, refer to Section III.B. In spite of this, proposed findings in QI selection and quality metrics can be extended to the second class of procedures for $Q^2A$ of PAN-sharpened MS images at high spatial scale h, without reference image, to replace the bivariate heterogeneous UIQI metric, Q, typically employed in the QNR formulation, refer to Section III.C.

The first step in the design of the novel procedure was the definition of an original taxonomy of PAN-sharpened MS image QIs and quality metrics, which are mapped onto four nominal scales, i.e., each QI is assigned with four categorical variables, see Table 4, alternative to traditional categorizations, like those proposed or surveyed in [18], [50], [53], [55], [95].

I. Homogeneous versus heterogeneous (claimed to be "universal") combinations of statistics, refer to Section

II.E. In general, the latter should be discouraged as a possible cause of non-redundant information loss, due to dimensionality reduction.

II. Univariate (one-channel) / bivariate / multivariate (multi-channel) analysis.

III. 1st-, 2nd- or 3rd-order statistics in the spatial domain, in line with the ETAU principle, refer to Section II.D.

IV. QI (or, vice versa, cost index) categories 1 to 4.

1. SPCTRL: Context-insensitive (pixel-based) Position (row and column)-independent Spectral cost indexes.
2. SPCTRL & SPTL1: Context-insensitive Position-dependent Spectral cost indexes.
3. SPCTRL & SPTL2: Context-sensitive Position-independent Spectral cost indexes.
4. SPCTRL & SPTL1 & SPTL2: Context-sensitive Position-dependent Spectral cost indexes.

For example, this fourfold QI taxonomy, I to IV, reveals that traditional multivariate QIs (or, vice versa, cost indexes), like Q4, ERGAS and average SAM, are all 1st-order (non-contextual) statistics in the spatial domain and all belong to the product QI category 2, SPCTRL & SPTL1 - Context-insensitive Position-dependent. Hence, from a statistical standpoint, due to their degree of similarity, they are expected to be correlated in the RS common practice.

In the rest of this section, first, product QIs are described and implemented, in agreement with Table 4. Some of these product QIs are extracted from the SIAM output products, generated automatically and in near real-time from the fused $MS^*_l$ and the reference $MS_l$ image. Second, process QIs are selected and instantiated. Finally, intra- and inter-category quantitative QI combination and ranking are discussed.

*1) First category of product QIs: SPCTRL - Context-insensitive (pixel-based) and Position (row and column)-independent (Rotation invariant)*

The first category of product QIs and quality metrics consists of traditional multidimensional (multi-band) absolute differences (e.g., implemented via the Minkowski distance of order 1 [61]) between univariate (one-channel) global (image-wide) gross (summary) characteristics of the 1st-order in the spatial domain, estimated from corresponding pairs of bands of the sensory and fused MS images at low spatial scale l, $MS_{l,b}$ and $MS^*_{l,b}$, b = 1, …, B. Given the univariate 1st-order (in the spatial domain, pixel-based) histogram (distribution), H(x), and its probability density function, p(x), with $\sum_{x \in GL} p(x)$ = 1, of a one-channel (univariate) image x (random variable), with scalar pixel values belonging to the set of gray levels GL, the implemented scalar summary statistics of H(x) are (see Table 4):

- Mean of H(x), 1st-degree moment, MeanUnvrt $= E[x] = \bar{x} \geq 0$. (12)
- Standard deviation of H(x), 2nd-degree moment about the mean, StDvUnvrt $= \sigma(x) = \sigma_x \geq 0$. (13)
- Skewness of H(x), 3rd-degree moment about the mean, SkwnsUnvrt $= \frac{1}{\sigma_x^3} E\left[(x - \bar{x})^3\right]$. (14)
- Kurtosis of H(x), fourth-degree moment about the mean, KrtsUnvrt $= \frac{1}{\sigma_x^4} E\left[(x - \bar{x})^4\right] \geq 0$. (15)
- Entropy of H(x) [33], EntrpyUnvrt $= -\sum_{x \in GL} p(x) \log_2(p(x)) \in [0, \log_2 GL]$, (16)
  where EntrpyUnvrt is maximum (in case of an equiprobable distribution) when Energy (EnrgyUnvrt) is minimum, with EnrgyUnvrt $= \sum_{x \in GL} p(x)^2 \in [0, 1]$, such that an alternative formulation of EntrpyUnvrt can be EntrpyUnvrt = (1 – EnrgyUnvrt) $\in$ [0, 1] [75]. In [75], in the domain of 2nd-order spatial statistics extracted from a GLCM, it was proved that the two highly correlated measures of image energy and entropy tend to be poorly correlated with features like image contrast and standard deviation, which are in turn highly correlated one another.

*2) Second category of product QIs: SPCTRL & SPTL1 - Context-insensitive and Position-dependent (Sensitive to Rotation)*

As mentioned in the introduction to Section IV.F, popular multivariate image QIs, like average SAM, ERGAS and Q4, belong to this second category of product QIs.

In our experiments, two features belonging to this category were implemented, one traditional and one innovative, see Table 4. The traditional feature is the bivariate PCC, computed pixel-based and band-specific between each pair of bands $MS_{l,b}$ and $MS^*_{l,b}$, b = 1, …, B. The inverse PCC parameter, InvrsCrltnBivrt = 1 – PCC, is a cost function, in range [0, 1], to be minimized for image quality improvement. Band-specific InvrsCrltnBivrt values are averaged across bands. In compliance with Section II.E, since PCC is sensitive to collinearities between the two random variables, the proposed evaluation procedure was planned to be validated with and without the contribution of PCC. The latter evaluation case was expected to be more in line with human photointerpretation results.

The innovative feature was the SIAM-based multivariate PostClChngDtctnMvrt statistic, mentioned in Table 4. The Q-SIAM prior knowledge-based color space quantizer was run automatically and in near real-time on the fused $MS^*_l$ and the sensory $MS_l$ image. Table 3 shows that the Q-SIAM subsystem delivers as output one pre-classification map at a so-called “shared” number of color levels. This “shared” color map vocabulary can be employed for automatic inter-sensor post-classification change/no-change detection, as shown in Fig. 10. The cumulative number of pixels featuring a change in the multivariate SIAM-based post-classification mapping provided the PostClChngDtctnMvrt statistic.

*3) Third category of product QIs: SPCTRL & SPTL2 - Context-sensitive Position-independent (Rotation invariant)*

Implemented features belonging to this category have no counterpart in state-of-the-art evaluation procedures with or without reference image, like those proposed in [18], [54], [63], [92].

*a) Original TIMS-GLCM calculator and TIMS-GLCM texture feature extractor*

To account for the ETAU principle presented in Section II.D and inspired by the third-order GLCM proposed in [76], a novel third-order isotropic multi-scale GLCM (TIMS-GLCM) was designed and implemented as an upper triangular three-dimensional array (where the typical symmetry of a GLCM is exploited to reduce memory size and computation time), transformed into a probability distribution, such that $\sum_{i=0}^{GL-1}\sum_{j=i+1}^{GL-1}\sum_{k=0}^{GL-1} p(i,j,k) = 1$, as shown in Fig. 11 and Fig. 12.

To reduce computation time (at the cost of a loss in sensitivity), pixel values were discretized into GL = 32 gray levels (histogram bins), in agreement with [76]. The maximum size S of the square moving window was fixed equal to 7 pixels, corresponding to $2.44 \times 7 \approx 18$ m in the $MS_l$ image, investigated in a three-scale TIMS-GLCM instance featuring ray r = displacement D = 1, 2, 3 in pixel unit, see Fig. 11.

The moving window was centered on each pixel of the fused and reference images, $MS^*_l$ and $MS_l$. One univariate TIMS-GLCM was instantiated per image channel, where 3-tuples collected from the pixel-centered moving window were cumulated. The mDMI set of univariate texture features extracted from a single-band TIMS-GLCM instance, transformed into a probability distribution, such that $\sum_{i=0}^{GL-1}\sum_{j=i+1}^{GL-1}\sum_{k=0}^{GL-1} p(i,j,k) = 1$, were contrast and energy, according to [75], and the large number emphasis (LNE), according to [76].

- Third-order second-degree Contrast $= \sum_{i=0}^{GL-1}\sum_{j=i+1}^{GL-1}\sum_{k=0}^{GL-1}\left[(i-j)^2+(j-k)^2+(i-k)^2\right]p(i,j,k)\cdot$ (17)
- Third-order second-degree Energy (Angular second moment) $= \sum_{i=0}^{GL-1}\sum_{j=i+1}^{GL-1}\sum_{k=0}^{GL-1} p^2(i,j,k)\cdot$ (18)
- Third-order second-degree Large Number Emphasis (LNE) $= \sum_{i=0}^{GL-1}\sum_{j=i+1}^{GL-1}\sum_{k=0}^{GL-1}\left(i^2+j^2+k^2\right)p(i,j,k)\cdot$ (19)

The absolute differences (Minkowski distances of order 1 [61]) between per band-specific 3rd-order texture statistics were computed between each pair of corresponding bands in the fused $MS^*_l$ and reference $MS_l$ images.

*b) Multivariate SIAM-based multi-level 8-adjacency cross-aura contour measure*

From each of the two multi-level SIAM maps, featuring fine, intermediate and coarse color discretization levels (refer to Table 3), automatically generated in near real-time from the test and reference images, $MS^*_l$ and $MS_l$, a three-level sum of 8-adjacency cross-aura measures was computed, so that the cross-aura value of each pixel ranges in interval {0, 24 = 8 × 3}, see Fig. 8 and Fig. 9. This SIAM-based three-level sum of 8-adjacency cross-aura measures provides a pixel-specific contour intensity, in range {0, 24}, increasing if the pixel is an isolated contour pixel (see Fig. 9) or if its color contrast is persistent through reductions of the color quantization levels (refer to Table 3). Next, the image-wide per-pixel average statistic was collected. The absolute difference (Minkowski difference of order 1 [61]) between these two image-wide multivariate statistics collected from images $MS^*_l$ and $MS_l$ was computed.

*4) Fourth category of product QIs: SPCTRL & SPTL1 & SPTL2 - Context-sensitive Position-dependent (Sensitive to rotation)*

Same as in Section IV.F.3, but the multi-level sum of cross-aura values was binarized into a binary contour pixel value: yes/no, coded as either 1 or 0, to avoid considering contour intensity values superior for isolated pixels. Next, an inter-image absolute difference was computed pixel-wise and, finally, averaged image-wise. In practice, this is an inter-image edge difference. This category of features has no counterpart in state-of-the-art evaluation procedures, with or without reference image, like those proposed in [18], [54], [63], [92].

*5) Process QIs*

In addition to the aforementioned product QIs, two process QIs, also called QIs of operativeness (QIOs) [8], [9], were considered for each MS image PAN-sharpening algorithm: the processing time and the number of system's free parameters, to be user-defined based on heuristics. The latter is monotonically decreasing with ease of use, i.e., it is inversely related to the system's degree of automation [8], [9].

*6) Quantitative QI combination and ranking*

To combine two (or more) quantitative variables whose units of measure, domains of change and/or sensitivity to changes in input data are different, there are two possible strategies: (i) quantitative variables are transformed into ranked variables, equivalent to categorical variables (affected by a quantization error), then ranks are combined, or (ii) their units of measure, domains of change and sensitivities are harmonized, so that the two harmonized quantitative variables can be combined before or after ranking (categorization). In line with past works [129], [130], in the present study heterogeneous quantitative variables were standardized for harmonization before combination. By definition [89], the standard score *z* of a raw population *x* is:

$$z = stndrd\,(x) = \frac{(x-\bar{x})}{\sigma(x)}\ , \qquad (20)$$

such that E[z] = 0 and σ(z) = 1. Since the standardized variable z represents the distance between the raw score and the population mean in units of the standard deviation, then z is negative when the raw score is below the population mean, positive when above. It is worth mentioning that the sum of T standardized variables is a variable with zero mean and variance equal to T.

In the present study, standardized variables were combined (summed) before ranking when they were belonging to the same product QI category 1 (SPCTRL) to 4 (SPCTRL & SPTL1 & SPTL2), refer to Section IV.F.1 to Section IV.F.4, as shown in Table 7. Otherwise, to account for inter- rather than intra-category differences, e.g., to combine statistical apples with statistical oranges, inter-category statistics were combined after ranking, see Table 9. The advantage of summing up quantitative standardized variables in place of ranked (categorical) variables is that the domain of change of the former is continuous rather than discrete, i.e., standardized variables are not affected by any discretization error.

To recapitulate, in any quantitative evaluation procedure defined beforehand, i.e., prior to looking at the dataset at hand, the arbitrary and application-specific choice of similarity (quality) or dissimilarity (distorsion) QIs and quality metrics does not allow to reach any "ultimate" conclusion about the "absolute" quality of outcomes or processes being assessed. In other words, any quantitative evaluation procedure provides nothing more than relative (subjective) conclusions about alternative solutions. Nonetheless, when a pool of QIs, individually equivalent to weak sources of evidence, is mDMI, then a multivariate convergence-of-evidence approach can be applied, to infer strong conjectures from weak sources of univariate evidence [58], refer to Section II.E.

## V. Results

In agreement with Section IV, the sensory image pair, $P_h$ and $MS_l$, was radiometrically calibrated into TOARF values (see Fig. 4) and the reference $MS_l$ image was automatically mapped by the Q-SIAM expert system for color quantization (see Table 3). The radiometrically calibrated downsampled $P_{h->l}$ and $MS_{l->s}$ image pair, with spatial scale s = l / 4 = 1 / (2.44 m × 4), such that the "artificial" spatial scale ratio (l : s) is equal to the "native" spatial scale ratio (h : l), where spatial scale h = 1 / 0.61 m, was generated through LPF filtering and filtered image downsampling in accordance with Section III.B. The fourteen alternative MS image PAN-sharpening algorithms selected for testing (refer to Table 1) were input with the radiometrically calibrated downsampled $P_{h->l}$ and $MS_{l->s}$ image pair to deliver as output fourteen alternative fused $MS^*_l$ images. Each fused $MS^*_l$ image was mapped automatically by the prior knowledge-based SIAM pre-classifier. In parallel, according to Section IV.D and Table 2, a perceptual visual assessment of the fourteen fused $MS^*_l$ images was conducted by sixteen human subjects, which led to the development of Table 5 as a reference baseline.

Summarized in Table 4, product QIs were collected from each fused $MS^*_l$ image in comparison with the reference $MS_l$ image, as shown in Table 6. According to Section IV.F.6, standardized QIs were combined (summed) per product QI category 1 (SPECL) to 4 (SPCTRL & SPTL1 & SPTL2) and ranked as shown in Table 7. The inter-category combination of per-category product ranks was accomplished in Table 9. This pool of per-category product ranks was combined with process ranks, shown in Table 8, as summarized in Table 10. To compare Table 10, delivered by the proposed evaluation procedure, with standard QIs (or cost indexes), the popular ERGAS, average SIAM and Q4 statistics were collected from each fused $MS^*_l$ image compared against the reference $MS_l$ image, as shown in Table 11. In particular, the final Q4 estimate was averaged over Q4 values estimated per image-block in a regular-grid image partition of BL × BL image blocks, with BL = 8, refer to Section IV.C. Table 12 summarizes final ranks of: (i) alternative fused $MS^*_l$ images, scored by visual inspection (refer to Table 5), (ii) products and products & processes, scored by the proposed evaluation procedure (refer to Table 9 and Table 10 respectively), and (iii) products, assessed in quality by traditional QIs (refer to Table 11).

Table 13 presents the Spearman's rank correlation coefficients (SRCCs) generated from pairwise comparisons of ranked variables generated in case of ERGAS, average SAM, Q4, Product Case C and Product & Process Case D, selected from Table 12. The SRCC index in range [-1, 1] is a nonparametric measure of statistical dependence between two ranked variables. It assesses how well the relationship between two ranked variables can be described by a monotonically increasing or decreasing function. If there are no repeated data values, a perfect Spearman correlation of +1 or −1 occurs when each of the variables is a monotonically increasing or decreasing function of the other, even if their relationship is not linear, which makes it quite different from the popular PCC.

## VI. Discussion

In this experiment, according to Table 5, the two qualitatively "best" PAN-sharpened MS images selected by human subjects were those generated by the PC2_B and HCS3_NN algorithm implementations.

In agreement with Section II.E, the well-known sensitivity of the bivariate PCC to collinearities between the two input random variables, which led to neglect the estimation of PCC as a viable texture feature from popular $2^{nd}$-order GLCMs [75], recommended the computation of a product rank Case C, where the InvrsCrltnBivrt cost term was omitted from the product QI category 2, SPCTRL & SPTL1 - Context-insensitive Position-dependent Spectral cost indexes, as a viable alternative to the product rank in Case A, see Table 7 and Table 9. The same omission of the InvrsCrltnBivrt cost term accounts for the product & process rank in Case D, alternative to the product & process rank in Case B, see Table 10. The conclusion is that, based on theoretical considerations in addition to experimental evidence summarized in Table 9 and Table 10, the removal of PCC from inter-image QIs is strongly recommended. In addition, in line with theoretical expectations, Table 9 and Table 10 show that the proposed convergence-of-evidence approach is robust to changes in one information source, like InvrsCrltnBivrt, in both product and product & process quality assessments.

The final summary Table 12 shows that, in this experiment, the first- and second-best choices of the novel quantitative estimation procedure comply with perceptual ranking by human subjects of fourteen alternative PAN-sharpened MS outcomes. Noteworthy, traditional multivariate QIs, like Q4, average SAM and ERGAS, completely fail detecting one-of-two best choices by human subjects, specifically, the outcome of the HCS3_NN algorithm implementation, see Fig. 13.

Although the goal of this experiment is validation by independent human subjects of an evaluation procedure, rather than selection of a "best" MS image PAN-sharpening algorithm, the conclusion that, according to the proposed product and product & process evaluation procedures, the implemented HCS3_NN algorithm [109] is first-best in both scores is of potential interest to those RS scientists and practitioners who, in the RS common practice, are recommended to comply with the QA4EO guidelines (refer to Section I).

Table 9 reveals that, in this experiments, the sole QI belonging to category 4, SPCTRL & SPTL1 & SPTL2 - Context-sensitive Position-dependent Spectral cost indexes, specifically, the BinaryCntourMvrt index (see Fig. 8 and Fig. 9), is the individual indicator that best approximates (which is highly correlated with) the Product final ranks, either Case A or Case C, although no single "universal" QI can exist on a theoretical basis (refer to Section II.B).

As observed in Section IV.F, since they belong to the same product QI category 2, SPCTRL & SPTL1 - Context-insensitive Position-dependent, traditional multivariate QIs (or cost indexes), like Q4, ERGAS and average SAM, are likely to be highly correlated (little informative, highly redundant) in many datasets. Traditionally, a correlation coefficient greater than 0.80 represents strong agreement, between 0.40 and 0.80 describes moderate agreement, and below 0.40 represents poor agreement [131]. This theoretical expectation about Q4, ERGAS and average SAM is clearly observable in works by other authors, like [18] and [63]. In line with theory, Table 13 shows that, also in our experiment, these traditional QIs feature high values of the SRCC, which means their pairwise relationships are nearly monotonically increasing or decreasing.

## VII. Conclusions

Provided with a relevant survey value, this paper reports on the design, implementation and validation by independent human subjects of a novel (to the best of these authors' knowledge, the first) procedure for perceptual visual quality assurance of MS image PAN-sharpening outcome and process. This is an inherently difficult (ill-posed) inductive learning-from-data problem, open to better-posed solutions in compliance with the QA4EO guidelines and the principles of human vision. Several conclusions of potential interest to a large segment of the RS and computer vision communities can be inferred from the high degree of convergence between theoretical considerations and experimental evidence highlighted in this paper.

The first conclusion is that, based on a critical analysis of existing literature, traditional PAN-sharpened MS image quality estimation procedures appear affected by conceptual and implementation drawbacks, which undermine their effectiveness in quality assurance.

(1) Belonging to the product QI category 2, SPCTRL & SPTL1 - Context-insensitive Position-dependent, "universal" multivariate Q/Q4/Q2$^n$ and SSIM indexes are implemented as heuristic mixtures, specifically, logical AND-combinations, of "heterogeneous" scalar QIs of signal fidelity, not to be confused with PVQMs. The same consideration holds for the fused image quality assessment without reference, QNR, where the "universal" Q metric is adopted. These AND-combinations of heterogeneous random variables featuring different statistical properties, domain of change and sensitivity to changes in input data, such as the AND-combination of pixel-based bivariate cross-correlation with differences in local univariate mean and standard deviation, are pursued in the erroneous attempt to find a "universal" summary (gross) statistic which cannot exist, due to the non-injective property of QIs, to be regarded as common knowledge by the scientific community. It means that for model comparison purposes, heuristic mixtures of heterogeneous QIs should be avoided before qualitative (categorical) ranking of each individual QI takes place. Otherwise, if any combination of heterogeneous QIs occurs before ranking, then each single QI in the combination should be standardized (z-scored) in advance to feature zero mean and unit variance, which accomplishes inter-QI harmonization of units of measure, domains of variation and sensitivities to changes in input data.

(2) Popular univariate (one-channel) QIs of signal fidelity [18], like relative bias in percent, relative difference of variances, relative standard deviation, etc., are all 1$^{st}$-order statistics in the spatial domain and all belong to the product QI category 1, SPCTRL - Context-insensitive Position-independent, exclusively. The bivariate PCC, which employs as input two spectral bands, together with popular multivariate QIs [18], like Q4, ERGAS and average SAM, are all 1$^{st}$-order statistics in the spatial domain (pixel-based, context-insensitive) and all belong to the MS image PAN-sharpening outcome QI category 2, SPCTRL & SPTL1 - Context-insensitive Position-dependent, exclusively. The conclusion is that existing MS image PAN-sharpening product evaluation procedures, e.g., refer to [18] and [63], employ no QI belonging to either category 3 (SPCTRL & SPTL2 - Context-sensitive Position-independent) or category 4 (SPCTRL & SPTL1 & SPTL2 - Context-sensitive Position-dependent). Consequences are twofold.

(i) Since they are pixel-based (context-insensitive), traditional MS image PAN-sharpening product quality estimation procedures consider a planar 2D dataset, i.e., a 2D array of vector data, equivalent to a 1D string of vector data. In practice, they adopt a pixel-based image analysis approach, which is a special case of 1D image analysis where the 2D spatial non-topological and topological properties of images are ignored. Image statistics are typically non-stationary; according to human pre-attentive vision, they should be investigated on a multi-scale basis (at least four spatial scales) up to third-order statistics in the 2D spatial domain [49], [71]. Actually, the recommended block-based implementation of Q/Q4/Q2$^n$, does not counterbalance this lack, at the cost of introducing yet-another system's free-parameter, the block size, to be user-defined based on heuristics, because Q/SSIM is a signal fidelity measure featuring both a statistical link and a formal connection with the conventional pixel-based mean squared error [137]. In image analysis, no "single" best spatial scale can exist [70], which means that no single-scale Q/Q4/Q2$^n$ can be robust to changes in block size. In convergence with these theoretical considerations, our experimental results provide a significant counter-example where the three popular Q4, ERGAS and average SAM indexes simultaneously fail detecting one-of-two best choices by human subjects, see Table 12.

(ii) Since they belong to the same MS image PAN-sharpening product QI category 2, SPCTRL & SPTL1 - Context-insensitive Position-dependent, popular QIs of signal fidelity, such as Q4, ERGAS and average SAM, are likely to be highly correlated (little informative, highly redundant) in common practice, in agreement with [137]. This theoretical expectation is clearly observable in works by other authors, like [18] and [63]. It is confirmed by the present work, see Table 12 and Table 13.

(3) The bivariate PCC statistic is widely adopted in existing estimation procedures, e.g., in the Q/Q4/Q2$^n$ indexes. The well-known sensitivity of PCC to linear transformations means that correlation is maximum between two images that are either identical or one the linear transformation of the other although, in this latter case, they can look (perceptually) very different. Its macroscopic inconsistency with PVQMs strongly discourages image (dis)similarity metrics, such as those surveyed in [55], from using the PCC as an input variable. Similar considerations led to neglect the estimation of correlation as a viable texture feature from popular GLCMs [75]. This recommendation is successfully put into practice in the present study, see Table 9 and Table 10.

The second conclusion is that four nominal taxonomies of image-pair QIs are proposed, alternative to existing categorizations, like the univariate/multivariate QI categorization adopted in [18], or various QI taxonomies discussed in [50], [55] and [95]. The four categorization criteria for fused image QIs proposed in this paper are summarized below.

I. Homogeneous versus heterogeneous (“universal”) statistics, like those proposed in [60].
II. Univariate (one-channel) or multivariate (multi-channel) statistics (where bivariate cross-correlation must be avoided).
III. 1$^{st}$-order (mean, stdev, skeweness, kurtosis, entropy) or 3$^{rd}$-order (contrast, energy, large number emphasis) statistics in the spatial domain.
IV. Inter-image comparison of type 1: SPCTRL = Context-insensitive (pixel-based) Position (row and column)-independent Spectral QI; 2: SPCTRL & SPTL1 = Context-insensitive Position-dependent Spectral QI; 3: SPCTRL & SPTL2 = Context-sensitive Position-independent Spectral QI; 4: SPCTRL & SPTL1 & SPTL2 = Context-sensitive Position-dependent Spectral QI.

These four nominal taxonomies of fused image QIs and quality metrics, I to IV, help users to select MS image PAN-sharpening product QIs that are minimally dependent and maximally informative (mDMI), in order to adopt a multivariate convergence-of-evidence approach [58]. For example, popular image-pair QIs, such as average SAM, ERGAS and Q/Q4/Q2$^n$, which are typically assessed together [18], [63], all belong to the aforementioned product QI category 2, SPCTRL & SPTL1, which means they together tend to be maximally redundant and minimally informative, refer to this Section above.

The third conclusion is that, in the proposed experiment, the implemented evaluation procedure, based on the analysis of multiple independent sources of converging evidence, successfully agrees with perceptual ranking by human subjects of fourteen alternative PAN-sharpened MS outcomes, whereas traditional product QIs, like Q4, SAM and ERGAS, completely fail detecting one-of-two best choices by human subjects, specifically, the outcome of the HCS3_NN algorithm’s implementation, see Table 12. Noteworthy, in our experiment, the sole MS image PAN-sharpening product QI belonging to category 4, SPCTRL & SPTL1 & SPTL2, is the individual indicator that best approximates (which is highly correlated with) the product final ranks A and C, see Table 9, although it does not mean a single "universal" quality indicator can exist (refer to Section II.B).

Despite the goal of this experiment is validation of an evaluation procedure by human subjects, rather than selection of a “best” MS image PAN-sharpening algorithm, the fourth conclusion is of potential interest to RS scientists and practitioners, required to comply with the QA4EO guidelines in their RS common practice: according to the proposed product and product & process evaluation procedures, the implemented HCS3_NN algorithm [109] is first-best in both scores, see Fig. 13.

The fifth conclusion is that, although the proposed procedure for Q$^2$A of PAN-sharpened MS outcome and process is a simplified one-statement instantiation of the Wald’s three-statement protocol, i.e., the proposed evaluation procedure adopts the raw MS$_l$ image as a reference benchmark, all the proposed 1$^{st}$- and 3$^{rd}$-order statistics in the spatial domain can be employed in Eq. (9) and Eq. (11) of the MS image PAN-sharpening “quality with no reference” index, QNR. It means that Eq. (8), where the “universal” univariate Q metric is typically adopted by QNR to calculate the dissimilarity between couples of bands, should be reformulated, refer to Section III.

The sixth conclusion is that, in compliance with the Yellott’s (E)TAU principle (refer to Section II.D), the proposed third-order isotropic multi-scale gray-level co-occurrence matrix (TIMS-GLCM) calculator and the TIMS-GLCM texture feature descriptor are expected to outperform the traditional 2$^{nd}$-order GLCM [75], still popular in the RS community and implemented in commercial RS image processing software toolboxes, like Exelis ENVI [101], Trimble eCognition [132] and ERDAS Imagine [100].

Last but not least, this study reveals another RS data application domain, specifically, Q$^2$A of PAN-sharpened MS images, where a prior knowledge-based MS data space quantization (partitioning) algorithm in operating mode, such as SIAM, is eligible for use. Documented applications of prior knowledge-based image mapping systems date back to the early 1980s [44], [45] and span from RS image enhancement (pre-processing) [113], [114] to RS image understanding [8], [9],

[12], [13], [14], [44], [45], [115], [116], [117], [118], refer to Section IV.E.

Planned future developments of this work will regard the validation of the proposed evaluation procedure for MS image PAN-sharpening outcome when input three-band test images, acquired by terrain-level cameras, depict natural landscapes, whose photointerpretation and perceptual quality assessment by independent human subjects is expected to be more intuitive and, therefore, reliable. These future developments are expected to agree with the PVQM proposed in [138] as a viable alternative to the signal fidelity measure SSIM by one of its authors. It is summarized as follows.

$$D(I_R, I_T) = \frac{1}{S}\sum_{s=0}^{S-1}\frac{1}{\sqrt{SF_s}}\left\|\hat{I}_{R,s} - \hat{I}_{T,s}\right\|^2, \qquad (21)$$

where $I_R$ and $I_T$ are the reference and the test image respectively, $\hat{I}_{R,s}$ and $\hat{I}_{T,s}$ denote vectors containing the transformed reference and distorted image data at scale s = 0, …, S-1, respectively, and where $SF_s$ is the number of spatial filters in the sub-band at scale s. In this equation a root mean squared error is computed for each scale, and then averaged over these scales giving larger weight to the lower frequency coefficients (which are fewer in number, due to subsampling). Per se, this scale-dependent weighting policy is not supported by any perceptual plausibility.

Alternative to the normalized Laplacian pyramid proposed in [138], an even-symmetric multi-scale filter bank proposed in [139] provides a near-orthogonal image decomposition and a zero-crossing (ZX) image-contour detection. For the sake of simplicity, in 1D signal processing, it is such that:

$$Rcnstrct(x) = f(x) \text{ o } G(x) + [f(x) \text{ o } \partial\text{^}2\ G/\partial x\text{^}2]/2, \qquad (22)$$

where G(x) is a 1D Gaussian low-pass filter and ∂^2 G/∂x^2 is the second-derivative of a 1D Gaussian function which mimics an even-symmetric spatial filter. Noteworthy, according to [140], the 2nd-order derivative $\partial^2/\partial n^2$ is a nonlinear operator, it neither commutes nor associates with the convolution [88]. Therefore, the filtered image ($\partial^2 G/\partial n^2$ o I) is different from the 2nd-order derivative $\partial^2/\partial n^2$ applied to the low-pass image adopted by both Canny [141] and Bertero, Torre and Poggio [46]. Hence, the inequality

$$( \partial^2 G/\partial n^2 \text{ o } I ) \neq \partial^2/\partial n^2 (G \text{ o } I) \qquad (23)$$

always holds true. As a consequence, in Eq. (22), term {f(x) o G(x)} ∈ [0, MaxGrayValue] and {[f(x) o ∂^2 G/∂x^2]/2} ∈ [-MaxGrayValue / 2, MaxGrayValue / 2]. According to these properties, the PVQM proposed in [139] is (attention: for the time being, this index ignores the multiple spatial scales):

$$D(I_R, I_T) = |f_R(x) \text{ o } G(x) - f_T(x) \text{ o } G(x)| + \left|\frac{\left[f_R(x) \text{ o } \frac{\partial^2 G}{\partial x^2}\right]}{2} - \frac{\left[f_T(x) \text{ o } \frac{\partial^2 G}{\partial x^2}\right]}{2}\right|, \qquad (24)$$

where $|f_R(x) \text{ o } G(x) - f_T(x) \text{ o } G(x)| \in [0, MaxGrayValue]$ and $|[f_R(x) \text{ o } \partial\text{^}2\ G/\partial x\text{^}2]/2 - [f_T(x) \text{ o } \partial\text{^}2\ G/\partial x\text{^}2]/2| \in [0, MaxGrayValue]$. In mathematical terms, this is a Minkowski distance with degree d equal to 1. If appropriate, d can be set equal 2 (to apply a Euclidean distance) or superior.

## Acknowledgments

To accomplish this work Andrea Baraldi was supported in part by the National Aeronautics and Space Administration under Grant No. NNX07AV19G issued through the Earth Science Division of the Science Mission Directorate. Francesca Despini and Sergio Teggi were funded by the Agenzia Spaziale Italiana (ASI), in the framework of the project "Analisi Sistema Iperspettrali per le Applicazioni Geofisiche Integrate - ASI-AGI" (n. I/016/11/0). The authors are grateful to the Modena and Reggio Emilia University staff and students who voluntarily participated in this work through the perceptual image quality assessment experiment. Andrea Baraldi thanks Prof. Raphael Capurro for his hospitality, patience, politeness and open-mindedness. He also thanks Prof. Christopher Justice, Chair of the Department of Geographical Sciences, University of Maryland, for his friendship and support. The authors also wish to thank the Editor-in-Chief, Associate Editor and reviewers for their competence, patience and willingness to help.

## References

[1] *A Quality Assurance Framework for Earth Observation, version 4.0*, Group on Earth Observation / Committee on Earth Observation Satellites (GEO/CEOS), 2010. [Online]. Available: http://qa4eo.org/docs/QA4EO_Principles_v4.0.pdf

[2] M. C. Hansen and T. R. Loveland, “A review of large area monitoring of land cover change using Landsat data,” *Remote Sens. of Env.*, vol. 122, pp. 66–74, 2012.

[3] A. Chen, G. G. Leptoukh, and S. J. Kempler, “Using KML and virtual globes to access and visualize heterogeneous datasets and explore their relationships along the A-Train tracks,” *IEEE J. Sel. Topics Appl. Earth Observ. Remote Sens*., vol. 3, no. 3, pp. 352-358, 2010.

[4] *The Global Earth Observation System of Systems (GEOSS) 10-Year Implementation Plan*, Group on Earth Observation (GEO), 2005. [Online]. Available: http://www.earthobservations.org/docs/10-Year%20Implementation%20Plan.pdf

[5] C. Shannon, “A mathematical theory of communication,” Bell System Technical Journal, vol. 27, pp. 379–423 and 623–656, 1948.

[6] I. Herrmann, A. Pimstein, A. Karnieli, Y. Cohen, V. Alchanatis, D.J. Bonfil, “LAI assessment of wheat and potato crops by VENμS and Sentinel-2 bands”, *Remote Sens. of Env*., vol. 115, pp. 2141–2151, 2011.

[7] R. Capurro and B. Hjørland, “The concept of information,” Annual *Review of Inf. Science and Tech*., vol. 37, pp. 343-411, 2003.

[8] A. Baraldi and L. Boschetti, “Operational automatic remote sensing image understanding systems: Beyond Geographic Object-Based and Object-Oriented Image Analysis (GEOBIA/GEOOIA) - Part 1: Introduction,” Remote Sensing, vol. 4, pp. 2694-2735, 2012.

[9] A. Baraldi and L. Boschetti, “Operational automatic remote sensing image understanding systems: Beyond Geographic Object-Based and Object-Oriented Image Analysis (GEOBIA/GEOOIA) - Part 2: Novel system architecture, information/knowledge representation, algorithm design and implementation,” *Remote Sens.*, vol. 4, pp. 2768-2817, 2012.

[10] *Land Product Validation (LPV)*, Committee on Earth Observation Satellites (CEOS) - Working Group on Calibration and Validation (WGCV), [Online]. Available: http://lpvs.gsfc.nasa.gov/. Accessed on March 20, 2015.

[11] G. Schaepman-Strub, M. E. Schaepman, T. H. Painter, S. Dangel, and J. V. Martonchik, “Reflectance quantities in optical remote sensing - Definitions and case studies,” *Remote Sens. Environ*., vol. 103, pp. 27–42, 2006.

[12] A. Baraldi, L. Boschetti, L., and M. Humber, “Probability sampling protocol for thematic and spatial quality assessments of classification maps generated from spaceborne/airborne very high resolution images,” *IEEE Trans. Geosci. Remote Sens*., vol. 52, no. 1, Part: 2, pp. 701-760, Jan. 2014.

[13] A. Baraldi, M. L. Humber, D. Tiede, and S. Lang, "Automatic pre-classification of Landsat image composites of the conterminous United States – Process and outcome quality indicators", *Remote Sens*., submitted for consideration for publication, Jan. 2015.

[14] A. Baraldi and M. Humber, "Quality assessment of pre-classification maps generated from spaceborne/airborne multi-spectral images by the Satellite Image Automatic Mapper™ and Atmospheric/Topographic Correction™-Spectral Classification software products: Part 1 – Theory," *IEEE J. Sel. Topics Appl. Earth Observ. Remote Sens*., vol. 8, no. 3, pp. 1307-1329, March 2015.

[15] V. Cherkassky and F. Mulier, *Learning from Data: Concepts, Theory, and Methods*. New York, NY: Wiley, 1998.

[16] S. Liang, *Quantitative Remote Sensing of Land Surfaces*. Hoboken, NJ, USA: John Wiley and Sons, 2004.

[17] N. Longbotham, F. Pacifici, T. Glenn, A. Zare, M. Volpi, D. Tuia, E. Christophe, J. Michel, J. Inglada, J. Chanussot, and Qian Du, "Multi-Modal Change Detection, Application to the Detection of Flooded Areas: Outcome of the 2009–2010 Data Fusion Contest," *IEEE J. Sel. Topics Appl. Earth Observ. Remote Sens*., vol. 5, no. 1, pp. 331-342, 2012.

[18] L. Alparone, L. Wald, J. Chanussot, C. Thomas, P. Gamba, et al., "Comparison of Pansharpening Algorithms: Outcome of the 2006 GRS-S Data Fusion Contest," *IEEE Trans. Geosci. Remote Sens*., vol. 45, no. 10, pp. 3012-3021, 2007.

[19] J. Li, "Spatial quality evaluation of fusion of different resolution images," *ISPRS Int. Arch. Photogramm. Remote Sens*., vol. 33, no. B2-2, pp. 339–346, 2000.

[20] C. Thomas and L. Wald, "Assessment of the quality of fused products," in *Proc. 24th EARSeL Annu. Symp. New Strategies Eur. Remote Sens*., Dubrovnik, Croatia, May 25–27, 2004. M. Oluic, Ed., Rotterdam, The Netherlands: Balkema, 2005, pp. 317–325.

[21] Z. Wang, D. Ziou, C. Armenakis, D. Li and Q. Li, "A comparative analysis of image fusion methods," *IEEE Trans. Geosci.Remote Sens*., vol. 43, issue 6, pp. 1391-1402, 2005.

[22] L. Wald, "Some terms of reference in data fusion," *IEEE Trans. Geosci. Remote Sens*., vol. 37, no. 3, pp. 1190–1193, May 1999.

[23] B. Aiazzi, L. Alparone, S. Baronti, A. Garzelli, F. Nencini, and M. Selva, "Spectral information extraction by means of MS+PAN fusion," *ESA-EUSC*, 2004.

[24] D. A. Quattrochi and M. F. Goodchild, Eds., *Scale in Remote Sensing and GIS*. CRC Press, 1997.

[25] A. Dadon, E. Ben-Dor, M. Beyth, and A. Karnieli, "Examination of spaceborne imaging spectroscopy data utility for stratigraphic and lithologic mapping," *J. Appl. Remote Sens*., vol. 5, no. 1, pp.-053507, 2011.

[26] J. G. Liu, "Smoothing filter-based intensity modulation: A spectral preserve image fusion technique for improving spatial details," *Int. J. Remote Sens*., vol. 21, no. 18, pp. 3461–3472, 2000.

[27] B. Garguet-Duport, J. Girel, J.-M. Chassery, and G. Pautou, "The use of multi-resolution analysis and wavelets transform for merging SPOT panchromatic and multi-spectral image data," *Photogramm. Eng. Remote Sens*., vol. 62, no. 9, pp. 1057–1066, 1996.

[28] J. Hadamard, "Sur les problemes aux derivees partielles et leur signification physique," *Princet. Univ. Bull*., vol. 13, pp. 49–52, 1902.

[29] M. Sonka, V. Hlavac, and R. Boyle, *Image Processing, Analysis and Machine Vision*. London, U.K.: Chapman & Hall, 1994.

[30] Q. Iqbal and J. K. Aggarwal, "Image retrieval via isotropic and anisotropic mappings," in *Proc. IAPR Workshop Pattern Recognit. Inf. Syst*., Setubal, Portugal, Jul. 2001, pp. 34–49.

[31] G. A. Miller, "The cognitive revolution: a historical perspective", in *Trends in Cognitive Sciences 7*, pp. 141-144, 2003.

[32] F. J. Varela, E. Thompson, and E. Rosch, *The Embodied Mind: Cognitive Science and Human Experience*. Cambridge, Mass.: MIT Press, 1991.

[33] C. M. Bishop, *Neural Networks for Pattern Recognition*. Oxford, U.K.: Clarendon, 1995.

[34] G. Patanè and M. Russo, "The enhanced-LBG algorithm," *Neural Networks*, vol. 14, no. 9, pp. 1219–1237, 2001.

[35] G. Patanè and M. Russo, "Fully automatic clustering system," *IEEE Trans. Neural Networks*, vol. 13, no. 6, pp. 1285-1298, 2002.

[36] B. Fritzke, "The LBG-U method for vector quantization—An improvement over LBG inspired from neural networks," *Neural Processing Lett.*, vol. 5, no. 1, 1997.

[37] Y. Linde, A. Buzo, and R. M. Gray, "An algorithm for vector quantizer design," *IEEE Trans. Commun.*, vol. 28, pp. 84–94, Jan. 1980.

[38] D. Lee, S. Baek, and K. Sung, "Modified k-means algorithm for vector quantizer design," *IEEE Signal Processing Lett.*, vol. 4, pp. 2–4, Jan. 1997.

[39] B. Fritzke, *Some competitive learning methods*. Draft document, 1997. [Online]. Available: http://www.demogng.de/JavaPaper/t.html

[40] A. Baraldi and E. Alpaydin, ``Constructive feedforward ART clustering networks - Part I," *IEEE Trans. Neural Networks*, vol. 13, no. 3, pp. 645-661, March 2002.

[41] A. Baraldi and E. Alpaydin, ``Constructive feedforward ART clustering networks - Part II," *IEEE Trans. Neural Networks*, vol. 13, no. 3, pp. 662 -677, March 2002.

[42] T. Martinetz, G. Berkovich, and K. Schulten, "Topology representing networks," *Neural Networks*, vol. 7, no. 3, pp. 507–522, 1994.

[43] R. Laurini and D. Thompson, *Fundamentals of Spatial Information Systems*. London, UK: Academic Press, 1992.

[44] T. Matsuyama and V. Shang-Shouq Hwang, *SIGMA – A Knowledge-based Aerial Image Understanding System*, Plenum Press, 1990.

[45] M. Nagao and T. Matsuyama, *A Structural Analysis of Complex Aerial Photographs*. New York, NY, USA: Plenum, 1980.

[46] M. Bertero, T. Poggio, and V. Torre, "Ill-posed problems in early vision," *Proc. IEEE*, vol. 76, pp. 869–889, 1988.

[47] S. P. Vecera and M. J. Farah, "Is visual image segmentation a bottom-up or an interactive process?," *Perception and Psychophysics*, vol. 59, pp. 1280–1296, 1997.

[48] L. Wald, T. Ranchin, and M. Mangolini, "Fusion of satellite images of different spatial resolutions: Assessing the quality of resulting images", *Photogramm. Eng. Remote Sens*, vol. 63, no. 6, pp. 691-699, 1997.

[49] Jain A. and G. Healey, "A multiscale representation including opponent color features for texture recognition," *IEEE Trans. Image Processing*, vol. 7, no. 1, pp. 124-128, Jan. 1998.

[50] J. Pa and A. V. Hegdeb, "A Review of quality metrics for fused image," *Aquatic Procedia*, vol. 4, pp. 133 – 142, 2015.

[51] Y. Zhang and R. K. Mishra, "A review and comparison of commercially available pan-sharpening techniques for high resolution satellite image fusion", in *IEEE Geosci. and Remote Sens. Symposium* (IGARSS) 2012, 22-27 July 2012, pp. 182-185.

[52] L. Alparone, B. Aiazzi, S. Baronti, A. Garzelli, and F. Nencini, "A new method for MS+Pan image fusion assessment without reference," *IEEE*, 2006.

[53] M. C. El-Mezouar, N. Taleb, K. Kpalma, and J. Ronsin, "A new evaluation protocol for image pan-sharpening methods," *ICCIT 2012*, pp. 144-148.

[54] L. Alparone, B. Aiazzi, S. Baronti, A. Garzelli, F. Nencini, and M. Selva, "Multispectral and panchromatic data fusion assessment without reference," *Photogramm. Eng. Remote Sens*., vol. 74, no. 2, pp. 193-200, 2008.

[55] Weisi Lin, C.-C. Jay Kuo, "Perceptual visual quality metrics: A survey," *J. Vis. Commun. Image R*., vol. 22, pp. 297–312, 2011.

[56] A. Baraldi, L. Bruzzone, and P. Blonda, "Quality assessment of classification and cluster maps without ground truth knowledge," *IEEE Trans. Geosci. Remote Sens.*, vol. 43, no. 4, pp. 857-873, Apr. 2005.

[57] D. Zhang and G. Lu, "Review of shape representation and description techniques," *Pattern Recognition*, vol. 37, no. 1, pp. 1–19, 2004.

[58] A. Baraldi and J. V. B. Soares, "Software library of two-dimensional shape descriptors in object-based image analysis," *IEEE Trans. Image Proc*., submitted, 2015.

[59] *World Happiness Report, 2012/2013*, Columbia University, Canadian Institute for Advanced Research, London School of Economics, 2014.

[60] Z. Wang and A. C. Bovik, "A universal image quality index," *IEEE Signal Proc. Letters*, vol. 9, no. 3, pp. 81-84, March 2002.

[61] Z. Wang, A. C. Bovik, H. R. Sheikh, and E. P. Simoncelli, "Image quality assessment: From error visibility to structural similarity," *IEEE Trans. Image Proc*., vol. 13, no. 4, p. 1-14, 2004.

[62] L. Alparone, S. Baronti, A. Garzelli, and F. Nencini, "A global quality measurement of pan-sharpened multispectral imagery," *IEEE Geosci. Remote Sens. Letters*, vol. 1, no. 4, pp. 313-317, Oct. 2004.

[63] G. Vivone, L. Alparone, J. Chanussot, M. Dalla Mura, A. Garzelli, G. A. Liardi, R., Restaino, and L. Wald, "A critical comparison among pansharpening algorithms," *IEEE Trans. Geosci. Remote Sens*., vol. 53, no. 5, pp. 2565 – 2586, 2015.

[64] A. Garzelli and F. Nencini, "Hypercomplex quality assessment of multi/hyper-spectral images," *IEEE Geosci. Remote Sens. Lett*., vol. 6, no. 4, pp. 662–665, Oct. 2009.

[65] L. A. Zadeh, "Fuzzy sets," *Inform. Control*, vol. 8, pp. 338–353, 1965.

[66] B. Kosko, *Fuzzy Thinking*. Flamingo, London, UK, 1994.

[67] C. Mason and E. R. Kandel, “Central visual pathways,” in *Principles of Neural Science*, E. Kandel and J. Schwartz, Eds. Norwalk, CT, USA: Appleton and Lange, 1991, pp. 420–439.

[68] P. Gouras, “Color vision,” in *Principles of Neural Science*, E. Kandel and J. Schwartz, Eds. Norwalk, CT, USA: Appleton and Lange, 1991, pp. 467–479.

[69] E. R. Kandel, “Perception of motion, depth and form,” in *Principles of Neural Science*, E. Kandel and J. Schwartz, Eds. Norwalk, CT, USA: Appleton and Lange, 1991, pp. 441–466.

[70] H. R. Wilson and J. R. Bergen, “A four mechanism model for threshold spatial vision,” *Vis. Res*., vol. 19, no. 1, pp. 19–32, 1979.

[71] J. I. Yellott, “Implications of triple correlation uniqueness for texture statistics and the Julesz conjecture,” *Opt. Soc. Am.*, vol. 10, no. 5, pp. 777-793, May 1993.

[72] B. Julesz, E. N. Gilbert, L. A. Shepp, and H. L. Frisch, “Inability of humans to discriminate between visual textures that agree in second-order statistics – revisited,” *Perception*, vol. 2, pp. 391-405, 1973.

[73] B. Julesz, “Texton gradients: The texton theory revisited,” in *Biomedical and Life Sciences Collection*, Springer, Berlin / Heidelberg, vol. 54, no. 4-5, Aug. 1986.

[74] J. Victor, “Images, statistics, and textures: Implications of triple correlation uniqueness for texture statistics and the Julesz conjecture: Comment,” *J. Opt. Soc. Am. A*, vol. 11, no. 5, pp. 1680-1684, May 1994.

[75] A. Baraldi and F. Parmiggiani, ``An investigation of textural characteristics associated with gray level cooccurence matrix statistical parameters,'' *IEEE Trans. Geosci. Remote Sens.*, vol. 33, no. 2, pp. 293-304, March 1995.

[76] H. Anys and D. C. He, “Evaluation of textural and multipolarization radar features for crop classification,” *IEEE Trans. Geosci. Remote Sens*., vol. 33, no. 5, pp. 1170-1181, 1995.

[77] D. Marr, *Vision.* New York, NY, USA: Freeman and C., 1982.

[78] R. M. Boynton, ‘‘Human color perception,’’ in *Science of Vision*, K. N. Leibovic, Ed., pp. 211–253, Springer-Verlag, New York, 1990.

[79] J. Rodrigues and J.M. Hans du Buf, “Multi-scale lines and edges in V1 and beyond: Brightness, object categorization and recognition, and consciousness,” *BioSystems*, vol. 1, pp. 1-21, 2008.

[80] J. Rodrigues and J.M.H. du Buf, “Multi-scale keypoints in V1 and beyond: Object segregation, scale selection, saliency maps and face detection”, *BioSystems*, vol. 86, pp. 75–90, 2006.

[81] F. Heitger, L. Rosenthaler, R. von der Heydt, E. Peterhans, and O. Kubler, “Simulation of neural contour mechanisms: from simple to end-stopped cells,” *Vision* Res., vol. 32, no. 5, pp. 963–981, 1992.

[82] L. Pessoa, “Mach Bands: How Many Models are Possible? Recent Experimental Findings and Modeling Attempts”, *Vision Res*., Vol. 36, No. 19, pp. 3205–3227, 1996.

[83] H. du Buf and J. Rodrigues, Image morphology: from perception to rendering, in *IMAGE - Computational Visualistics and Picture Morphology*, 2007.

[84] D. C. Burr and M. C. Morrone, ‘‘A nonlinear model of feature detection,’’ in *Nonlinear Vision: Determination of Neural Receptive Fields, Functions, and Networks*, R. B. Pinter and N. Bahram, Eds., pp. 309–327, CRC Press, Boca Raton, FL, 1992.

[85] E. H. Adelson and J. R. Bergen, “Spatio-temporal energy models for the perception of motion,” *J. Opt. Soc. Am. A*, vol. 2, pp. 284-299, 1985.

[86] D. Lowe, "Distinctive image features from scale-invariant keypoints," *Int. J. Comp. Vision*, vol. 60, no. 2, pp. 91–110, 2004.

[87] A. Baraldi and F. Parmiggiani, ``Combined detection of intensity and chromatic contours in color images,'' *Optical Engineering*, vol. 35, no. 5, pp. 1413-1439, May 1996.

[88] M. Hagenlocher, S. Lang, D. Hölbling, D. Tiede, and S. Kienberger, “Modeling hotspots of climate change in the Sahel using object-based regionalization of multidimensional gridded datasets,” *IEEE J. Sel. Topics Appl. Earth Observ. Remote Sens*., vol. 7, no. 1, pp. 229-234, Jan. 2014.

[89] E. Kreyszig, *Applied Mathematics*.Wiley Press, 1979.

[90] B. Aiazzi, L. Alparone, S. Baronti, A. Garzelli, and M. Selva, “MTF-tailored multiscale fusion of high-resolution MS and Pan imagery,” *Photogramm. Eng. Remote Sens*., vol. 72, no. 5, pp. 591–596, May 2006.

[91] F. Laporterie-Déjean, H. de Boissezon, G. Flouzat, and M.-J. Lefèvre-Fonollosa, “Thematic and statistical evaluations of five panchromatic/multispectral fusion methods on simulated PLEIADES-HR images,” *Inf. Fusion*, vol. 6, no. 3, pp. 193–212, Sep. 2005.

[92] B. Aiazzi, L. Alparone, S. Baronti, A. Garzelli, and M. Selva, “Twenty-five years of pansharpening: A critical review and new developments,” in *Signal and Image Processing for Remote Sensing*, 2nd ed., C.-H. Chen, Ed. Boca Raton, FL, USA: CRC Press, 2012, pp. 533–548.

[93] L. Wald, *Data Fusion. Definitions and Architectures - Fusion of Images of Different Spatial Resolutions*. Paris, France: Les Presses, Ecole des Mines de Paris, 2002.

[94] R. H. Yuhas, A. F. H. Goetz, and J. W. Boardman, “Discrimination among semi-arid landscape endmembers using the Spectral AngleMapper (SAM) algorithm,” in *Proc. Summaries 3rd Annu. JPL Airborne Geosci. Workshop*, 1992, pp. 147–149.

[95] C. Thomas and L. Wald, “Comparing distances for quality assessment of fused products,” in *Proc. 26th EARSeL Annu. Symp. New Develop. Challenges Remote Sens*., Warsaw, Poland, May 29–31, 2006. Z. Bochenek, Ed., Rotterdam, The Netherlands: Balkema, 2007, pp. 101–111.

[96] P.S. Pradhan, R.L. King, N.H. Younan, and D.W. Holcomb, “Estimation of the number of decomposition levels for a wavelet-based multiresolution multisensor image fusion ,” *IEEE Trans. Geosci. Remote Sens*., vol. 44, pp. 3674 – 3683, May 2006.

[97] T. Ranchin and L. Wald, “Fusion of high spatial and spectral resolution images: The ARSIS concept and its implementation”, *Photogramm. Eng. Remote Sens*., vol. 66, no. 1, pp. 49-61, 2000.

[98] P. Lukowicz, E. Heinz, L. Prechelt, and W. Tichy, ``Experimental evaluation in computer science: a quantitative study,'' *Tech. Rep*. 17/94, Univ. Karlsruhe (Germany), 1994.

[99] L. Prechelt, ``A quantitative study of experimental evaluations of neural network learning algorithms: Current research practice,'' *Neural Networks*, vol. 9, 1996.

[100] *ERDAS IMAGINE 2015 User's Guide*, Hexagon Geospatial. [Online] Available: http://www.hexagongeospatial.com/products/remote-sensing/erdas-imagine.

[101] *ENVI EX User Guide 5.0*, ITT Visual Information Solutions, Dec. 2009. [Online]. Available: http://www.exelisvis.com/portals/0/pdfs/enviex/ENVI_EX_User_Guide.pdf

[102] J. Zhou , D. L. Civco, and J. A. Silander, “A wavelet transform method to merge Landsat TM and SPOT panchromatic data,” *Int. J. Remote Sens*., vol. 19, no. 4, pp. 743–757, 1998.

[103] C A. Laben and B. V. Brower, “Process for Enhancing the Spatial Resolution of Multispectral Imagery Using Pan-Sharpening”, *US Patent* 6.011.875, 1998.

[104] P. A. Brivio, G. Lechi, and E. Zilioli, *Principi e metodi di Telerilevamento*. Cittàstudi Edizioni, 2006.

[105] K. Amolins, Y. Zhang, and P. Dare. “Wavelet based image fusion techniques - An introduction, review and comparison,” *ISPRS J. Photogram. & Remote Sens.*, vol. 62, pp. 249–263, 2007.

[106] M. J. Canty, *Image Analysis, Classification and Change Detection in Remote Sensing: With Algorithms for ENVI/IDL and Python*. Crc Press, 2014.

[107] B. Aiazzi, L. Alparone, S. Baronti, and A. Garzelli, “Context-driven fusion of high spatial and spectral resolution images based on oversampled multi-resolution analysis,” *IEEE Trans. Geosci. Remote Sens*., vol. 40, no. 10, pp. 2300–2312, Oct. 2002.

[108] J. Núñez, X. Otazu, O. Fors, A. Prades, V. Palà, and R. Arbiol, (1999). Multiresolution- based image fusion with additive wavelet decomposition, *IEEE Trans. Geosci. Remote Sens*., vol. 37, no. 3, pp. 1204–1211, May 1999.

[109] C. Padwick, M. Deskevich, F. Pacifici, and S. Smallwood, “WorldView-2 pan-sharpening,” in *American Society for Photogrammetry and Remote Sensing*, 2010.

[110] M. Ehlers, “Spectral characteristics preserving image fusion based on Fourier domain filtering,” in *Proc. of SPIE*, Maspalomas, Spain, 2004, vol. 5574, pp. 1–13.

[111] P.S. Chavez, S.C. Sides, J.A. Anderson, “Comparison of three different methods to merge multiresolution and multispectral data: Landsat TM and SPOT Panchromatic,” *Photogram. Eng. Remote Sens*., vol. 57, no. 3, pp. 295-303, 1991.

[112] *The 5th International Conference on Computing for Geospatial Research and Application* (COM.Geo 2014), Call For Presentation. [Online]. Available: http://www.com-geo.org/conferences/2014/topics.htm. Accessed on March 20, 2015.

[113] A. Baraldi, M. Humber and L. Boschetti, "Quality assessment of pre-classification maps generated from spaceborne/airborne multi-spectral images by the Satellite Image Automatic Mapper™ and Atmospheric/Topographic Correction™-Spectral Classification software products: Part 2 – Experimental results," *Remote Sens*., vol. 5, pp. 5209-5264, Oct. 2013.

[114] A. Baraldi, M. Gironda, and D. Simonetti, “Operational three-stage stratified topographic correction of spaceborne multi-spectral imagery employing an automatic spectral rule-based decision-tree preliminary classifier,” *IEEE Trans. Geosci. Remote Sens*., vol. 48, no. 1, pp. 112-146, Jan. 2010.

[115] S. A. Ackerman, K. I. Strabala, W. P. Menzel, R. A. Frey, C. C. Moeller, and L. E. Gumley, "Discriminating clear sky from clouds with MODIS", *J. Geophys. Res*., vol. 103, D24, no. 32, pp. 141-157, 1998.

[116] Y. Luo, A. P. Trishchenko and K. V. Khlopenkov, “Developing clear-sky, cloud and cloud shadow mask for producing clear-sky composites at 250-meter spatial resolution for the seven MODIS land bands over Canada and North America,” *Remote Sens. of Env*., vol. 112, pp. 4167–4185, 2008.

[117] Zhe Zhu and C. E. Woodcock, “Object-based cloud and cloud shadow detection in Landsat imagery”, *Remote Sens. of Env*., vol. 118, pp. 83–94, 2012

[118] D. Simonetti, E. Simonetti, Z. Szantoi, A. Lupi and H. D. Eva, “First results from the phenology-based synthesis classifier using Landsat 8 imagery,” *IEEE Geosci. Remote Sens. Letters*, accepted for publication, Feb. 2015.

[119] J. van de Weijer, C. Schmid, J. Verbeek, D. Larlus, Learning color names for real-world applications, *IEEE Trans. Image Proc*., vol. 18, no. 7, pp. 1512 – 1523, 2009.

[120] R. Benavente, M. Vanrell, and R. Baldrich, “Parametric fuzzy sets for automatic color naming,“ *J. Opt. Soc. Am. A*, vol. 25, pp. 2582-2593, 2008.

[121] T. Gevers, A. Gijsenij, J. van de Weijer, J. M. Geusebroek, *Color in Computer Vision*. Hoboken, NJ, USA: Wiley, 2012

[122] E. C. Tsao, J. C. Bezdek, and N. R. Pal, “Fuzzy Kohonen clustering network,” *Pattern Recognition*, vol. 27, no. 5, pp. 757–764, 1994.

[123] E. Erwin, K. Obermayer, and K. Schulten, “Self-organizing maps: Ordering, convergence properties and energy functions,” *Biol. Cybern*., vol. 67, pp. 47–55, 1992.

[124] S. P. Luttrell, “A Bayesian analysis of self-organizing maps,” *Neural Comput*., vol. 6, pp. 767–794, 1994.

[125] A. Baraldi, P. Puzzolo, P. Blonda, L. Bruzzone, and C. Tarantino, “Automatic spectral rule-based preliminary mapping of calibrated Landsat TM and ETM+ images,” *IEEE Trans. Geosci. Remote Sens*., vol. 44, pp. 2563-2586, 2006.

[126] *OWL Web Ontology Language*, World Wide Web Consortium (W3C). [Online] Available: http://www.w3.org/TR/owl-features/

[127] H. Couclelis, “Ontologies of geographic information,” *Int. J. Geo. Info. Science*, vol. 24, no. 12, pp. 1785-1809, 2010.

[128] P. S. Chavez, “An improved dark-object subtraction technique for atmospheric scattering correction of multispectral data,” *Remote Sens. Environ*., vol. 24, pp. 459–479, 1988.

[129] A. Baraldi, L. Bruzzone, P. Blonda, and L. Carlin, “Badly-posed classification of remotely sensed images - An experimental comparison of existing data labeling systems,” *IEEE Trans. Geosci. Remote Sens*., vol. 44, no. 1, pp. 214-235, Jan. 2006.

[130] A. Baraldi, L. Bruzzone and P. Blonda, "A multi-scale Expectation-Maximization semisupervised classifier suitable for badly-posed image classification," *IEEE Trans. Image Proc.*, vol. 15, no. 8, pp. 2208-2225, Aug. 2006.

[131] R. G. Congalton and K. Green, *Assessing the Accuracy of Remotely Sensed Data*, Boca Raton, FL: Lewis Publishers,1999.

[132] *eCognition® Developer 9.0 Reference Book*, Trimble, 2015.

[133] H. Couclelis, “Ontologies of geographic information,” *Int. J. Geo. Info. Science*, vol. 24, no. 12, pp. 1785-1809, 2010.

[134] G. Ginesu, F. Massidda, and D. Giusto, “A multi-factors approach for image quality assessment based on a human visual system model,” Signal Processing: Image Communication, vol. 21, pp. 316–333, 2006.

[135] C. Chubb and J. I. Yellott, “Every discrete, finite image is uniquely determined by its dipole histogram,” Vision Research, vol. 40, pp. 485–492, 2000.

[136] Z. Wang, E. P. Simoncelli, and A. C. Bovik, “Multi-scale structural similarity for image quality assessment,” in Proc. IEEE Asilomar Conf. on Signals, Systems, and Computers, pp. 1398–1402, 2003.

[137] R. Dosselmann and Xue Dong Yang, “A comprehensive assessment of the structural similarity index,” SIViP 2011, vol. 5, pp. 81–91, 2011.

[138] V. Laparra, J. Ballé, A. Berardino, and E. P. Simoncelli, Perceptual image quality assessment using a normalized Laplacian pyramid, Proc. IS&T Int'l Symposium on Electronic Imaging, Conf. on Human Vision and Electronic Imaging, vol. 2016(16), Feb 2016.

[139] A. Baraldi, Operational automatic stratified multi-scale image-contour, keypoint, texel and texture-boundary detection in panchromatic and color images: Developments and open challenges, *Preliminary report No. 1, version 10.1*, Baraldi Consultancy in Remote Sensing of Andrea Baraldi, August 2016.

[140] V. Torre and T. Poggio, “On edge detection,” *IEEE Trans. Pattern Anal. and Mach. Intell*. vol. 8, no. 2, pp. 147–163, 1986.

[141] J. Canny, ‘‘A computational approach to edge detection,’’ *IEEE Trans. Pattern Anal. and Mach. Intell*. vol. 8, no. 6, pp. 679–698, 1986.

FIGURE CAPTIONS

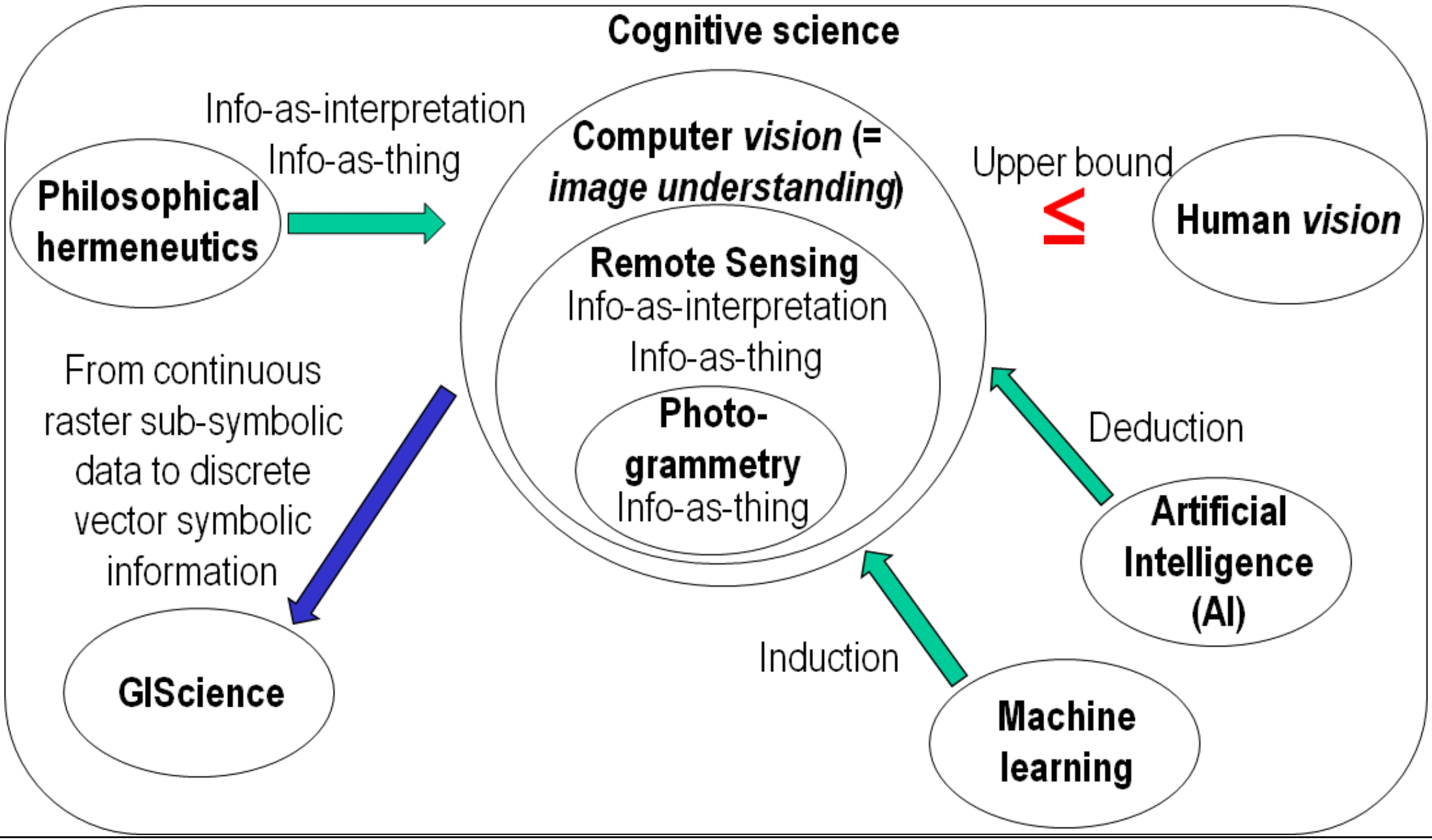

Fig. 1. Like engineering, remote sensing (RS) is a metascience, whose goal is to transform knowledge of the world, provided by other scientific disciplines, into useful user- and context-dependent solutions in the world [133]. Cognitive science is the interdisciplinary scientific study of the mind and its processes. It examines what cognition (learning) is, what it does and how it works. It especially focuses on how information/knowledge is represented, acquired, processed and transferred within nervous systems (humans or other animals) and machines (e.g., computers) [31], [32].

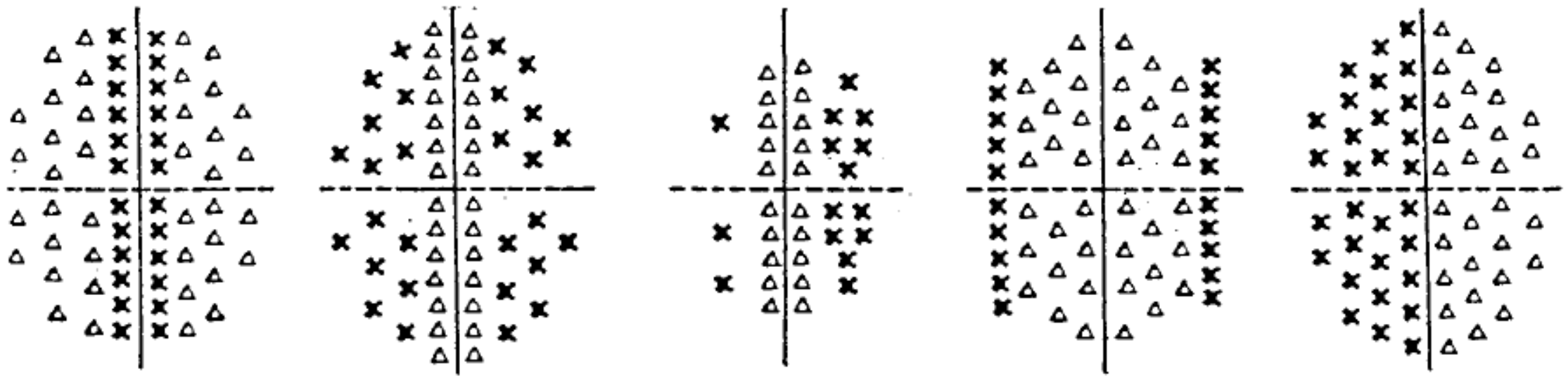

Fig. 2. Excitatory and inhibitory terms, shown as triangles ad crosses, in activation domains of even- and odd-symmetric S-cells found in the PVC of mammals by Mason and Kandel in their seminal work on neuroscience [67].

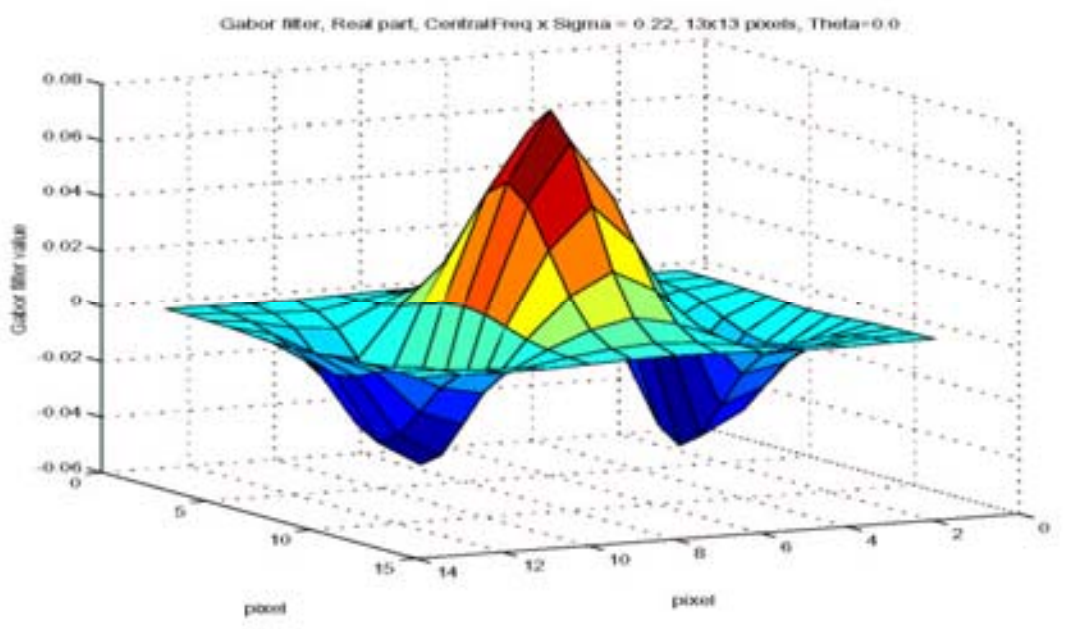

(a)

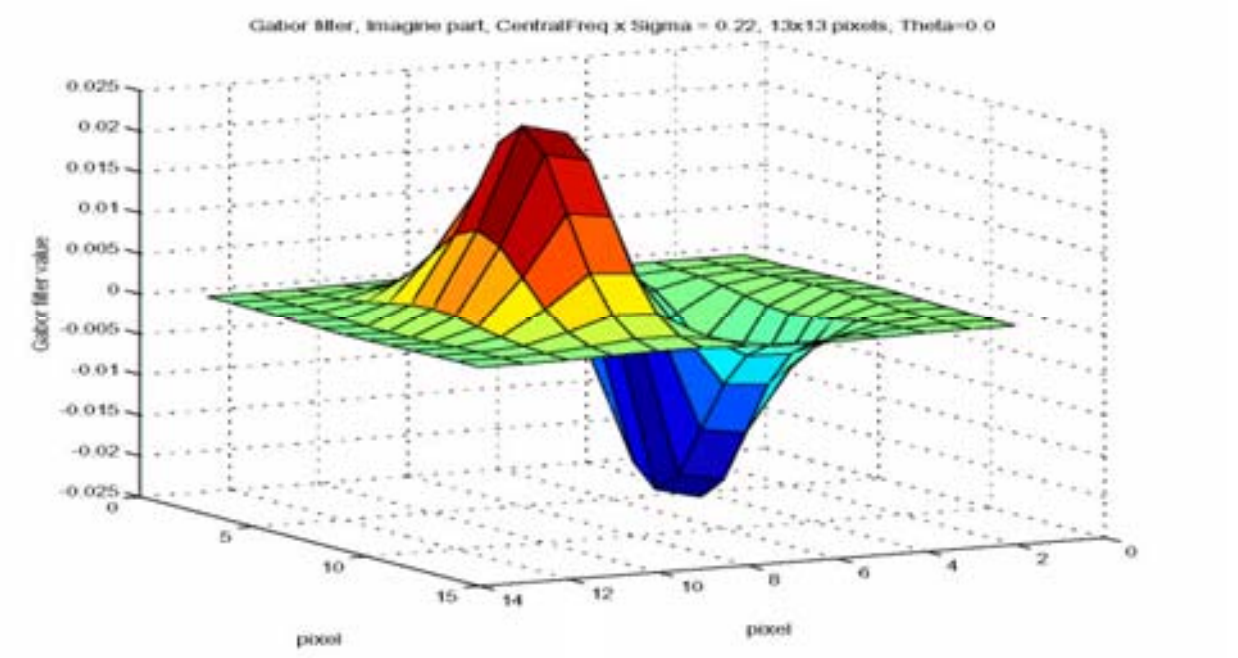

(b)

Fig. 3. Multi-scale multi-orientation filter bank consisting of Gabor wavelets such that: (I) A Gabor filter is a Gaussian function modulated by a complex sinusoid. (II) The real part of an oriented Gabor mother-wavelet is selected as an even-symmetric 2nd-order derivative of a Gaussian function, equivalent to a 3rd-order spatial statistic, in line with the works by Yellot [71] and Victor [74], see Fig. 3(a). According to [77], [87], this local filter is necessary and sufficient to detect any sort of image contours, namely, step edge, roof, line (ridge) and ramps (in compliance with the Mach band illusion [82]), as zero-crossings of the even-symmetric filtered image. (III) The imaginary part of an oriented Gabor mother-wavelet, shown in Fig. 3(b), provides an odd-symmetric 1st-order derivative of a Gaussian function, equivalent to a 2nd-order spatial statistic. According to the first author of the present study, in the raw primal sketch, this odd-symmetric filter is not employed in image contour detection, in agreement with [77], [87], but in multi-scale keypoint extraction exclusively, in line with [120], [121]. (IV) The oriented Gabor mother-wavelet is designed with a zero DC-component (to be insensitive to ramps and constant offsets), in line with [87]. (V) To provide the best compromise between computation time and the quality of the image decomposition/synthesis the following filter bank design is selected [87]. (i) Four dyadic spatial scales (one octave apart), with filter size equals to 3, 3×2, $3\times2^2$, $3\times2^3$ pixels, in agreement with [70]. (ii) Two spatial orientations: 0 and 90 degrees.

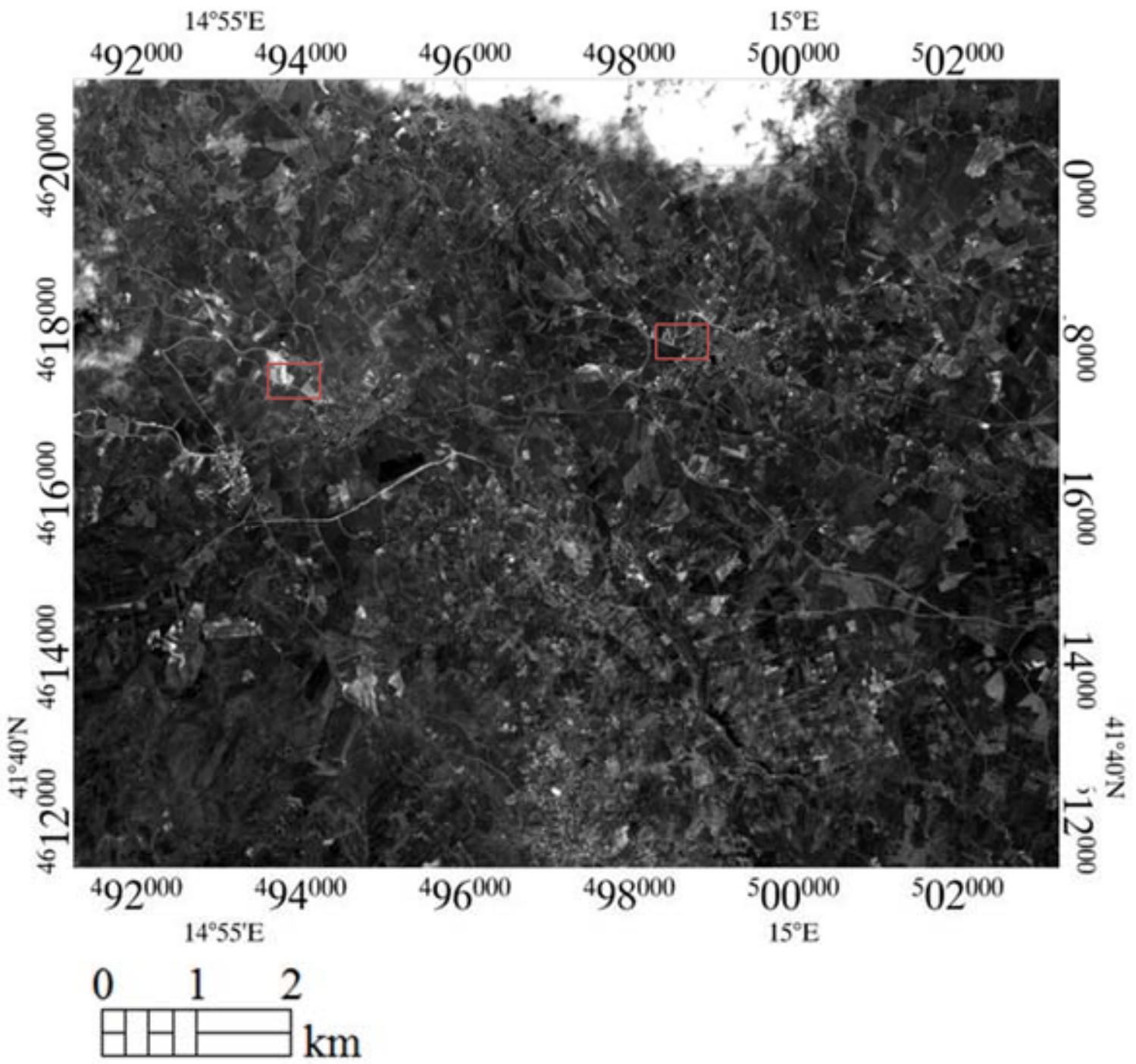

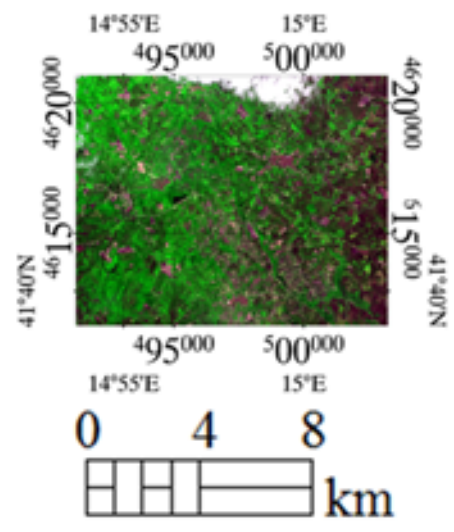

Fig. 4. (a) Right. Radiometrically calibrated QuickBird-2 MS image of the Campania region, Italy, acquired on 2004-06-13 at 09:58 a.m., depicted in false colors (R: band Red, G: band NIR, B: band Blue). Spatial resolution: 2.44 m. Default ENVI 2% linear histogram stretching applied for visualization purposes. (b) Left. Radiometrically calibrated QuickBird-2 PAN image of the Campania region, Italy, acquired on 2004-06-13 at 09:58 a.m. Spatial resolution: 0.61 m. Default ENVI 2% linear histogram stretching applied for visualization purposes. The two red rectangular outlines represent two zoomed areas, shown in Fig. 5 and Fig. 6 respectively.

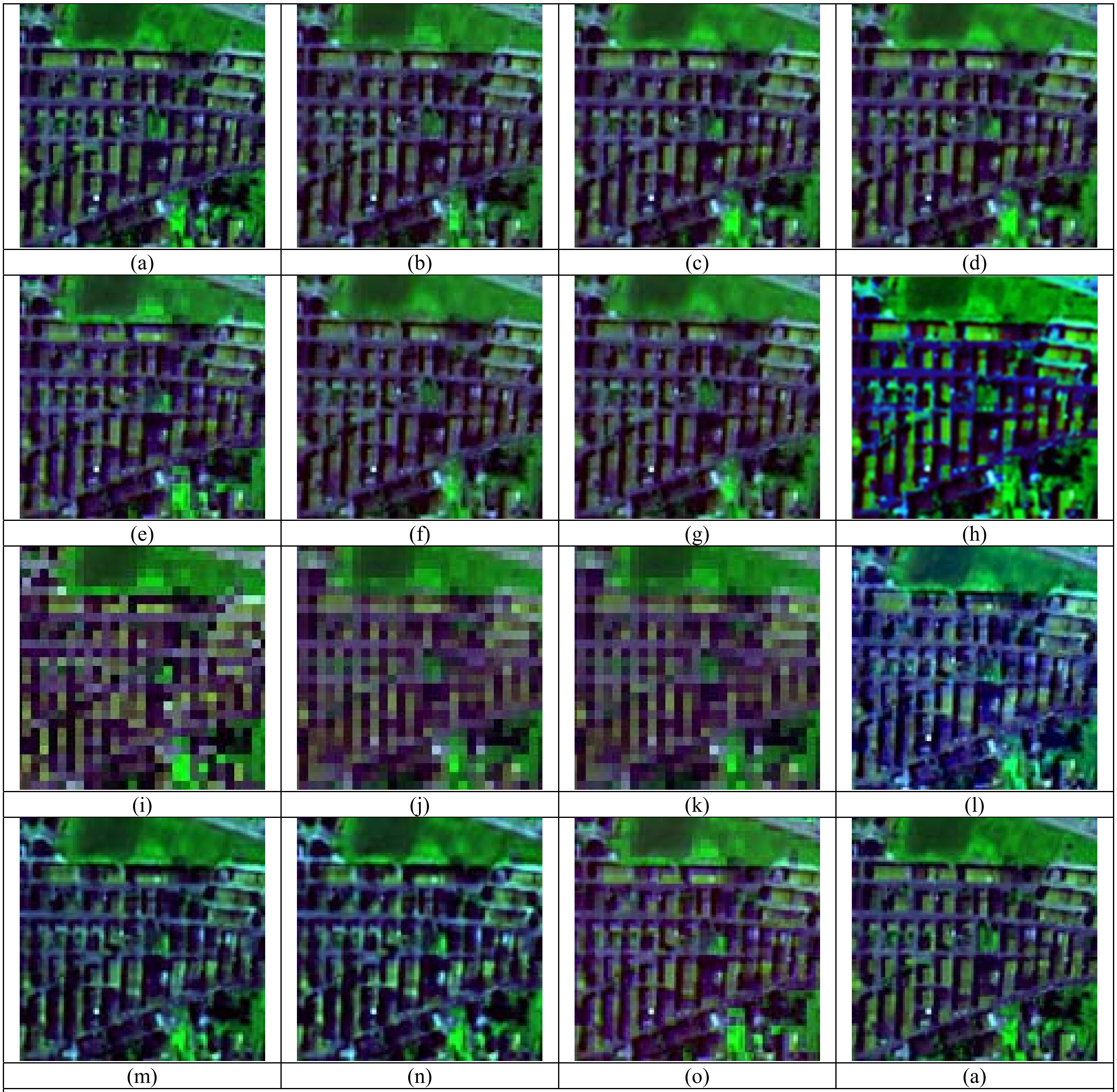

Fig. 5. Zoom of an urban area selected from the validation image shown in Fig. 4. Comparison of the reference sensory image, MSl, with the fourteen fused image. All images are depicted in false colors (R: band Red, G: band NIR, B: band Blue). Spatial resolution: 2.44 m. No histogram stretching is applied for visualization purposes. (a) Reference image (depicted top-left and bottom-right); (b) PC1_NN_PA; (c) PC2_B; (d) PC3_CC; (e) GS1_NN; (f) GS2_B_PA; (g) GS3_CC_PA; (h) CN2_PA; (i) DWT1; (j) DWT1_PA; (k)ATW2_PA; (l) EH1_PA; (m) HCS3_NN; (n) HCS7_CC; (o) RM.

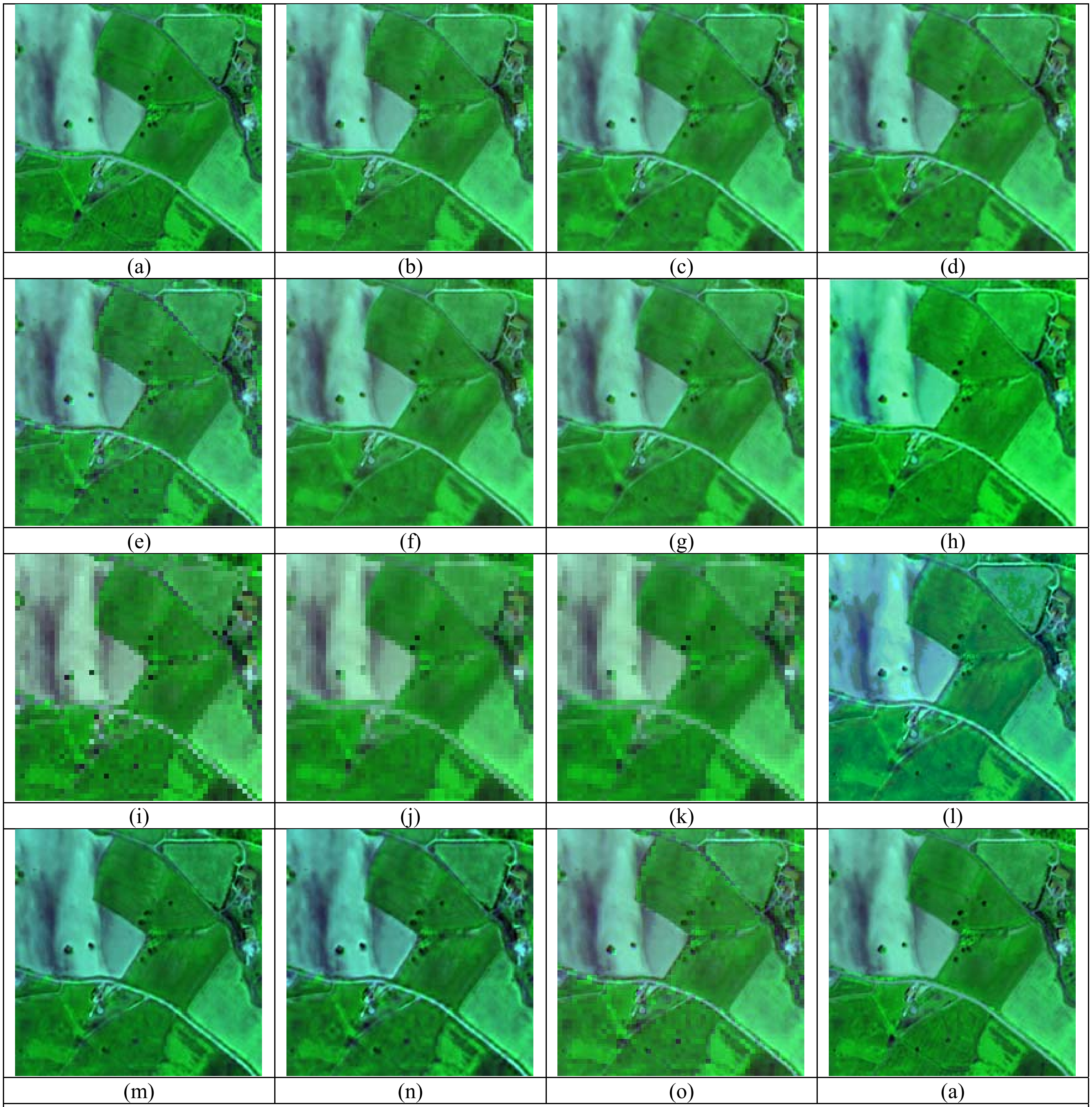

Fig. 6. Zoom of an agricultural area selected from the validation image shown in Fig. 4. Comparison of the reference sensory image, $MS_l$, with the fourteen fused image. All images are depicted in false colors (R: band Red, G: band Near Infrared, B: band Blue). Spatial resolution: 2.44 m. No histogram stretching is applied for visualization purposes. (a) Reference image (depicted top-left and bottom-right); (b) PC1_NN_PA; (c) PC2_B; (d) PC3_CC; (e) GS1_NN; (f) GS2_B_PA; (g) GS3_CC_PA; (h) CN2_PA; (i) DWT1; (j) DWT1_PA; (k)ATW2_PA; (l) EH1_PA; (m) HCS3_NN; (n) HCS7_CC; (o) RM.

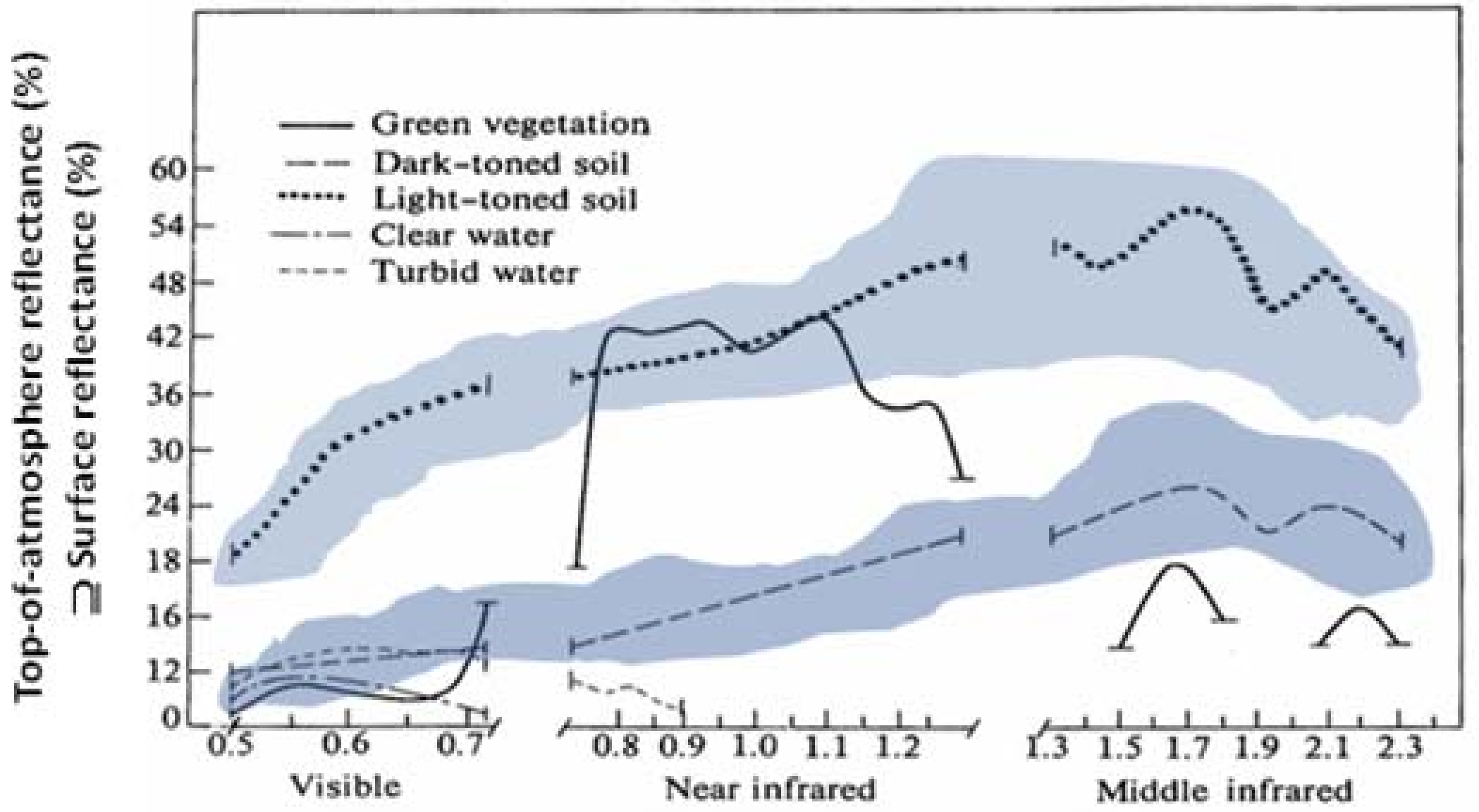

Fig. 7. Examples of land cover (LC)-class specific families of spectral signatures in top-of-atmosphere reflectance (TOARF) values. A within-class family of spectral signatures (e.g., dark-toned soil) in TOARF values forms a buffer zone (support area, envelope) which includes surface reflectance (SURF) values as a special case in clear sky and flat terrain conditions.

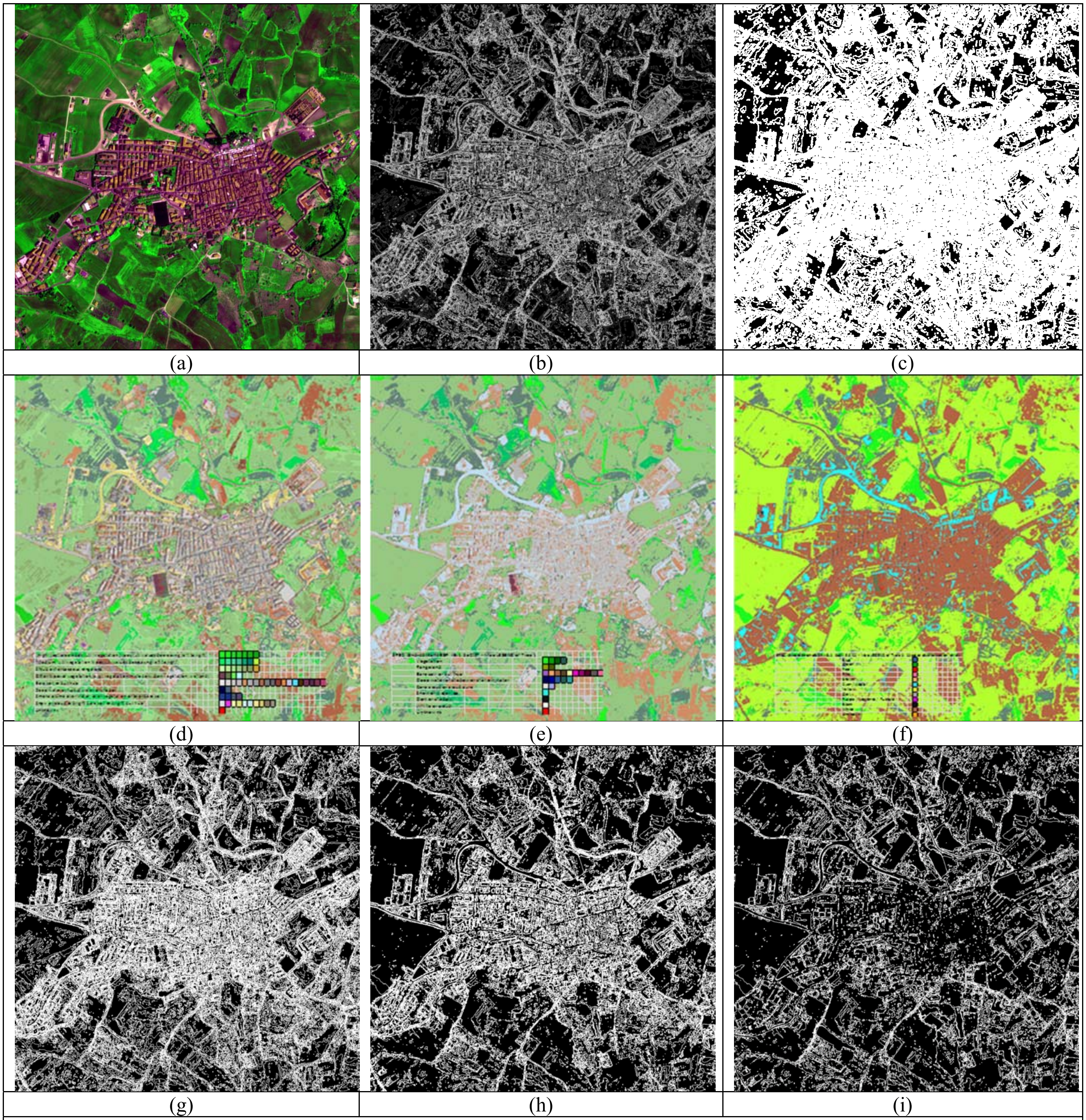

Fig. 8. (a) Zoomed area selected from the validation image shown in Fig. 4, depicted in false colors (R: band Red, G: band Near Infrared, B: band Blue), with histogram stretching for visualization purposes. (b) Three-level sum of 8-adjacency cross-aura measures shown in Fig. 8(g), (h) and (i). The per-pixel three-level cross-aura measure belongs to range {0, 24 = 8 × 3}. (c) Binarization of the three-level sum of cross-aura measures, shown in Fig. 8(b). (d), (e), (f) Q-SIAM pre-classification map, at fine/ intermediate/ coarse discretization levels, corresponding to 61/28/12 spectral categories (see Table 3). (g), (h), (i) 8-adjacency cross-aura measure in range {0, 8} per pixel, generated from the Q-SIAM pre-classification map at fine/ intermediate/ coarse discretization levels shown in Fig. 8(d), (e) and (f) respectively.

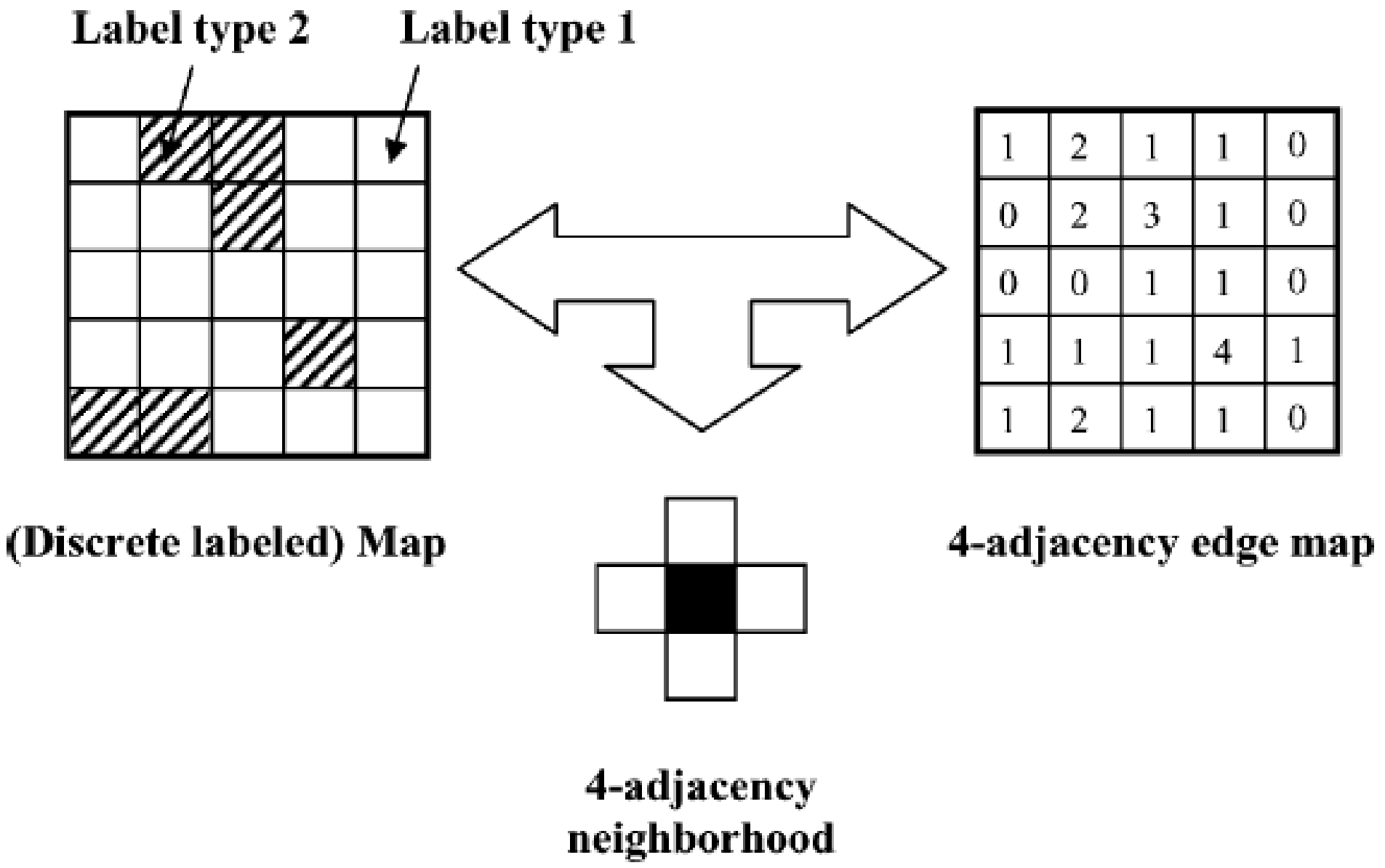

Fig. 9. Taken from [56]. Example of a 4-adjacency cross-aura map generated from a multi-level image, like a binary classification map.

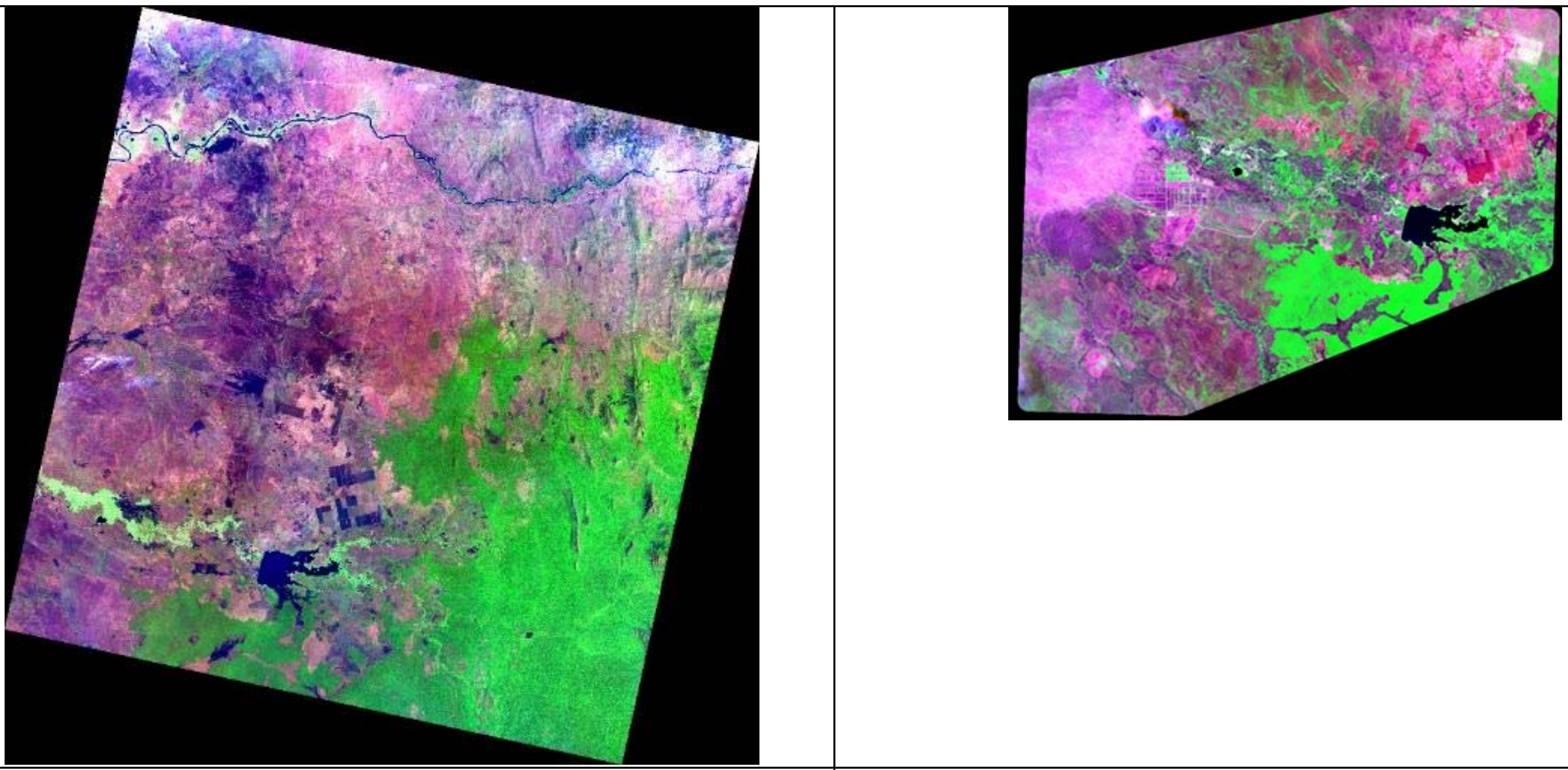

(a) (b)

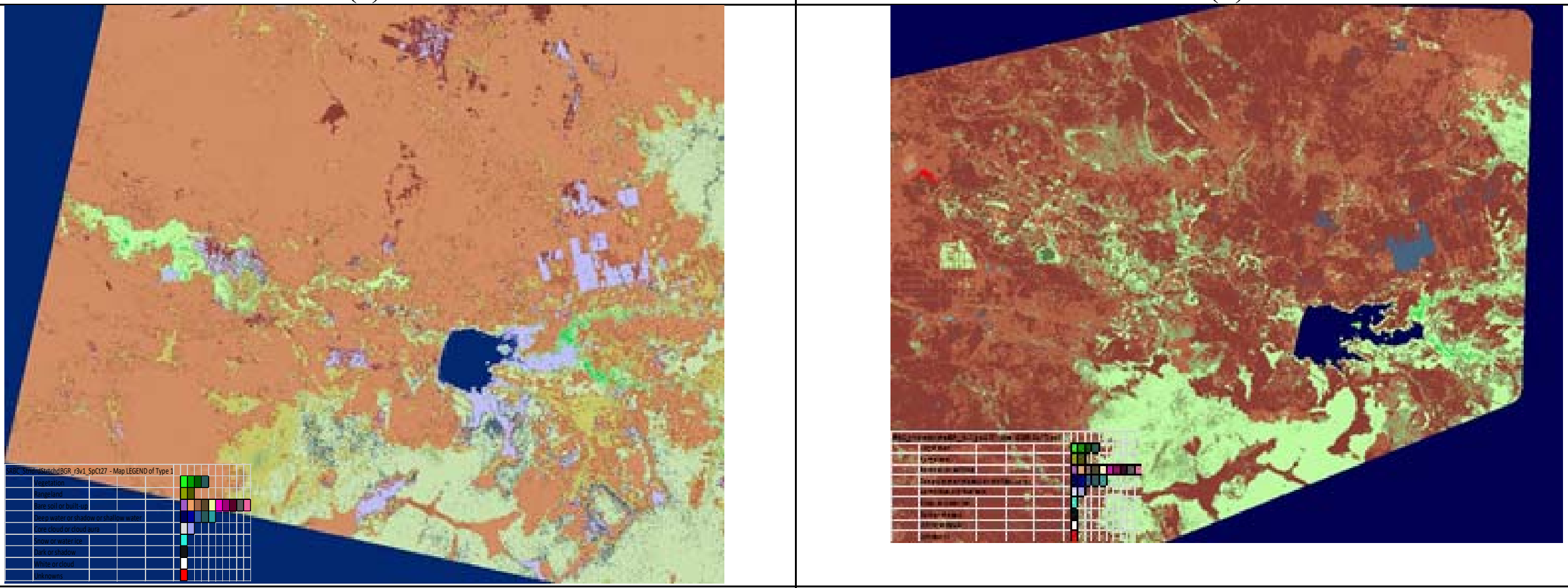

(c) (d)

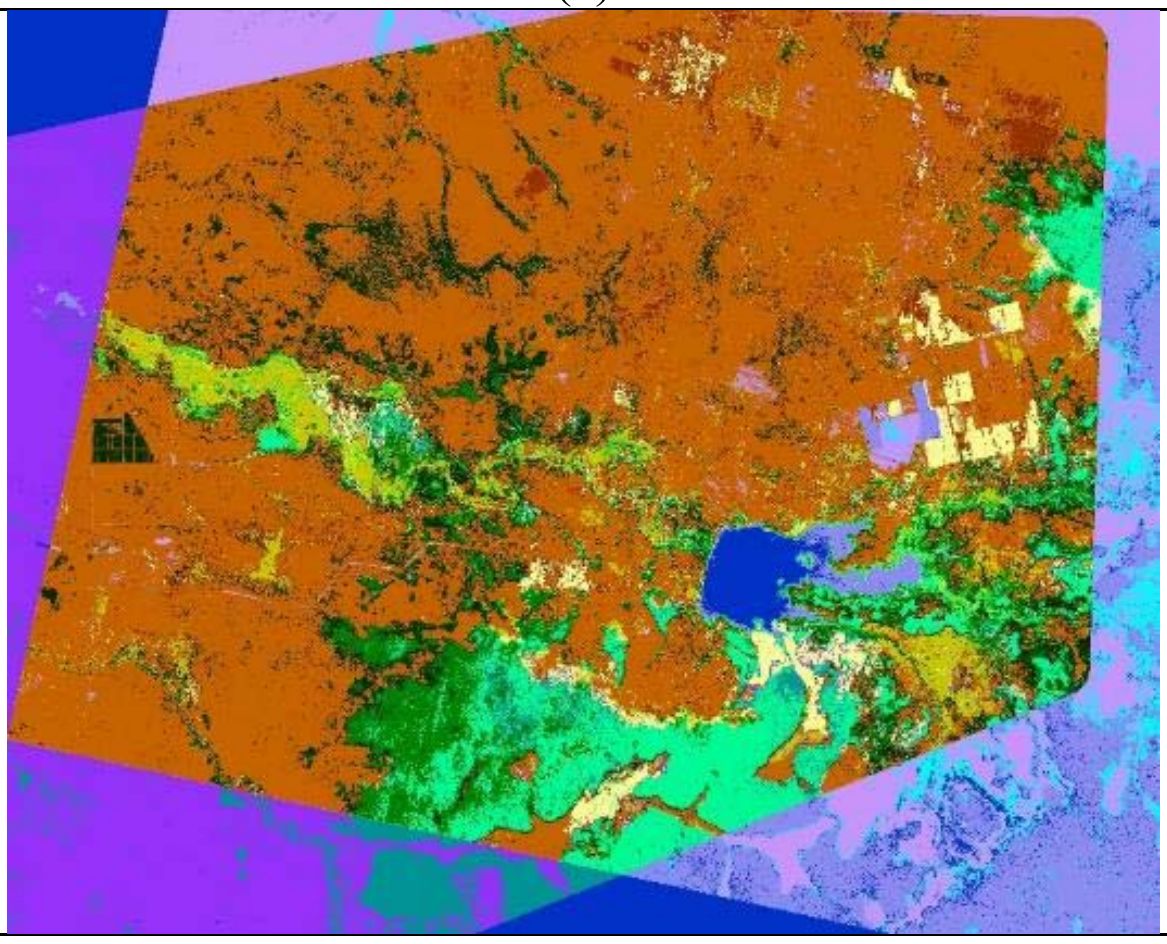

| | |
|---|---|
| 1 | Constant vegetation |
| 2 | Vegetation decrease |
| 3 | Vegetation increase |
| 4 | Vegetation total gain from bare soil or built-up or fire |
| 5 | Vegetation total loss into bare soil or built-up |
| 6 | Vegetation total gain from water (or shadow) |
| 7 | Vegetation total loss into water (or shadow) |
| 8 | Single-date vegetation (affected by data noise at either T1 or T2) |
| 9 | Bare soil or built-up total gain from water (or shadow) |
| 10 | Bare soil or built-up total loss into water (or shadow) |
| 11 | Constant water (or shadow) |
| 12 | Single-date water (or shadow) (affected by data noise at either T1 or T2) |
| 13 | Constant bare soil or built-up |
| 14 | Within-bare soil or built-up change |
| 15 | Single-date bare soil (affected by data noise at either T1 or T2) |
| 16 | Constant cloud or single-date cloud (affected by noise at either T1 or T2) |
| 17 | Constant snow (or bright bare soil/built-up or cloud in VHR imagery) |
| 18 | Single-date snow (or shadowed snow) (affected by data noise at either T1 or T2) |
| 19 | Snow (or bright bare soil/built-up or cloud in VHR imagery) total gain |
| 20 | Vegetation from snow (or bright bare soil/built-up or cloud in VHR imagery) |
| 21 | Bare soil or built-up from snow (or bright bare soil/built-up or cloud in VHR imagery) |
| 22 | Water (or shadow) from snow (or bright bare soil/built-up or cloud in VHR imagery) |
| 23 | Constant shadowed snow |
| 24 | Single-date shadowed snow (affected by data noise at either T1 or T2) |
| 25 | Constant shadow |
| 26 | Constant flame |
| 27 | Single-date flame (affected by data noise at either T1 or T2) |
| 28 | Active flame |
| 29 | Constant unknown or noisy |

(e) (f)

Fig. 10. Automatic SIAM-based post-classification change/no-change detection. (a) SPOT-5 image acquired on 2008-02-14, covering a surface area around Gambella, Ethiopia (DATASET_NAME: SCENE 5 130-334), spatial resolution: 10 m, depicted in false colors (R: MIR, G: NIR, B: G). Radiometrically calibrated into TOARF values. Default ENVI histogram stretching, linear 2%. (b) Mosaic of 6 RapidEye images acquired on 2014-04-08, covering a surface area around Gambella, Ethiopia, spatial resolution: 5 m, depicted in false colors (R: R, G: NIR, B: B). Radiometrically calibrated into TOARF values. Default ENVI

histogram stretching, linear 2%. (c) S-SIAM pre-classification map depicted in pseudo colors, generated from the SPOT-5/RapidEye inter-image overlapping area, upscaled to 5 m. This inter-sensor SIAM's map legend consists of 33 "shared" spectral categories. (d) Q-SIAM pre-classification map depicted in pseudo colors, generated from the SPOT-5/RapidEye inter-image overlapping area, 5 m resolution. This inter-sensor SIAM's map legend consists of 33 "shared" spectral categories. (e) Bi-temporal inter-sensor post-classification land surface change/no-change detection, automatically generated from the two SIAM's pre-classification maps. (f) SIAM-based post-classification change/no-change map's legend, consisting of 29 spectral categories.

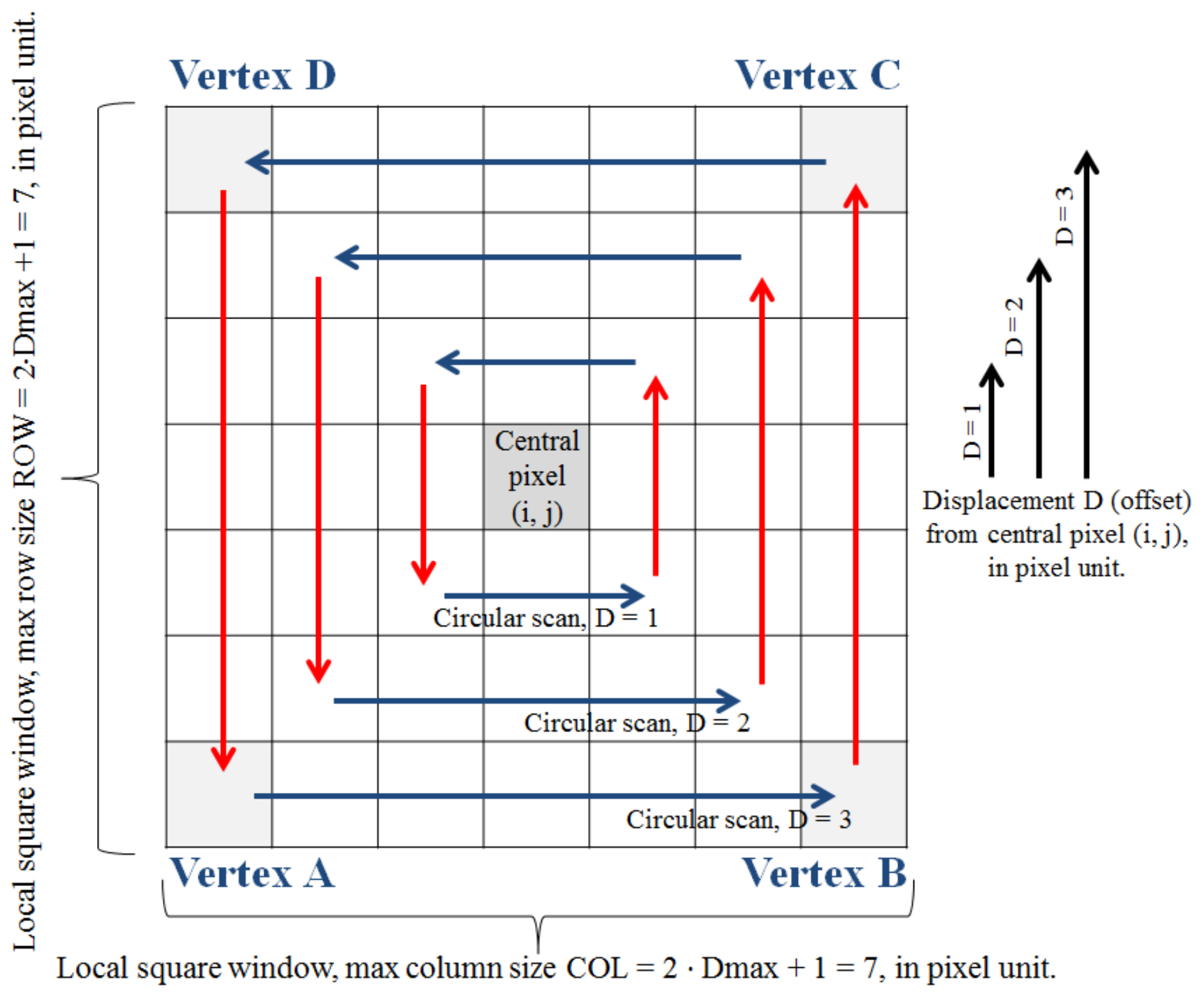

Fig. 11. Sketch of the TIMS-GLCM implementation, to collect 3rd-order statistics in the spatial domain. In this example, the scanning order of the local window centered on pixel (i,j) is counter-clockwise, for each spatial scale (radius, in this example, r = 1, 2, 3 in pixel unit, where r is equivalent to a displacement D from the center pixel). The central pixel, whose gray level is identified as GL1, provides the first GL value of each 3-tuple collected while scanning a complete circumference around the central pixel. The 2nd pixel is collected on the Right side – Bottom up (see arrow in red, pointing up), while the 3rd pixel is collected on the Left side – Top down (see arrow in red, pointing down). To close the circumference at a given radius r, the 2nd pixel is collected on the Top side – Right to left, while the 3rd pixel is collected on the Bottom side – Left to right. In each collected 3-tuple, the gray levels (GLs) of the 2nd and 3rd pixels are sorted, such that GL2 < GL3. The upper triangular TIMS-GLCM is input with values: row = GL2, column = GL3, depth = GL1, where GL2 < GL3, also refer to Fig. 12.

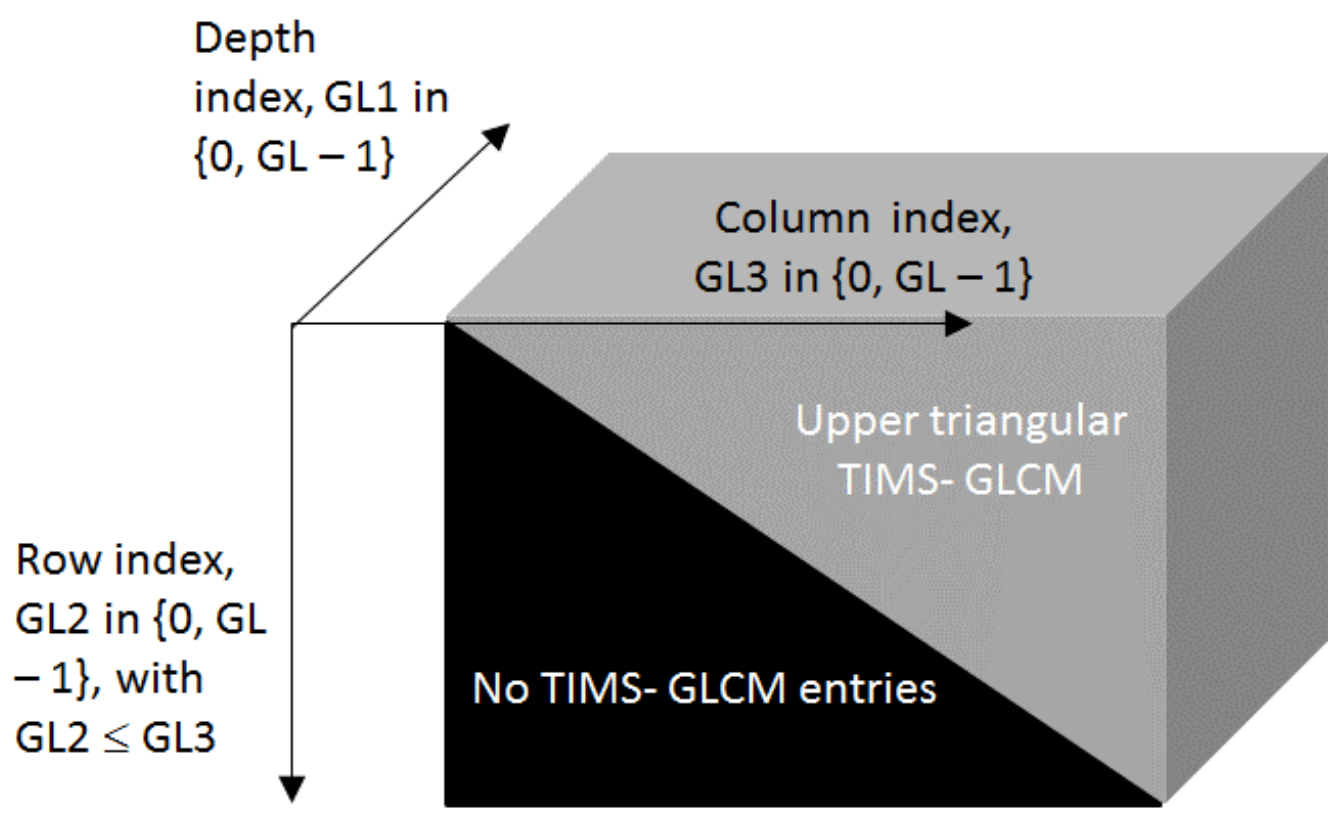

Fig. 12. Upper triangular TIMS-GLCM. An entry 3-tuple is: Row = GL2, Column = GL3, Depth = GL1, with GL2 ≤ GL3.

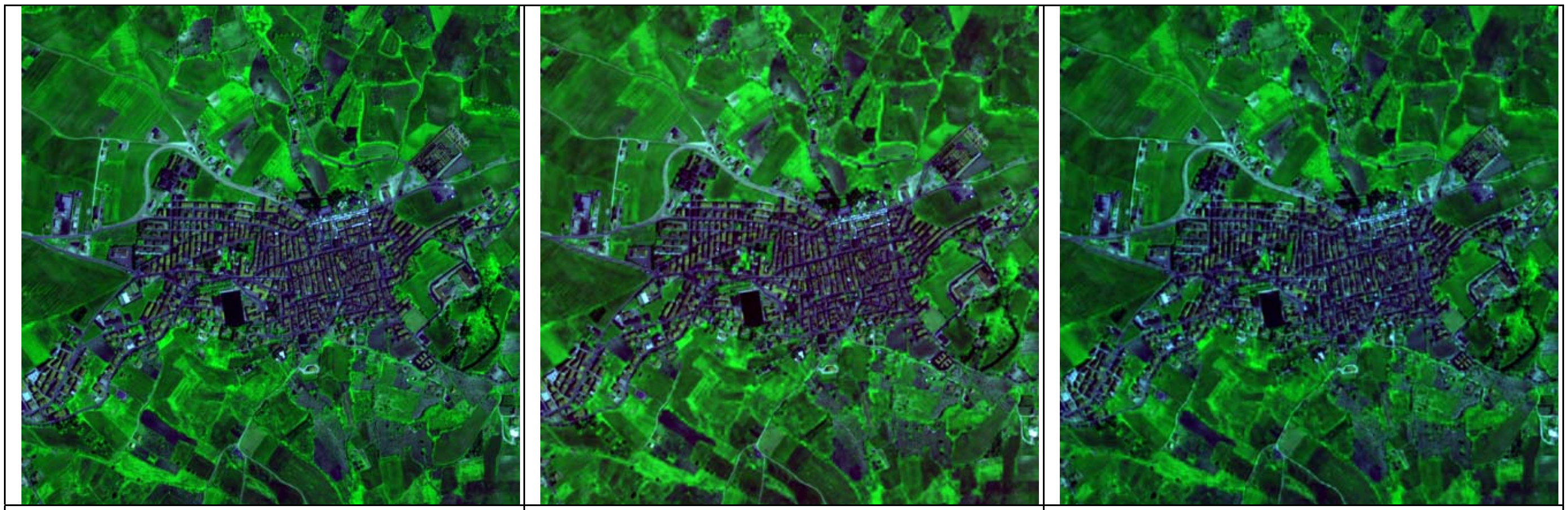

(a) (b) (c)

Fig. 13. Zoomed area selected from the validation image shown in Fig. 4, same as that shown in Fig. 8. Comparison of the reference image, $MS_l$, with the two fused $MS^*_l$ images considered "best" by human subjects. All images are depicted in false colors (R: band Red, G: band Near Infrared, B: band Blue). Spatial resolution: 2.44 m. No histogram stretching is applied for visualization purposes. (a) Reference image; (b) PC3_CC; (c) HCS3_NN.

TABLE CAPTIONS

Table 1. Test set of MS image PAN-sharpening algorithms. For each algorithm, there were one or more system's free-parameters to be user-defined. One input parameter was selected for discriminative purposes, i.e., its changes in value led to different runs of the same algorithm with different outcome. Other input parameters, if any, were kept fixed in the different runs by the same algorithm.

| **Acronym** | **Tested MS image PAN-sharpening algorithm** | **Discriminative input parameter - Resampling algorithm** |
|---|---|---|
| PC1_NN_PA | Principal Component | Nearest Neighbor (ENVI) |
| PC2_B | | Bilinear (ENVI) |
| PC3_CC | | Cubic Convolution (ENVI) |
| GS1_NN | Gram - Schmidt | Nearest Neighbor (ENVI) |
| GS2_B_PA | | Bilinear (ENVI) |
| GS3_CC_PA | | Cubic Convolution (ENVI) |
| CN2_PA | Color Normalized | Nearest Neighbor (ENVI) |
| DWT1 | Discrete Wavelet Transform | Nearest Neighbor (ENVI) |
| DWT1_PA | | Pixel Aggregate (IDL) |
| ATW2_PA | A Trous Wavelet Transform | Pixel Aggregate (IDL) |
| HCS3_NN | Hyperspectral color space | Nearest Neighbor (ERDAS) |
| HCS7_CC | | Cubic Convolution (ERDAS) |
| EH1 | Ehlers | Nearest Neighbor (ERDAS) |
| RM | Resolution Merge | Nearest Neighbor (ERDAS) |

Table 2. Perceptual inter-image quality (similarity) assessment by human subjects: legend of the qualitative (categorical) variable.

| **Either spatial or spectral quality** | **Human assessment** |
|---|---|
| A | 1 - Excellent |
| B | 2 - Very good |
| C | 3 - Good |
| D | 4 - Fairly good |
| E | 5 – Sufficient |
| F | 6 - Insufficient (bad) |
| G | 7 - Very bad |

Table 3. The SIAM prior knowledge-based MS data quantizer is an EO system of systems, scalable to any past, existing or future MS imaging sensors, in compliance with the GEOSS implementation plan [4].

| **SIAM, r88v5** | **Input bands** | **Preliminary classification map output products: Number of output spectral categories** | | | |
|---|---|---|---|---|---|
| | | **Fine discretization levels** | **Intermediate discretization levels** | **Coarse discretization levels** | **Inter-sensor discretization levels (*)** |
| **L-SIAM** | **7 – B, G, R, NIR, MIR1, MIR2, TIR** | **96** | **48** | **18** | **33**<br>* employed for inter-sensor post-classification change/no-change detection |
| **S-SIAM** | **4 – G, R, NIR, MIR1** | **68** | **40** | **15** | |
| **AV-SIAM** | **4 – R, NIR, MIR1, TIR** | **83** | **43** | **17** | |
| **Q-SIAM** | **4 – B, G, R, NIR** | **61** | **28** | **12** | |

Table 4. Novel categorization of MS image PAN-sharpening product QIs and quality metrics, based on three nominal scales: (a) univariate (one-channel, Unvrt)/ bivariate (two-channel, Bivrt)/ multivariate (multi-channel, Mvrt), (b) 1st- to 3rd-order statistic in the spatial domain, (c) categories 1 to 4: SPCTRL, SPCTRL & SPTL1, SPCTRL & SPTL2, SPCTRL & SPTL1 & SPTL2. The Inter-QI metric functions (Metric) considered are: Minkowski distance of order 1 (MD) across bands (MDB), Across-band average (ABA).

| **Product QI category** | **QI name and description** | **Metric** | **Acronym** | **Statistic analysis** | **Statistic order in the spatial domain** |
|---|---|---|---|---|---|
| 1. Context-insensitive (pixel-based) Position (row and column)-independent Spectral cost indexes (SPCTRL) | Mean scalar value, band-specific | MDB | MeanUnvrt | Unvrt | 1st |
| | Standard deviation scalar value, band-specific | MDB | StDvUnvrt | Unvrt | 1st |
| | Skewness scalar value, band-specific | MDB | SkwnsUnvrt | Unvrt | 1st |
| | Kurtosis scalar value, band-specific | MDB | KrtsUnvrt | Unvrt | 1st |
| | Entropy scalar value, band-specific | MDB | EntrpyUnvrt | Unvrt | 1st |
| 2. Context-insensitive Position-dependent Spectral cost indexes (SPCTRL & SPTL1) | Cumulative (image-wide) per-pixel absolute difference in a pair of SIAM-based pre-classification maps (post-classification change detection) | - | PostClChngDtctn Mvrt | Mvrt | 1st |
| | Inverse correlation coefficient (to be considered a cost value, to be minimized) computed inter-image band-specific | ABA | InvrsCrltnBivrt | Bivrt | 1st |
| 3. Context-sensitive Position-independent Spectral cost indexes (SPCTRL & SPTL2) | 3rd-order Contrast scalar value, band-specific | MDB | 3ordrCntrstUnvrt | Unvrt | 3rd |
| | 3rd-order Energy scalar value, band-specific | MDB | 3ordrEnrgyUnvrt | Unvrt | 3rd |
| | 3rd-order Large Number Emphasis scalar value, image band-specific | MDB | 3ordrLneUnvrt | Unvrt | 3rd |
| | Mean (image-wide) per-pixel SIAM-based multi-level 8-adjacency cross-aura contour measure (in range {0, 8} per pixel) | MD | CntourXauraMvrt | Mvrt | 1st |
| 4. Context-sensitive Position-dependent Spectral cost indexes (SPCTRL & SPTL1 & SPTL2) | Cumulative (image-wide) per-pixel absolute difference in a pair of SIAM-based multi-level 8-adjacency cross-aura binary contour maps (in range {0, 1} per pixel) | MD | BinaryCntourMvrt | Mvrt | 1st |

Table 5. Visual score of the tested MS image PAN-sharpened outcome. Green highlight: first- and second-best choice by human subjects.

| **Algorithm** | **Acronym** | **Subjective quality assessment** | |
|---|---|---|---|
| | | **Spectral quality** | **Spatial quality** |
| Principal Component | PC1_NN_PA | C | A |
| | PC2_B | B | A |
| | PC3_CC | C | A |
| Gram - schmidt | GS1_NN | C | D |
| | GS2_B_PA | D | D |
| | GS3_CC_PA | D | D |
| Color Normalized | CN2_PA | E | E |
| Discrete Wavelet Transform | DWT1 | F | G |
| | DWT1_PA | F | G |
| A Trous Wavelet Transform | ATW2_PA | F | G |
| Hyperspherical Color Space | HCS3_NN | A | B |
| | HCS7_CC | B/C | D |
| Ehlers | EH1 | G | E |
| Resolution Merge | RM | C | E |

Table 6. Test QuickBird MS image PAN-sharpening: product QI (actually, cost index) values and standardized QI values. Product QI (actually, cost index) category 1: SPCTRL. Product QI (actually, cost index) category 2: SPCTRL & SPTL1. Product QI (actually, cost index) category 3: SPCTRL & SPTL2. Product QI (actually, cost index) category 4: SPCTRL & SPTL1 & SPTL2. Cost index values to be minimized (best when smaller).

| Algorithm | SPCTRL | | | | | SPCTRL & SPTL1 | | SPCTRL & SPTL2 | | | | SPCTRL & SPTL1 & SPTL2 |
|---|---|---|---|---|---|---|---|---|---|---|---|---|
| | **Mean Unvrt** | **StDvUnvrt** | **SkwnsUnvrt** | **KrtsUnvrt** | **Entrpy Unvrt** | **PostClChngDtctnMvrt** | **InvrsCrltnBivrt** | **3ordr Cntrst Unvrt** | **3ordr EnrgyUnvrt** | **3ordr LneUnvrt** | **Cntour Xaura Mvrt** | **BinaryCntourMvrt** |
| PC1_NN_PA | 0.033 | 0.180 | 0.366 | 2.324 | 0.011 | 0.271 | 0.006 | 0.014 | 0.011 | 0.011 | 3.867 | 0.233 |
| PC2_B | 0.001 | 0.199 | 0.348 | 2.213 | 0.037 | 0.261 | 0.005 | 0.012 | 0.010 | 0.170 | 3.753 | 0.231 |
| PC3_CC | 0.002 | 0.029 | 0.356 | 2.270 | 0.058 | 0.264 | 0.006 | 0.066 | 0.013 | 0.030 | 3.699 | 0.232 |
| GS1_NN | 0.000 | 0.171 | 0.300 | 1.847 | 0.082 | 0.336 | 0.003 | 0.113 | 0.016 | 0.300 | 4.214 | 0.249 |
| GS2_B_PA | 0.153 | 0.178 | 0.299 | 1.855 | 0.081 | 0.299 | 0.007 | 0.019 | 0.011 | 0.510 | 4.093 | 0.254 |
| GS3_CC_PA | 0.218 | 0.176 | 0.302 | 1.901 | 0.106 | 0.299 | 0.008 | 0.009 | 0.012 | 0.724 | 4.088 | 0.243 |
| CN1 | 3.702 | 0.411 | 0.312 | 2.117 | 0.149 | 0.323 | 0.005 | 0.289 | 0.019 | 15.762 | 4.221 | 0.301 |
| DWT1 | 0.086 | 0.169 | 0.174 | 0.635 | 1.457 | 0.329 | 0.085 | 0.046 | 0.015 | 0.113 | 4.433 | 0.381 |
| DWT1_PA | 0.072 | 0.226 | 0.044 | 1.206 | 1.650 | 0.241 | 0.039 | 0.741 | 0.050 | 0.518 | 4.261 | 0.394 |
| ATW2_PA | 0.020 | 0.359 | 0.222 | 1.792 | 1.753 | 0.238 | 0.031 | 0.791 | 0.052 | 0.332 | 4.272 | 0.409 |
| HCS3_NN | 0.099 | 0.216 | 0.104 | 0.577 | 0.028 | 0.268 | 0.010 | 0.129 | 0.001 | 0.502 | 3.918 | 0.223 |
| HCS7_CC | 0.409 | 0.627 | 0.090 | 0.002 | 0.171 | 0.319 | 0.039 | 0.710 | 0.017 | 0.916 | 4.020 | 0.237 |
| EH1 | 46.67 | 7.388 | 2.124 | 12.517 | 0.718 | 0.950 | 0.201 | 1.069 | 0.127 | 277.94 | 4.885 | 0.383 |
| RM | 4.349 | 0.961 | 0.338 | 2.229 | 0.059 | 0.462 | 0.007 | 0.156 | 0.009 | 16.923 | 4.916 | 0.330 |
| | | | | | | | | | | | | |
| MEAN | 3.987 | 0.806 | 0.384 | 2.392 | 0.454 | 0.347 | 0.032 | 0.297 | 0.026 | 22.482 | 4.189 | 0.293 |
| STDV | 12.37 | 1.909 | 0.512 | 3.004 | 0.658 | 0.182 | 0.054 | 0.364 | 0.033 | 73.752 | 0.366 | 0.071 |
| **Standardized QI (z = (QI – MEAN) / STDV, such that E[z] = 0, STDV[z] = 1.** | | | | | | | | | | | | |
| PC1_NN_PA | -0.320 | -0.328 | -0.036 | -0.023 | -0.673 | -0.419 | -0.484 | -0.777 | -0.456 | -0.305 | -0.879 | -0.837 |
| PC2_B | -0.322 | -0.318 | -0.071 | -0.059 | -0.634 | -0.472 | -0.501 | -0.783 | -0.494 | -0.303 | -1.190 | -0.865 |
| PC3_CC | -0.322 | -0.407 | -0.055 | -0.040 | -0.603 | -0.458 | -0.497 | -0.634 | -0.412 | -0.304 | -1.338 | -0.849 |
| GS1_NN | -0.322 | -0.333 | -0.164 | -0.181 | -0.566 | -0.062 | -0.552 | -0.506 | -0.312 | -0.301 | 0.070 | -0.620 |
| GS2_B_PA | -0.310 | -0.329 | -0.167 | -0.179 | -0.567 | -0.263 | -0.469 | -0.764 | -0.449 | -0.298 | -0.261 | -0.544 |
| GS3_CC_PA | -0.305 | -0.330 | -0.161 | -0.163 | -0.529 | -0.265 | -0.463 | -0.791 | -0.415 | -0.295 | -0.275 | -0.705 |
| CN1 | -0.023 | -0.207 | -0.142 | -0.091 | -0.465 | -0.131 | -0.512 | -0.022 | -0.207 | -0.091 | 0.089 | 0.113 |
| DWT1 | -0.315 | -0.334 | -0.410 | -0.585 | 1.524 | -0.102 | 0.986 | -0.689 | -0.345 | -0.303 | 0.668 | 1.239 |
| DWT1_PA | -0.317 | -0.304 | -0.664 | -0.395 | 1.817 | -0.583 | 0.133 | 1.218 | 0.749 | -0.298 | 0.198 | 1.422 |
| ATW2_PA | -0.321 | -0.234 | -0.317 | -0.200 | 1.973 | -0.597 | -0.027 | 1.353 | 0.790 | -0.300 | 0.228 | 1.625 |
| HCS3_NN | -0.314 | -0.309 | -0.547 | -0.604 | -0.647 | -0.431 | -0.415 | -0.463 | -0.753 | -0.298 | -0.739 | -0.976 |
| HCS7_CC | -0.289 | -0.094 | -0.574 | -0.795 | -0.430 | -0.155 | 0.130 | 1.131 | -0.286 | -0.292 | -0.461 | -0.789 |
| EH1 | 3.451 | 3.447 | 3.398 | 3.370 | 0.401 | 3.304 | 3.141 | 2.117 | 3.106 | 3.464 | 1.902 | 1.264 |
| RM | 0.029 | 0.081 | -0.090 | -0.054 | -0.601 | 0.632 | -0.471 | -0.389 | -0.516 | -0.075 | 1.987 | 0.522 |
| | | | | | | | | | | | | |
| MEAN | 0.000 | 0.000 | 0.000 | 0.000 | 0.000 | 0.000 | 0.000 | 0.000 | 0.000 | 0.000 | 0.000 | 0.000 |
| STDV | 1.000 | 1.000 | 1.000 | 1.000 | 1.000 | 1.000 | 1.000 | 1.000 | 1.000 | 1.000 | 1.000 | 1.000 |

Table 7. Sum of within-category standardized product QIs (actually, cost indexes), to be minimized (best when more negative) and partial ranking for each category of product QIs (product partial rank, PDPR). Product QI (actually, cost index) category 1: SPCTRL - Sum of five standardized cost indexes to be minimized: MeanUnvrt, StDvUnvrt, SkwnsUnvrt, KrtsUnvrt, En-trpyUnvrt. Product QI (actually, cost index) category 2: SPCTRL & SPTL1 – Case (i) = Sum of two standardized cost indexes to be minimized: PostClChngDtctnMvrt, InvrsCrltnBivrt; Case (ii) = InvrsCrltnBivrt is omitted, i.e., only PostClChng-DtctnMvrt is considered. Product QI (actually, cost index) category 3: SPCTRL & SPTL2 - Sum of four standardized cost indexes to be minimized: 3ordrCntrstUnvrt, 3ordrEnrgyUnvrt, 3ordrLneUnvrt, CntourXauraMvrt. Product QI (actually, cost index) category 4: SPCTRL & SPTL1 & SPTL2 - Single standardized cost index to be minimized: BinaryCntourMvrt.

| Algorithm | SPCTRL | | SPCTRL & SPTL1 | | | | SPCTRL & SPTL2 | | SPCTRL & SPTL1 & SPTL2 | |
|---|---|---|---|---|---|---|---|---|---|---|
| | **Sum** | **PDPR** | **Sum, case (i)** | **PDPR, case (i)** | **Stndrdzd PostClChngDtctnMvrt, case (ii)** | **PDPR, case (ii)** | **Sum** | **PDPR** | **Stndrdzd BinaryCntour Mvrt** | **PDPR** |
| PC1_NN_PA | -1.38 | 8 | -0.90 | 3 | -0.41859 | 6 | -2.42 | 3 | -0.837 | 4 |
| PC2_B | -1.41 | 7 | -0.97 | 1 | -0.47177 | 3 | -2.77 | 1 | -0.865 | 2 |
| PC3_CC | -1.43 | 6 | -0.96 | 2 | -0.45796 | 4 | -2.69 | 2 | -0.849 | 3 |
| GS1_NN | -1.57 | 3 | -0.61 | 9 | -0.06234 | 12 | -1.05 | 7 | -0.620 | 7 |
| GS2_B_PA | -1.55 | 4 | -0.73 | 5 | -0.26270 | 8 | -1.77 | 6 | -0.544 | 8 |
| GS3_CC_PA | -1.49 | 5 | -0.73 | 6 | -0.26511 | 7 | -1.78 | 5 | -0.705 | 6 |
| CN1 | -0.93 | 9 | -0.64 | 7 | -0.13064 | 10 | -0.23 | 9 | 0.113 | 9 |
| DWT1 | -0.12 | 11 | 0.88 | 13 | -0.10176 | 11 | -0.67 | 8 | 1.239 | 11 |
| DWT1_PA | 0.14 | 12 | -0.45 | 10 | -0.58338 | 2 | 1.87 | 12 | 1.422 | 13 |
| ATW2_PA | 0.90 | 13 | -0.62 | 8 | -0.59653 | 1 | 2.07 | 13 | 1.625 | 14 |
| HCS3_NN | -2.42 | 1 | -0.85 | 4 | -0.43093 | 5 | -2.25 | 4 | -0.976 | 1 |
| HCS7_CC | -2.18 | 2 | -0.02 | 11 | -0.15457 | 9 | 0.09 | 10 | -0.789 | 5 |
| EH1 | 14.07 | 14 | 6.44 | 14 | 3.30399 | 14 | 10.59 | 14 | 1.264 | 12 |
| RM | -0.63 | 10 | 0.16 | 12 | 0.63228 | 13 | 1.01 | 11 | 0.522 | 10 |

Table 8. Battery of process QIs (actually, cost indexes, to be minimized) adopted by the new evaluation procedure. The number of system's free-parameters is a cost index, inversely related to the degree of automation. Process QI (actually, cost index) par-tial rank: PSPR.

| **Algorithm** | **Processing time (cost index 1, in seconds)** | **PSPR1** | **No. of system's free-parameters (cost index 2)** | **PSPR2** |
|---|---|---|---|---|
| PC1_NN_PA | 00:02:56:541 | **8** | 1 | **1** |
| PC2_B | 00:03:10:061 | **9** | 1 | **1** |
| PC3_CC | 00:03:23:870 | **10** | 1 | **1** |
| GS1_NN | 00:01:24:735 | **6** | 2 | **2** |
| GS2_B_PA | 00:01:18:118 | **5** | 2 | **2** |
| GS3_CC_PA | 00:01:39:022 | **7** | 2 | **2** |
| CN1 | 00:00:28:820 | **1** | 1 | **1** |
| DWT1 | 00:22:10:514 | **14** | 3 | **3** |
| DWT1_PA | 00:21:44:012 | **13** | 3 | **3** |
| ATW2_PA | 00:16:45:224 | **12** | 1 | **1** |
| HCS3_NN | 00:01:12:256 | **4** | 4 | **4** |
| HCS7_CC | 00:01:10:852 | **3** | 4 | **4** |
| EH1 | 00:04:42:274 | **11** | 5 | **5** |
| RM | 00:00:58:713 | **2** | 5 | **5** |

Table 9. Product final ranks (PDFR), computed from the sum of the individual product quality partial ranks (PDPRs) collected from the four categories of product QIs, refer to Table 7. Case A is alternative to Case C. The latter holds when the InvrsCrltnBivrt cost index is removed from the product QI category SPCTRL & SPTL1. Noteworthy, in these experiments, the sole QI be-longing to category 4, SPCTRL & SPTL1 & SPTL2, is the individual indicator that best approximates (which is highly cor-related with) the final ranks A and C, although no single "universal" quality indicator can exist. Yellow highlight: first-best choice by quantitative quality estimation. Red highlight: second-best choice by quantitative quality estimation.

| Algorithm | SPCTRL, PDPR | SPCTRL & SPTL1, case (i), PDPR | SPCTRL & SPTL1, case (ii), PDPR | SPCTRL & SPTL2, PDPR | SPCTRL & SPTL1 & SPTL2, PDPR | Sum of PDPRs, with (i) - Case A | PDFR, with (i) - Case A | Sum of PDPRs, with (ii) - Case C | PDFR, with (ii) - Case C |
|---|---|---|---|---|---|---|---|---|---|
| PC1_NN_PA | 8 | 3 | 6 | 3 | 4 | 18 | **4** | 21 | **4** |
| PC2_B | 7 | 1 | 3 | 1 | 2 | 11 | **2** | 13 | **2** |
| PC3_CC | 6 | 2 | 4 | 2 | 3 | 13 | **3** | 15 | **3** |
| GS1_NN | 3 | 9 | 12 | 7 | 7 | 26 | **7** | 29 | **8** |
| GS2_B_PA | 4 | 5 | 8 | 6 | 8 | 23 | **6** | 26 | **6** |
| GS3_CC_PA | 5 | 6 | 7 | 5 | 6 | 22 | **5** | 23 | **5** |
| CN2_PA | 9 | 7 | 10 | 9 | 9 | 34 | **9** | 37 | **9** |
| DWT1 | 11 | 13 | 11 | 8 | 11 | 43 | **10** | 41 | **11** |
| DWT1_PA | 12 | 10 | 2 | 12 | 13 | 47 | **12** | 39 | **10** |
| ATW2_PA | 13 | 8 | 1 | 13 | 14 | 48 | **13** | 41 | **11** |
| HCS3_NN | 1 | 4 | 5 | 4 | 1 | 10 | **1** | 11 | **1** |
| HCS7_CC | 2 | 11 | 9 | 10 | 5 | 28 | **8** | 26 | **6** |
| EH1 | 14 | 14 | 14 | 14 | 12 | 54 | **14** | 54 | **14** |
| RM | 10 | 12 | 13 | 11 | 10 | 43 | **10** | 44 | **13** |

Table 10. Product & Process final ranks (PPFRs), computed from the sum of product quality partial ranks (PDPRs) collected from each of the four categories of product's QI, refer to Table 7, in addition to the two process quality partial ranks (PSPRs), reported in Table 8. Case B is alternative to Case D. The latter holds when the InvrsCrltnBivrt cost index is removed from the product QI category SPCTRL & SPTL1, refer to case (ii). Yellow highlight: first-best choice by quantitative quality estimation. Red highlight: second-best choice by quantitative quality estimation.

| Algorithm | SPCTRL, PDPR | SPCTRL & SPTL1, case (i), PDPR | SPCTRL & SPTL1, case (ii), PDPR | SPCTRL & SPTL2, PDPR | SPCTRL & SPTL1 & SPTL2, PDPR | PSPR1, Processing time | PSPR2, No. of system's free-parameters | Sum of PDPRs and PSPRs, Case B | PPFR, Case B | Sum of PDPRs and PSPRs, Case D | PPFR, Case D |
|---|---|---|---|---|---|---|---|---|---|---|---|
| PC1_NN_PA | 8 | 3 | 6 | 3 | 4 | 8 | 1 | 27 | **4** | 30 | **4** |
| PC2_B | 7 | 1 | 3 | 1 | 2 | 9 | 1 | 21 | **2** | 23 | **2** |
| PC3_CC | 6 | 2 | 4 | 2 | 3 | 10 | 1 | 24 | **3** | 26 | **3** |
| GS1_NN | 3 | 9 | 12 | 7 | 7 | 6 | 2 | 34 | **7** | 37 | **8** |
| GS2_B_PA | 4 | 5 | 8 | 6 | 8 | 5 | 2 | 30 | **5** | 33 | **6** |
| GS3_CC_PA | 5 | 6 | 7 | 5 | 6 | 7 | 2 | 31 | **6** | 32 | **5** |
| CN2_PA | 9 | 7 | 10 | 9 | 9 | 1 | 1 | 36 | **9** | 39 | **9** |
| DWT1 | 11 | 13 | 11 | 8 | 11 | 14 | 3 | 60 | **11** | 58 | **13** |
| DWT1_PA | 12 | 10 | 2 | 12 | 13 | 13 | 3 | 63 | **13** | 55 | **12** |
| ATW2_PA | 13 | 8 | 1 | 13 | 14 | 12 | 1 | 61 | **12** | 54 | **11** |
| HCS3_NN | 1 | 4 | 5 | 4 | 1 | 4 | 4 | 18 | **1** | 19 | **1** |
| HCS7_CC | 2 | 11 | 9 | 10 | 5 | 3 | 4 | 35 | **8** | 33 | **6** |
| EH1 | 14 | 14 | 14 | 14 | 12 | 11 | 5 | 70 | **14** | 70 | **14** |
| RM | 10 | 12 | 13 | 11 | 10 | 2 | 5 | 50 | **10** | 51 | **10** |

Table 11. Product final ranks (PDFRs) obtained from three popular multivariate (multi-band) QIs: SAM (angle in range [0, 90 degrees]), it is a cost to be minimized; ERGAS ≥ 0, it is a cost to be minimized; "universal" (heterogeneous) Q4 (in range [0, 1]), it is a quality index to be maximized. Yellow highlight: first-best choice by quantitative quality estimation. Red highlight: second-best choice by quantitative quality estimation.

| Algorithm | SAM (cost index) | PDFR, SAM | ERGAS (cost index) | PDFR, ERGAS | Q4 (quality index) | PDFR, Q4 |
|---|---|---|---|---|---|---|
| PC1_NN_PA | 2.1950 | **3** | 2.2207 | **3** | 0.996300 | **4** |
| PC2_B | 2.0378 | **2** | 2.13219 | **2** | 0.996791 | **2** |
| PC3_CC | 2.0369 | **1** | 2.12856 | **1** | 0.996972 | **1** |
| GS1_NN | 3.0045 | **6** | 2.92803 | **10** | 0.993670 | **6** |
| GS2_B_PA | 2.6781 | **4** | 2.42699 | **5** | 0.996542 | **3** |
| GS3_CC_PA | 3.8092 | **5** | 2.40221 | **4** | 0.995557 | **5** |
| CN2_PA | 3.9176 | **7** | 2.5552 | **7** | 0.991644 | **7** |
| DWT1 | 5.9411 | **13** | 4.32709 | **13** | 0.918766 | **13** |
| DWT1_PA | 4.5145 | **10** | 2.80043 | **8** | 0.963695 | **11** |
| ATW2_PA | 4.1037 | **9** | 2.5721 | **6** | 0.972055 | **10** |
| HCS3_NN | 3.9481 | **8** | 2.92367 | **9** | 0.975420 | **9** |
| HCS7_CC | 5.6735 | **12** | 3.44937 | **11** | 0.960850 | **12** |
| EH1 | 6.1258 | **14** | 14.0384 | **14** | 0.437443 | **14** |
| RM | 5.1103 | **11** | 3.95332 | **12** | 0.987463 | **8** |

Table 12. Comparison between the new protocol's final ranks of Product QIs (PDFRs in Cases A and C, refer to Table 9; Case C is alternative to Case A, because the former omits cost index InvrsCrltnBivrt) and Product & Process QIs (PPFRs in Cases B and D, refer to Table 10; Case D is alternative to Case B, because the former omits cost index InvrsCrltnBivrt), the subjective ranks collected from human subjects (refer to Table 5) and the product final ranks (PDFRs) provided by three popular QIs, specifically, SAM, ERGAS and Q4, refer to Table 11. Green highlight: first-best and second-best choice by human subjects. Yellow highlight: first-best choice by quantitative quality estimation. Red highlight: second-best choice by quantitative quality estimation.

| Algorithm | PDFR, Case A | PPFR, Case B | PDFR, Case C | PPFR, Case D | Subjective quality assessment | | PDFR | | |
|---|---|---|---|---|---|---|---|---|---|
| | | | | | Spectral | Spatial | SAM | ERGAS | Q4 |
| PC1_NN_PA | 4 | 4 | 4 | 4 | C | A | 3 | 3 | 4 |
| PC2_B | 2 | 2 | 2 | 2 | B | A | 2 | 2 | 2 |
| PC3_CC | 3 | 3 | 3 | 3 | C | A | 1 | 1 | 1 |
| GS1_NN | 7 | 7 | 8 | 8 | C | D | 6 | 10 | 6 |
| GS2_B_PA | 6 | 5 | 6 | 6 | D | D | 4 | 5 | 3 |
| GS3_CC_PA | 5 | 6 | 5 | 5 | D | D | 5 | 4 | 5 |
| CN2_PA | 9 | 9 | 9 | 9 | F | E | 7 | 7 | 7 |
| DWT1 | 10 | 11 | 11 | 13 | B | F | 13 | 13 | 13 |
| DWT1_PA | 12 | 13 | 10 | 12 | B | F | 10 | 8 | 11 |
| ATW2_PA | 13 | 12 | 11 | 11 | B | F | 9 | 6 | 10 |
| HCS3_NN | 1 | 1 | 1 | 1 | A | B | 8 | 9 | 9 |
| HCS7_CC | 8 | 8 | 6 | 6 | B/C | E | 12 | 11 | 12 |
| EH1 | 14 | 14 | 14 | 14 | G | G | 14 | 14 | 14 |
| RM | 10 | 10 | 13 | 10 | C | E | 11 | 12 | 8 |

Table 13. The Spearman's rank correlation coefficient (SRCC) values, in range [-1, 1], generated from pairwise comparisons of ranked variables: ERGAS, average SAM, Q4, PDFR - Case C and PPFR - Case D, refer to Table 12. Unlike the Pearson's correlation coefficient (PCC), the SRCC assesses how well the relationship between two ranked variables can be described using a monotonically increasing or decreasing function, even if their relationship is not linear.

| | Spearman's rank correlation coefficient (SRCC) | | | | |
|---|---|---|---|---|---|
| | **ERGAS** | **SAM** | **Q4** | **PDFR, Case C** | **PPFR, Case D** |
| **ERGAS** | x | 0.9253 | 0.8593 | 0.6967 | 0.6725 |
| **SAM** | x | x | 0.9692 | 0.7495 | 0.7560 |
| **Q4** | x | x | x | 0.6659 | 0.7209 |
| **PDFR, Case C** | x | x | x | x | 0.9626 |
| **PPFR, Case D** | x | x | x | x | x |